%% file: iccv_2023_paper.tex
\documentclass[10pt,twocolumn,letterpaper]{article}

\def\paperID{11935} 
\def\printOption{final} 
\def\codeurl{https://github.com/yetigurbuz/generalized-sum-pooling} 

\input{packages.tex}

\input{definitions.tex}

\begin{document}

\title{Generalized Sum Pooling for Metric Learning}

\input{author_def.tex}

\input{authors.tex}

\maketitle

\input{abstract.tex}

\fancypagestyle{firststyle}
{
   \fancyhead{}
   \lhead{Accepted as a conference paper at ICCV 2023}
   \renewcommand{\headrulewidth}{0pt} 
}
\thispagestyle{firststyle}
\section{Introduction}
\label{sec:intro}
Distance metric learning (DML) addresses the problem of finding an embedding function such that the semantically similar samples are embedded close to each other while the dissimilar ones are placed relatively apart in the Euclidean sense. Although the prolific and diverse literature of DML includes various architectural designs \cite{kim2018attention, lin2018deep, ermolov2022hyperbolic}, loss functions \cite{musgrave2020metric}, and data-augmentation techniques \cite{roth2020revisiting, venkataramanan2022it}, many of these methods have a shared component: a convolutional neural network (CNN) followed by a global pooling layer, mostly global average pooling (GAP) \cite{musgrave2020metric}.



Common folklore to explain the effectiveness of GAP is considering each pixel of the CNN feature map as corresponding to a separate semantic entity \cite{gurbuz2023generalizable}. For example, spatial extent of one pixel can correspond to a \textit{"tire"} object making the resulting feature a representation for \textit{"tireness"} of the image. If this explanation is correct, the representation space defined via output of GAP is a convex combination of semantically independent representations defined by each pixel in the feature map. Although this folklore is later empirically studied in \cite[and references therein]{zeiler2014visualizing, zhou2016learning,  zhou2018interpreting} %
 and further verified for classification in \cite{gurbuz2019novel,xu2020attribute}, 
its algorithmic implications are not clear. If each feature is truly representing a different semantic entity, should we really average over all of them? Surely, some classes belong to the background and should be discarded as nuisance variables. Moreover, is uniform average of them the best choice? Aren't some classes more important than others? In this paper, we try to answer these questions within the context of metric learning. We propose a learnable and generalized version of GAP which learns to choose the subset of the semantic entities to utilize as well as weights to assign them while averaging.


In order to generalize the GAP operator to be learnable, we re-define it as a solution of an optimization problem. We let the solution space to include $0$-weight effectively enabling us to choose subset of the features as well as carefully regularize it to discourage degenerate solution of using all the features. Crucially, we rigorously show that the original GAP is a specific case of our proposed optimization problem for a certain realization. Our proposed optimization problem closely follows optimal transport based \textit{top-$k$} operators \cite{cuturi2019differentiable} and we utilize its literature to solve it. Moreover, we present an algorithm for an efficient computation of the gradients over this optimization problem enabling learning. A critical desiderata of such an operator is choosing subset of features which are discriminative and ignoring the background classes corresponding to nuisance variables. Although supervised metric learning losses provide guidance for seen classes, they carry no such information to generalize the behavior to unseen classes. To enable such a behavior, we adopt a \textit{zero-shot prediction loss} as a regularization term which is built on expressing the class label embeddings as a convex combination of attribute embeddings \cite{demirel2017attributes2classname, xu2020attribute}.

In order to validate the theoretical claims, we design a synthetic empirical study. The results confirm that our pooling method chooses better subsets and improve generalization ability. Moreover, our method can be applied with any DML loss as GAP is a shared component of them. We applied our method on 6 DML losses and test on 4 datasets. Results show consistent improvements with respect to direct application of GAP as well as other pooling alternatives.

\section{Related Work}
\label{sec:review}
\textbf{Our contributions.} Briefly, our contributions include that $i)$ we introduce a general formulation for weighted sum pooling, $ii)$ we formulate local feature selection as an optimization problem which admits closed form gradient expression without matrix inversion, and $iii)$ we propose a meta-learning based zero-shot regularization term to explicitly impose unseen class generalization to the DML problem. We now discuss the works which are most related to ours.

\textbf{Distance Metric Learning (DML).} Primary thrusts in DML include  $i)$ tailoring pairwise loss terms \cite{musgrave2020metric} that enforces the desired intra- and inter-class proximity constraints, $ii)$ pair mining \cite{roth2020revisiting}, $iii)$ generating informative samples \cite{ko2020embedding, liu2021noise, proxysynthesis, venkataramanan2022it}, and $iv)$ augmenting the mini-batches with virtual embeddings called \textit{proxies} \cite{wang2020cross, teh2020proxynca++}. To improve generalization; learning theoretic ideas  \cite{dong2020generalization, lei2021generalization, gurbuz2021asap}, disentangling class-discriminative and class-shared features \cite{lin2018deep, Roth_2019_ICCV}, intra-batch feature aggregation \cite{intrabatch,lim2022hypergraph}, ranking surrogates \cite{patel2022recall}, and further regularization terms \cite{Jacob_2019_ICCV, zhang2020deepSEC, profs, kim2021multi, roth2022non} are utilized. To go beyond of a single model, ensemble \cite{xuan2018deep, kim2018attention, Sanakoyeu_2019_CVPR, zheng2021deep2, zheng2021deep} and multi-task based approaches \cite{milbich2020diva, roth2021s2sd} are also used. Different to them, we propose a learnable pooling method generalizing GAP, a shared component of all of the mentioned works. Hence, our work is orthogonal to all of these and can be used jointly with any of them.

\textbf{Prototype-based pooling.} Most related to ours are trainable VLAD \cite{zhang2017deep,vlad} and optimal transport based aggregation \cite{kolouri2020wasserstein,mialon2021trainable}. Such methods employ similarities to the prototypes to form a vector of aggregated local features for each prototype and build ensemble of representations. Although their embeddings enjoy $\ell 2$ metric for similarity, they typically have very large sizes limiting their applicability for DML. Recently, reducing the dimension of VLAD embedding via non-linear transforms is addressed for DML \cite{kancoded}. Nevertheless, such methods still map a set of features to another set of features without discarding any and do not provide a natural way to aggregate the class-discriminative subset of the features. On the contrary, our pooling machine effectively enables learning to select discriminative features and maps a set of features to a single feature that is distilled from nuisance information.

\textbf{Attention-based pooling.} CroW \cite{crow}, Trainable-SMK \cite{tamsk}, and CBAM \cite{cbam} reweight the CNN features before pooling. They build on feature magnitude based saliency, assuming that the backbone functions must be able to zero-out nuisance information. Yet, such a requirement is restrictive for the parameter space and annihilation of the non-discriminative information might not be feasible in some problems. Similarly, attention-based weighting methods DeLF \cite{delf}, GSoP \cite{gsop} do not have explicit control on feature selection behavior and might result in poor models when jointly trained with the feature extractor \cite{delf}. Differently, our method unifies attention-based feature masking practices (\eg, \textit{convolution}, \textit{correlation}) with an efficient-to-solve optimization framework and lets us do away with engineered heuristics in obtaining the masking weights (\eg, \textit{normalization}, \textit{sigmoid}, \textit{soft-plus}) without restricting the solution space.





\textbf{Optimal transport (OT) based operators.}  OT distance \cite{cuturi2013sinkhorn} to match local features is used as the DML distance metric instead of $\ell2$ in \cite{zhao2021towards}. Despite effective, replacing $\ell2$ with OT increases memory cost for image representation as well as computation cost for the distance computation. Different to them, we shift OT based computation in pooling (\ie, feature extraction) stage while having OT's merits and hence, do not affect the memory and computation costs of the inference by sticking to $\ell2$ metric. Moreover, our feature selection and aggregation formulation has close relation to OT \cite{cuturi2013sinkhorn} based top-$k$ \cite{xie2020differentiable}, ranking \cite{cuturi2019differentiable} and aggregation \cite{kolouri2020wasserstein, mialon2021trainable, naderializadeh2021pooling} operators which are not effectively applied to DML before. Their aggregation is built on concatenating the feature ensembles resulting in very large embedding sizes. What makes our method different is the unique way we formulate the feature selection problem to fuse aggregation into it (\emph{see Appendix for technical details}). Our selection operator allows computationally appealing and matrix inversion free gradient computation unlike its OT based counterparts \cite{luise2018differential}.

\input{paper/figures/fig_method.tex}

\section{Preliminaries}
\label{sec:notations}
Consider the data distribution $p_{\mathcal{X}\sxtimes\mathcal{Y}}$ over $\mathcal{X}\sxtimes \mathcal{Y}$ where $\mathcal{X}$ is the space of data points and $\mathcal{Y}$ is the space of labels. Given \textit{iid.} samples from $p_{\mathcal{X}\sxtimes\mathcal{Y}}$ as $\{(x_i, y_i) \}$, distance metric learning problem aims to find the parameters $\theta$ of an embedding function $e(\cdot; \theta): \mathcal{X} \rightarrow \RR^d$ such that the Euclidean distance in the space of embeddings is consistent with the label information where $d$ is the embedding dimension. More specifically, \mbox{$\Vert e(x_i;\theta) - e(x_j;\theta)\Vert_2$} is small whenever $y_i=y_j$, and large whenever $y_i \neq y_j$. In order to enable learning, this requirement is represented via loss function $l((x_i, y_i), (x_j,y_j); \theta)$ (\eg, \textit{contrastive} \cite{wu2017sampling}, \textit{triplet} \cite{schroff2015facenet}, \textit{multi-similarity} \cite{Wang_2019_CVPR_MS}).

The typical learning mechanism is gradient descent of an empirical risk function defined over a batch of data points $B$. To simplify notation throughout the paper, we will use $b=\{b(i)\mid x_i,y_i\in B \}_i$ to index the samples in a batch. Then, the typical empirical risk function is defined as:

\begin{equation} \label{eq:empi}
\mathcal{L}_\text{DML}(b; \theta) \coloneqq \tfrac{1}{\vert b \vert^2}\textstyle\sum\limits_{i \in b}\textstyle\sum\limits_{j\in b}l((x_i, y_i), (x_j,y_j); \theta)\, .
\end{equation}

We are interested specific class of embedding functions where a global average pooling is used. Specifically, consider the composite function family $e = g \circ f$ such that $g$ is pooling and $f$ is feature computation. We assume a further structure over the functions $g$ and $f$. The feature function $f$ maps the input space $\mathcal{X}$ into $\RR^{w\sxtimes h \sxtimes d}$ where $w$ and $h$ are spatial dimensions. Moreover, $g$ performs averaging as;

\begin{equation}\label{eq:gap}
    g(f(x;\theta)) = \frac{1}{w \cdot h} \textstyle\sum\limits_{i\in [w \cdot h]} f_i\, ,
\end{equation}
where $[n]{=}{1, \ldots, n}$ and we let $f_i{\in}\RR^d$ denote $i^\text{th}$ spatial feature of $f(x;\theta)$ to simplify notation. In the rest of the paper, we generalize the pooling function $g$ into a learnable form and propose an algorithm to learn it.

\section{Method}
\label{sec:method}

Consider the pooling operation in \eqref{eq:gap}, it is a simple averaging over pixel-level feature maps ($f_{i}$). As we discuss in \cref{sec:intro}, one explanation for the effectiveness of this operation is considering each $f_{i}$ as corresponding to a different semantic entity corresponding to the spatial extend of the pixel, and the averaging as convex combination over these semantic classes. Our method is based on generalizing this averaging such that a specific subset of pixels (correspondingly subset of semantic entities) are selected and their weights are adjusted according to their importance.

We generalize the pooling \eqref{eq:gap} in \cref{sec:generalized_gap} by formulating a feature selection problem in which we prioritize a subset of the features that are closest to some trainable prototypes. If a feature is to be selected, its weight will be high. We then formulate our pooling operation as a differentiable layer to facilitate the prototype learning along with the rest of the embedding function parameters in \cref{sec:learnable_gap}. We learn the prototypes with class-level supervision, however in metric learning, learned representations should generalize to unseen classes. Thus, we introduce a zero-shot prediction loss to regularize prototype training for zero-shot setting in \cref{sec:zero_shot}.

\subsection{Generalized Sum Pooling as a Linear Program}
\label{sec:generalized_gap}

Consider the pooling function $g$ with adjustable weights as $g(f(x;\theta); \omega)=\sum_{i\in[n]} p_i f_i$ where $n=w\,h$. Note that, $p_i {=} \nicefrac{1}{n}$ corresponds to average pooling. Informally, we want to control the weights to ease the metric learning problem. Specifically, we want the weights corresponding to background classes to be $0$ and the ones corresponding to discriminative features to be high.

If we were given representations of discriminative semantic entities, we could simply compare them with the features ($f_i$) and choose the ones with high similarity. Our proposed method is simply learning these representations and using them for weight computations. We first discuss the weight computation part before discussing learning the representations of prototypes.  

Assume that there are $m$ discriminative semantic entities which we call \textit{prototypes} with latent representations $\omega=\lbrace\omega_i\rbrace_{i\in[m]}$ of appropriate dimensions (same as $f_i$). Since we know that not all features ($\{f_i\}_{i \in [n]}$) are relevant, we need to choose a subset of $\{f_i\}_{i \in [n]}$. We perform this top-$k$ selection process by converting it into an optimal transport (OT) problem. 
 
Consider a cost map $c_{ij}=\Vert \bar{\omega}_i \shortminus \bar{f}_j\Vert_2$ which is an $m$ (number of prototypes) by $n$ (number of features)  matrix representing the closeness of prototypes $\omega_i$ and features $f_j$ after some normalization \mbox{$\bar{u}=\nicefrac{u}{{\max\lbrace1,\Vert u\Vert_2\rbrace}}$}. We aim to find a transport map $\pi$ which re-distributes the uniform mass from features to prototypes. Since we do not have any prior information over features, we also consider its marginal distribution (importance of each feature to begin with) to be uniform. As we need to choose a subset, we set $\mu{\in}[0,1]$ ratio of mass to be transported. The resulting OT problem is:
\begin{equation}\tag{P1}\label{eq:problem}
\rho^{\ast}, \pi^{\ast} = \! \argmin_{\rho,\pi\geqslant0}
\! \textstyle\sum_{ij}c_{ij}\pi_{ij} \text{\quad s.to }
{\small\begin{array}[t]{r} \rho_j {+} \Sigma_i\pi_{ij}{=}\tfrac{1}{n}\\[-0.2ex]\Sigma_{ij}\pi_{ij}{=}\mu \end{array}} .
\end{equation}
Different to typical OT literature, we introduce decision variables $\rho$ to represent residual weights to be discarded. Specifically modelling discarded weight instead of enforcing another marginalization constraint is beneficial beyond stylistic choices as it allows us to very efficiently compute gradients. When the introduced transport problem is solved, we perform weighting using residual weights as:
\begin{equation}\label{eq:pooling_weights}
g(f(x;\theta); \omega) = \textstyle\sum_i p_i f_i = \textstyle\sum_i \tfrac{\nicefrac{1}{n}-\rho^\ast_i}{\mu}f_i\,.
\end{equation}

Given set of prototypes  $\lbrace\omega_i\rbrace_{i\in[m]}$, solving the problem in \eqref{eq:problem} is a strict generalization of GAP since setting $\mu=1$ recovers the original GAP. We formalize this equivalence in the following claim.

\begin{theorem}\label{claim:gap}
If $\mu=1$, the operation in \eqref{eq:pooling_weights} reduces to global average pooling in \eqref{eq:gap}.
\end{theorem}

We defer the proof to Appendix. Having generalized GAP to a learnable form, we introduce a method to learn the prototypes $\lbrace\omega_i\rbrace_{i\in[m]}$ in the next section.

\subsection{GSP as a Differentiable Layer}
\label{sec:learnable_gap}
Consider the generalized form of pooling, defined as solution of \eqref{eq:problem}, as a layer of a neural network. The input is the feature vectors $\{f_i\}_{i \in [n]}$, the learnable parameters are prototype representations $\lbrace\omega_i\rbrace_{i\in[m]}$, and the output is residual weights $\rho^\ast$. To enable learning, we need partial derivatives of $\rho^\ast$ with respect to $\lbrace\omega_i\rbrace_{i\in[m]}$. However, this function is not smooth. More importantly it requires the $\mu$ parameter to be known a priori. 

We use a toy example to set the stage for rest of the formulation. Consider a $10{\sxtimes}10{\sxtimes}3$ feature map visualized as RGB-image in \cref{fig:smooth_effect} and corresponding two prototypes with representations $(1, 0, 0)$ (red) and $(0, 0, 1)$ (blue). The true $\mu=0.5$ since the half of the image corresponds to red and blue, and other half is background class of green. Consider an under-estimation of $\mu=0.2$, the global solution (shown as linear programming) is explicitly ignoring informative pixels (part of red and blue region). To solve this issue, we use entropy smoothing which is first introduced in \cite{cuturi2013sinkhorn} to enable fast computation of OT. Consider the entropy smoothed version of the original problem in \eqref{eq:problem} as:

\begin{equation}\tag{P2}\label{eq:smoothed_problem}
\rho^{(\varepsilon)}, \pi^{(\varepsilon)} = \!\!\!\!\! \argmin_{\substack{\rho,\pi\geqslant0 \\
\rho_j + \Sigma_i\pi_{ij}=\nicefrac{1}{n}\\
\Sigma_{ij}\pi_{ij}=\mu}}
\!\!\!\!\! \textstyle\sum_{ij}c_{ij}\pi_{ij} +
\tfrac{1}{\varepsilon}(H(\pi) + H(\rho)),
\end{equation}
and obtain pooling weights by replacing $\rho^\ast$ with $\rho^{(\varepsilon)}$ in \eqref{eq:pooling_weights}, where $H(u)\coloneqq\Sigma_i u_{i}\log u_i$.

%
\input{paper/figures/fig_smoothing.tex}
%
When smoothing is high ($\varepsilon{\to}0$), the resulting solution is uniform over features similar to GAP. When it is low, the result is similar to top-$k$ like
behavior. For us, $\varepsilon$ controls the trade-off between picking $\mu$ portion of the features that are closest to the prototypes and including as much features as possible for weight transfer. We further visualize the solution of the entropy smoothed problem in \cref{fig:smooth_effect} showing desirable behavior even with underestimated $\mu$.

Beyond alleviating the under-estimation of $\mu$ problem, entropy smoothing also makes the problem strictly convex and smooth. Thus, the solution of the problem enables differentiation and in fact, admits closed-form gradient expression. We state the solution of \eqref{eq:smoothed_problem} and its corresponding gradient in the following propositions and defer their proofs to Appendix.
\begin{proposition} \label{proposition1}
Given initialization $t^{(0)} = 1$, consider the following iteration:
\begin{equation*}
    \begin{split}
        &\rho^{(k{+1})} = \nicefrac{1}{n}\,(1 + t^{(k)}\exp({\shortminus}\varepsilon c)^\intercal\bm{1}_m)^{\shortminus 1}\!,\\
        &t^{(k{+}1)}=\mu\,(\bm{1}_m^\intercal\exp({\shortminus}\varepsilon c) \rho^{(k{+}1)})^{\shortminus 1}
    \end{split}
\end{equation*}
where $\exp$ and $(\cdot)^{\shortminus 1}$ are element-wise and $\bm{1}_m$ is $m$-dimensional vector of ones. Then, $(\rho^{(k)}, t^{(k)})$ converges to the solution of \eqref{eq:smoothed_problem} defining transport map via \mbox{$\pi^{(k)} = t^{(k)}\exp({\shortminus}\varepsilon c)Diag(\rho^{(k)})$}.
\end{proposition}

\begin{proposition} \label{proposition2}
Consider any differentiable loss function $\mathcal{L}$ as a function of $(\rho,\pi)$. Given gradients $\frac{\partial \mathcal{L}}{\partial \rho}$ and $\frac{\partial \mathcal{L}}{\partial \pi}$, with $(\rho,\pi)$ is the solution of \eqref{eq:smoothed_problem}. Let $q=\rho\odot\frac{\partial \mathcal{L}}{\partial \rho} + (\pi\odot\frac{\partial \mathcal{L}}{\partial \pi})^\intercal\bm{1}_m$ and $\eta=(\rho\odot\frac{\partial \mathcal{L}}{\partial \rho})^\intercal\bm{1}_n\shortminus n\,q^\intercal\rho$, the gradient of $\mathcal{L}$ with respect to 
$c$ reads:
\begin{equation}
\frac{\partial \mathcal{L}}{\partial c} = \shortminus\varepsilon\Big(\pi\odot\frac{\partial \mathcal{L}}{\partial \pi} - n\pi Diag\big(q- \tfrac{\eta}{1\shortminus\mu\shortminus n\rho^{\intercal}\rho}
\big)\rho\Big)\,\, ,
\end{equation}
where $\odot$ denotes element-wise multiplication.
\end{proposition}

Proposition \ref{proposition1} and \ref{proposition2} suggest that our feature selective pooling can be implemented as a differentiable layer. Moreover, Proposition \ref{proposition2} gives a matrix inversion free computation of the gradient with respect to the costs unlike optimal transport based operators \cite{luise2018differential}. Thus, the prototypes, $\omega$, can be jointly learned with the feature extraction efficiently.

\subsection{Cross-batch Zero-shot Regularization}
\label{sec:zero_shot}

We formulated a prototype based feature pooling and learn the prototypes using class labels. Simply classifying the labels as prototypes is a degenerate solution.  We rather want the prototypes to capture transferable attributes so that the learning can be transferred to the unseen classes as long as the attributes are shared. Learning with prototype based pooling shapes the embedding geometry in such a way that we have clusters corresponding to the prototypes in the embedding space. We want such clusters to have transferable semantics rather than class-specific information. To enable this, we now formulate a mechanism to predict class embedding vectors from the prototype assignment vectors and use that mechanism to tailor a loss regularizing the prototypes to have transferable representations. 

Our feature selection layer should learn discriminative feature prototypes $\omega$ using top-down label information. Consider two randomly selected batches $(b_1, b_2)$ of data sampled from the distribution. If the prototypes are corresponding to discriminative entities, the weights transferred to them (\ie, marginal distribution of prototypes) should be useful in predicting the classes and such behavior should be consistent between batches for zero-shot prediction. Formally, if one class in $b_2$ does not exist in $b_1$, a predictor on class labels based on marginal distribution of prototypes for each class of $b_1$ should still be useful for $b_2$. DML losses do not carry such information. We thus formulate a zero-shot prediction loss to enforce such zero-shot transfer.
 
We consider that we are given an embedding vector $\upsilon_i$ for each class label $i$, \ie, $\Upsilon=[\upsilon_i]_{i\in[c]}$ for $c$-many classes. We are to predict such embeddings from the marginal distribution of the prototypes. In particular, we use linear predictor $A$ to predict label embeddings as \mbox{$\hat{\upsilon} = A\,z$} where $z$ is the normalized distribution of the weighs on the prototypes;
\begin{equation}\label{eq:attr_weights}
z = \tfrac{1}{\mu}\textstyle\sum_i\pi^{(\varepsilon)}_i \quad \text{where} \quad \pi^{(\varepsilon)}=[\pi^{(\varepsilon)}_i]_{i\in[n]}\, .
\end{equation}
If we consider the prototypes as semantic vectors of some auxiliary labels such as \textit{attributes} similar to zero-shot learning (ZSL) \cite{demirel2017attributes2classname, xu2020attribute, huynh2020fine}, then we can interpret $z$ as \textit{attribute} predictions. Given attribute predictions $\{z_i\}_{i\in b}$ and corresponding class embeddings for a batch $b$ we fit the predictor as;
\begin{equation}\tag{P3}\label{eq:regression}
A_b \! = \!\!\! \argmin_{A=[a_i]_{i\in[m]}} \!\!\! \textstyle\sum_{i\in b} \Vert A\,z_i \shortminus \upsilon_{y_i} \Vert^2_2 + \epsilon\textstyle\sum_{i\in[m]}\Vert a_i\Vert^2_2\, ,
\end{equation}
which admits a closed form expression enabling back propagation $A_b= \Upsilon_b\,(Z_b\T Z_b + \epsilon I)^{\shortminus 1}Z_b\T$ where $\Upsilon_b=[\upsilon_{y_i}]_{i\in b}$, $Z_b=[z_i]_{i\in b}$. In practice, we are not provided with the label embeddings $\Upsilon=[\upsilon_i]_{i\in[c]}$. Nevertheless, having a closed-form expression for $A_b$ enables us to exploit a meta-learning scheme like \cite{bertinetto2018meta} to formulate a zero-shot prediction loss to learn them jointly with the rest of the parameters.

Specifically, we split a batch $b$ into two\footnote{Although we considered the simplest form which already worked well, repeating this splitting process can be beneficial.} as $b_1$ and $b_2$ such that classes are disjoint. We then estimate attribute embeddings $A_{b_k}$ according to \eqref{eq:regression} using one set and use that estimate to predict the label embeddings of the other set to form zero-shot prediction loss. Formally, our loss becomes:
\begin{equation}
\begin{split}
 \mathcal{L}_\text{ZS}(b; \theta) &= \tfrac{1}{\vert b_2 \vert}\textstyle\sum\limits_{i\in b_2} \log\big(1+\textstyle\sum\limits_{j\in[c]}\mathrm{e}^{(\upsilon_j\shortminus \upsilon_{y_i})\T A_{1}\,z_i}\big)\\
 &+ \tfrac{1}{\vert b_1 \vert}\textstyle\sum\limits_{i\in b_1} \log\big(1+\textstyle\sum\limits_{j\in[c]}\mathrm{e}^{(\upsilon_j\shortminus \upsilon_{y_i})\T A_{2}\,z_i}\big),
\end{split}
\end{equation}
\ie, rearranged \textit{soft-max cross-entropy}  where $A_k{=}A_{b_k}$ with the abuse of notation, and $\theta=\lbrace\theta_f, \omega, \Upsilon\rbrace$ (\ie, CNN parameters, prototype vectors, label embeddings). 



We learn attribute embeddings (\ie, columns of $A$) as sub-task and can define such learning as a differentiable operation. Intuitively, such a regularization should be useful in better generalization of our pooling operation to unseen classes since attribute predictions are connected to prototypes and the local features. We combine this loss with the metric learning loss using $\lambda$ mixing (\ie, $(1{\shortminus}\lambda)\mathcal{L}_\text{DML} + \lambda\mathcal{L}_\text{ZS}$) and jointly optimize.

\input{paper/figures/fig_ablation_synthetic}
\subsection{Implementation Details}
\label{sec:implementation}

\textbf{Embedding function.} For the embedding function $f(\cdot;\theta)$ we use ResNet20 \cite{he2016identity} for Cifar \cite{cifar} experiments, and ImageNet \cite{russakovsky2015imagenet} pretrained BN-Inception \cite{normalization2015accelerating} for the rest. We exploit architectures until the output before the global average pooling layer. We add a per-pixel linear transform (\ie, $1{\sxtimes}1$ convolution), to the output to obtain the local embedding vectors of size $128$.

\textbf{Pooling layer.} For baseline methods, we use global average pooling. For our method, we perform parameter search and set the hyperparameters accordingly. Specifically, we use 64- or 128-many prototypes depending on the dataset. We use $\varepsilon{=}0.5$ for proxy-based losses and $\varepsilon{=}5.0$ for non-proxy losses. For the rest, we set  $\mu{=}0.3$, $\epsilon{=}0.05$, $\lambda{=}0.1$ and we iterate until $k{=}100$ in \cref{proposition1}. The embedding vectors upon pooling are $\ell2$ normalized to have unit norm.

\section{Experiments}
We start our empirical study with a synthetic study validating the role of GAP in learning and the impact of GSP on the feature geometry. We further examine the effectiveness of our generalized sum pooling in metric learning for various models and datasets. We further perform ablation studies for the implications of our formulation as well as effects of the hyperparameters. We share the implementation details and the complete Tensorflow \cite{abadi2016tensorflow} code base in the supplemental materials.

\subsection{Analysis of the Behavior}
\label{sec:synthetic}

\textbf{Synthetic study.} We designed a synthetic empirical study to evaluate GSP in a fully controlled manner. We considered 16-class problem such that classes are defined over trainable tokens. In this setting, tokens correspond to semantic entities but we choose to give them a specific working to emphasize that they are trained as part of the learning. Each class is defined with 4 distinct tokens and there are also 4 background tokens shared by all classes. For example, a \emph{"car"} class would have tokens like \emph{"tire"} and \emph{"window"} as well as background tokens of \emph{"tree"} and \emph{"road"}. We sampled class representations from both class specific and background tokens according to a mixing ratio $\Tilde{\mu}\sim\mathcal{N}(0.5, 0.1)$. Such a 50-many feature collection corresponds to a training sample (\ie, we are mimicking CNN's output with trainable tokens). We then obtained global representations using GAP and GSP. We visualize the geometry of the embedding space in \cref{fig:synthetic}-(a) along with the DML performances. With GAP, we observe overlapping class convex hulls resulting in poor DML performance. However, GSP gives well separated class convex hulls, further validating that it learns to ignore background tokens.

\textbf{Selective pooling.} We further extended this synthetic study to image domain by considering the 20 \textit{super-classes} of Cifar100 dataset \cite{cifar}
where each has 5 sub-classes. For each super-class, we split the sub-classes for train (2), validation (1), and test (2). We consider 4 super-classes as the shared classes and compose $4{\sxtimes}4$-stitched collage images for the rest 16 classes (see supplementary material for details). In particular, we sample an image from a class and then sample 3 images from shared classes (\cref{fig:cifarcollage}). We used ResNet20 backbone pretrained on Cifar100 classification task and followed the implementation explained in \cref{sec:implementation}. We provide the evaluation results in \cref{fig:synthetic}-(b). GSP and the proposed zero shot loss effectively increase MAP@R. We also provide sample train and test images to showcase that our pooling can transfer well to unseen classes.

\input{supplementary/figures/fig_cifar_collage}

\textbf{Zero-shot regularization.} We also evaluated the zero-shot prediction performance of the attribute vectors. We trained on Cifar10 
dataset with 8 prototypes using ProxyNCA++ \cite{teh2020proxynca++} (PNCA) loss with and without $\mathcal{L}_\text{ZS}$. We then extracted attribute histograms for each class and visualized them in \cref{fig:attributes_eg}. We observe transferable representations with $\mathcal{L}_\text{ZS}$ and we visually show in \cref{fig:attributes_eg} that the semantic entities represented by the prototypes transfer across classes. We quantitatively evaluated such behavior by randomly splitting the classes into half and apply cross-batch zero-shot prediction explained in \cref{sec:zero_shot}. Namely, we fit $A$ in \eqref{eq:regression} for one subset and used it to predict the class embeddings for the other set. We pre-computed class embeddings from the dataset as the class mean. To this end, our evaluation assesses generalization of both the features and the prototypes. We used MAP with both $\ell2$ distance and \textit{cosine} similarity in our evaluation. We repeated the experiment 1000 times. We observe in \cref{fig:attributes_eg} that zero-shot performance of the prototypes learned with $\mathcal{L}_\text{ZS}$ is substantially superior. We also see that our feature aggregation method enables approximate localization of the semantic entities. Recent ZSL approaches \cite{huynh2020fine, xu2020attribute} can provide attribute localization and share a similar spirit with our method. However, attribute annotations must be provided for those methods whereas we exploit only class labels to extract attribute-like features. Thus, our method can be considered as an attribute-unsupervised alternative  to them.

\input{paper/figures/fig_ablation_zsl_attr}

\input{paper/figures/fig_summary}

\subsection{Deep Metric Learning Experiments}
We largely rely on the recent work explicitly studying the fair evaluation strategies for metric learning \cite{roth2020revisiting, musgrave2020metric, fehervari2019unbiased}. Specifically, we follow the procedures proposed in \cite{musgrave2020metric} to evaluate our method as well as the other methods. We additionally follow the relatively old-fashioned conventional procedure \cite{oh2016deep} for the evaluation of our method and provide those results in the supplementary material. We provide full detail of our experimental setup in the supplementary material for complete transparency and reproducibility.

\textbf{Datasets.} CUB \cite{wah2011caltech}, Cars196 \cite{krause2014submodular}, In-shop \cite{liu2016deepfashion}, and SOP \cite{oh2016deep} with typical DML data augmentation \cite{musgrave2020metric}.

\textbf{Evaluation metrics.} We report mean average precision (MAP@R) at R where R is defined for each query and is the total number of true references of the query. 

\textbf{Hyperparameters.} We use Adam \cite{kingma2014adam} optimizer with learning rate $10^{\shortminus 5}$, weight decay $10^{\shortminus 4}$, batch size 32 (4 per class). We train 4-fold: 4 models (1 for each $\nicefrac{3}{4}$ train set partition).

\textbf{Evaluation.} Average performance (128D) where each of 4-fold model is trained 3 times resulting in realization of $3^4{=}81$ different model collections. In our results we provide mean of 81 evaluations.

\textbf{Baselines.} We implement our method on top of and compare with \textit{Contrastive (C2)}: Contrastive with positive margin \cite{wu2017sampling}, \textit{MS}: Multi-similarity \cite{Wang_2019_CVPR_MS}, \textit{Triplet}: Triplet \cite{schroff2015facenet}, \textit{XBM}: Cross-batch memory \cite{wang2020cross} with contrastive loss \cite{hadsell2006dimensionality}, \textit{PNCA}: ProxyNCA++ \cite{teh2020proxynca++}, \textit{PAnchor}: ProxyAnchor \cite{kim2020proxy}.

\subsubsection{Results}
We compared our method (GSP) against direct application of GAP with 6 DML methods in 4 datasets. We also evaluated 14 additional pooling alternatives on \textit{Ciffar Collage} and CUB datasets. We provide the tabulated results in supplementary material. Based on CUB performances, we picked generalized mean pooling (GeMean)  \cite{gmeanp} and DeLF \cite{delf} to compare against in 4 DML benchmarks. We also evaluated max pooling (GMP) and its combination with GAP as we typically observe GAP+GMP in the recent works \cite{venkataramanan2022it,teh2020proxynca++,kim2020proxy, wang2020cross}. We also applied our method with GMP (GMP+GSP) and with GeMean (GeMean+GSP) to show that per channel selection is orthogonal to our method and thus, GSP can marginally improve those methods as well. 

We present the tabulated evaluation results in \cref{tab:all_results}, while \cref{fig:test_summary_gsp} provides a concise summary of the relative MAP@R orderings of the methods employing 128D embeddings. We observe consistent improvements upon direct application of GAP in all datasets. On the average, we consistently improve the baselines ${\approx} 1\%$ points in MAP@R. Our improvement margins are superior to ones of attention based DeLF pooling. We improve state-of-the-art (SOTA) XBM method up to $2\%$ points, which is a good evidence that application of GSP is not limited to loss terms but can be combined with different DML approaches. We also consistently improve GMP and GeMean pooling methods in all datasets, yet another evidence that our method can be combined with max pooling based methods.

We additionally evaluated our method with different architectures and methods in conventional setting \cite{oh2016deep} for the comparison with SOTA. The tabulated results are provided in supplementary material (\S~1.1), where we observe that we achieve SOTA performances with XBM \cite{wang2020cross} and LIBC \cite{intrabatch} methods.

\input{paper/figures/table_components}

\subsubsection{Ablations}
\label{sec:ablation}

\textbf{Effect of $\vect{\mathcal{L}_\text{ZS}}$.} We empirically showed the effect of $\mathcal{L}_\text{ZS}$ on learned representations in \cref{sec:synthetic}. We further examined the effect of $\mathcal{L}_\text{ZS}$ quantitatively by enabling/disabling it. We also evaluated its effect without GSP by setting $\mu{=}1$ where we used GAP with attribute vectors. The results are summarized in \cref{tab:components} showing that both components improve the baseline and their combination brings the best improvement. We observe similar behavior in \textit{Cifar Collage} experiment (\cref{fig:synthetic}-(b)) where the effect of $\mathcal{L}_\text{ZS}$ is more substantial.

\textbf{Computation efficiency.} Through a series of $k$ iterations, our pooling mechanism utilizes lightweight matrix-vector products to determine the pooling weights. While back propagation can be achieved through \emph{automatic-differentiation}, 
%
%
it can become computationally intensive as $k$ increases for certain problems. However, our pooling mechanism boasts a desirable feature of having a closed-form gradient expression for its backward computation, resulting in minimal scaling of computation load as $k$ increases, as evidenced by our analysis in \cref{fig:ablation_duo}. We further provided the inference times for
various $k$ in supplementary material (\S~1.5).



\textbf{Effect of $\vect{\mu}$.} As outlined in \cref{sec:learnable_gap}, GSP is similar to top-$k$ operator with an adaptive $k$ thanks to entropy smoothing. We empirically validated such behavior by sweeping $\mu$ parameter controlling top-$k$ behavior. The results, plotted in \cref{fig:ablation_duo}, show similar performance for lower $\mu$ values, with a decrease as $\mu$ increases, possibly due to overestimation of the foreground ratio. Hence a suggested value for $\mu$ is $0.3$.




\input{paper/figures/fig_ablation_duo}

\input{supplementary/figures/table_all_results}
\section{Conclusion}
We proposed a learnable and generalized version of GAP. Our proposed generalization GSP is a trainable pooling layer that selects the feature subset and re-weight it during pooling. To enable effective learning of our layer, we also proposed a cross-batch regularization improving zero-shot transfer. With extensive empirical studies, we validated the effectiveness of the proposed pooling layer in various DML benchmarks. We also established and empirically validated a valuable link between the computed transport maps for pooling and the prototypes, enabling attribute learning via meta-learning without explicit localized annotations. We believe such a connection is interesting, and has potential to offer pathways for enhanced embedding learning for DML as well as unsupervised attribute learning.

{\small
\bibliographystyle{IEEEbib-abbrv}
\bibliography{iccv_2023_paper.bbl}
}

\input{supplementary/supplementary_append}

\end{document}

%% file: packages.tex
\def\printFinal{final}

\usepackage{iccv}

\usepackage[accsupp]{axessibility}  

\usepackage{import}

\usepackage[utf8]{inputenc} 
\usepackage[T1]{fontenc}    

\usepackage{amsmath}
\usepackage{amsthm}
\usepackage{mathtools}
\usepackage{commath}
\usepackage{mathrsfs}
\usepackage{amssymb}
\usepackage{amsfonts}       
\usepackage{bbm}
\usepackage{bm}
\usepackage{fancyhdr}
\usepackage{cancel}
\usepackage{nicefrac}       
\usepackage{pifont}
\newcommand{\cmark}{\ding{51}}%
\newcommand{\xmark}{\ding{55}}%

\usepackage{microtype}      
\usepackage{fontawesome } 

\usepackage{algorithm}
 \usepackage[noend]{algpseudocode}
\usepackage{eqparbox}

\usepackage{wrapfig}

\usepackage{stfloats}

\usepackage[pdftex]{graphicx}
\usepackage{epsfig}
\usepackage[utf8]{inputenc}
\usepackage{tabularx}
\usepackage{booktabs, multicol, multirow}	
\usepackage[table, xcdraw]{xcolor}

\usepackage{caption}
\usepackage{subcaption}

\usepackage{appendix}

\usepackage{cite}[sorted]
\usepackage{url}            

\ifx \printOption \printFinal
\usepackage[breaklinks=true,bookmarks=false]{hyperref}
\iccvfinalcopy
\else
\usepackage[pagebackref=true,breaklinks=true,letterpaper=true,colorlinks,bookmarks=false]{hyperref}
\fi

\usepackage[capitalize]{cleveref}
\crefname{section}{Sec.}{Secs.}
\Crefname{section}{Section}{Sections}
\Crefname{table}{Table}{Tables}
\crefname{table}{Tab.}{Tabs.}



%% file: definitions.tex
\def\iccvPaperID{\paperID} 

\newcommand{\vect}[1]{\boldsymbol{#1}}
\DeclareMathOperator*{\argmin}{arg\,min}

\newcommand{\RR}{\ensuremath{\mathbb{R}}}
\newcommand{\Real}{I\!\!R} 

\DeclareMathSymbol{\shortminus}{\mathbin}{AMSa}{"39}

\newcommand{\sxtimes}{\mathsf{x}\mskip1mu}
\newcommand{\T}{\ensuremath{^\intercal}}

\usepackage{xspace}
\makeatletter
\DeclareRobustCommand\onedot{\futurelet\@let@token\@onedot}
\def\@onedot{\ifx\@let@token.\else.\null\fi\xspace}

\def\eg{\emph{e.g}\onedot} 
\def\ie{\emph{i.e}\onedot}

\makeatother

\theoremstyle{plain}
\newtheorem{theorem}{Theorem}[section]
\newtheorem{proposition}[theorem]{Proposition}

\theoremstyle{definition}

\theoremstyle{remark}

\numberwithin{equation}{section}

\makeatletter
\def\BState{\State\hskip-\ALG@thistlm}
\makeatother



%% file: author_def.tex
\def\yeti{Yeti~Z.~G\"{u}rb\"{u}z}
\def\ada{Ada~G\"{o}rg\"{u}n}
\def\aydin{A.~Ayd{\i}n~Alatan}
\def\ozan{Ozan~\c{S}ener}

\def\instmetu{OGAM and METU}
\def\addrssmetu{METU, TR}

\def\insttub{RSiM, TU Berlin}
\def\addrsstub{TU Berlin, DE}

\def\instuni{Faculty of Elect. Elec. Eng. and Computer Science}
\def\addrssuni{Somewhere in Europe}

\def\instapple{Intel Labs}

\def\emailyeti{y.guerbuez@tu-berlin.de}
\def\emailala{alatan@metu.edu.tr}
\def\emailada{ada.gorgun@metu.edu.tr}


\newcommand{\affinfo}{}

%% file: authors.tex




\def\fdagger{\rotatebox[origin=c]{180}{$\dagger$}}
\renewcommand\footnotemark{}



\author{\begin{tabular}{ccc}
\yeti\textsuperscript{$\dagger$}\thanks{\textsuperscript{$\dagger$}Affiliated with OGAM-METU during the research.} & \ozan & \aydin \\
     \insttub & \instapple & \instmetu \\
\end{tabular}
}

%% file: abstract.tex
\begin{abstract}
\label{sec:abstract}
A common architectural choice for deep metric learning is a convolutional neural network followed by global average pooling (GAP). Albeit simple, GAP is a highly effective way to aggregate information. One possible explanation for the effectiveness of GAP is considering each feature vector as representing a different semantic entity and GAP as a convex combination of them. Following this perspective, we generalize GAP and propose a learnable generalized sum pooling method (GSP). GSP improves GAP with two distinct abilities: i) the ability to choose a subset of semantic entities, effectively learning to ignore nuisance information, and ii) learning the weights corresponding to the importance of each entity. Formally, we propose an entropy-smoothed optimal transport problem and show that it is a strict generalization of GAP, \ie, a specific realization of the problem gives back GAP. We show that this optimization problem enjoys analytical gradients enabling us to use it as a direct learnable replacement for GAP. We further propose a zero-shot loss to ease the learning of GSP. We show the effectiveness of our method with extensive evaluations on 4 popular metric learning benchmarks. Code is available at: \href{\codeurl}{GSP-DML Framework}
\end{abstract}


%% file: paper/figures/fig_method.tex
\begin{figure*}[!t]
  \centering
  \centerline{\includegraphics[width=1\linewidth,keepaspectratio]{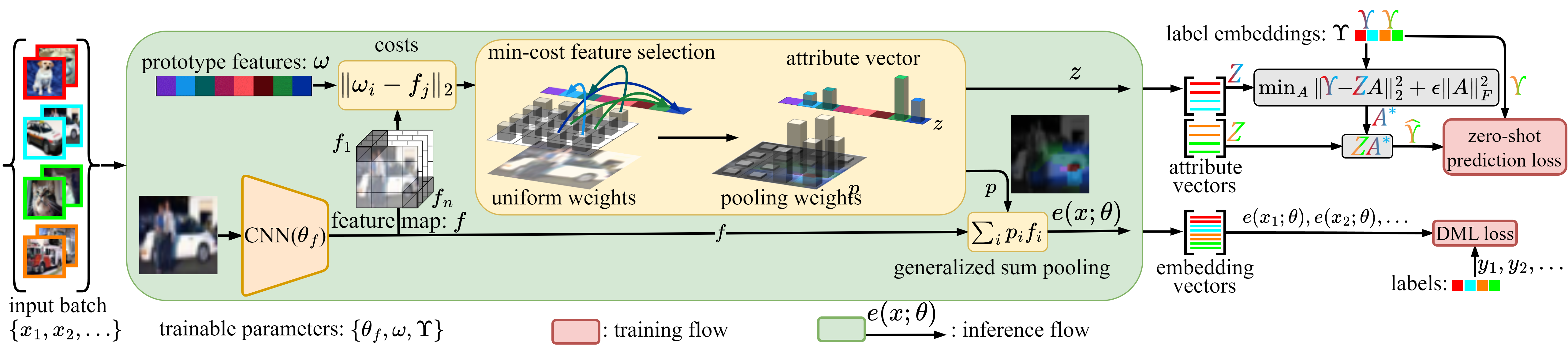}}
  \caption{Sketch of the method, where $Z{=}[z_i]_i$  \eqref{eq:attr_weights} and GSP vectors \eqref{eq:pooling_weights} are coloured w.r.t. their class label.}
	\label{fig:method}
  \end{figure*}

%% file: paper/figures/fig_smoothing.tex
\begin{wrapfigure}[15]{r}[-5pt]{0.41\linewidth}
\vspace{-.4\intextsep}
  \centerline{\includegraphics[width=1.0\linewidth,keepaspectratio]{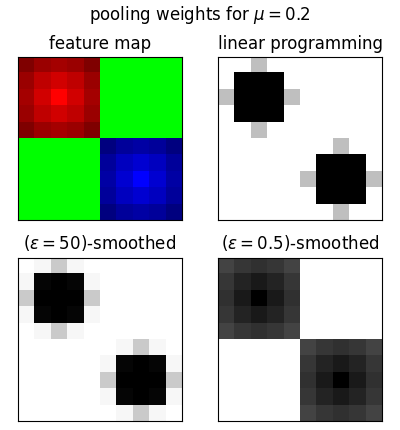}}
\caption{The resultant pooling  weights (higher the darker) of different problems.}
\label{fig:smooth_effect}
\end{wrapfigure}


%% file: paper/figures/fig_ablation_synthetic.tex
\begin{figure*}[t]
  \centering
  \centerline{\includegraphics[width=1\linewidth,keepaspectratio]{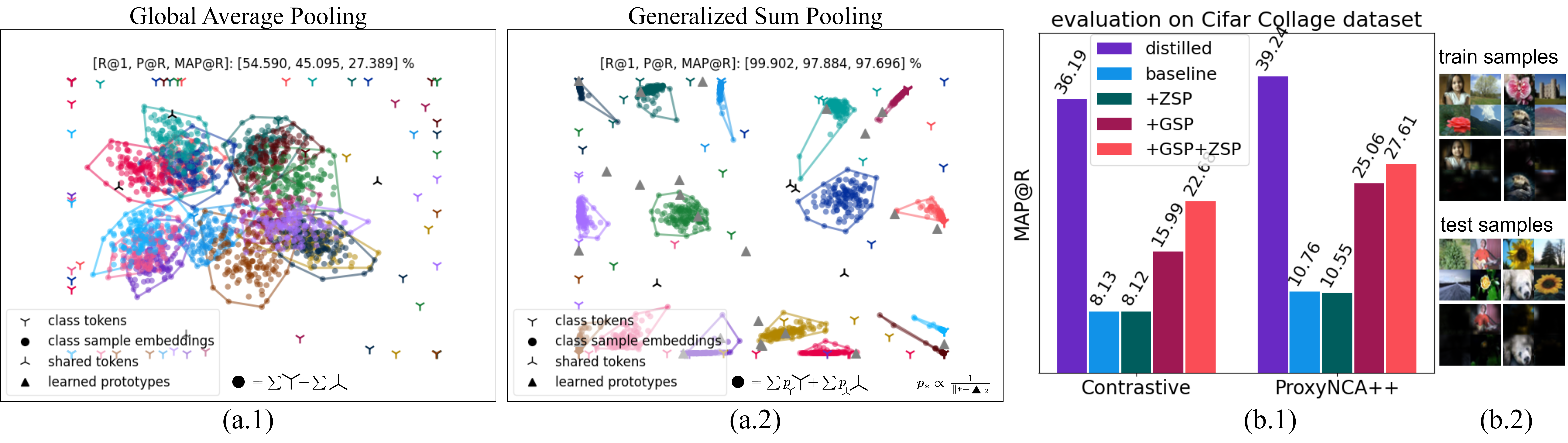}}
  \caption{Behavior analysis of GSP: (a) GAP vs GSP in aggregating features. The tokens represent learned embedding vectors, and the samples are derived from their aggregation. (b) Experiments on Cifar Collage dataset: (b.1) presents the results, while (b.2) displays sample train and test images along with their attention maps in terms of pooling weights. \textit{Distilled} denotes baseline performance on non-collage dataset, excluding the shared classes.}
	\label{fig:synthetic}
  \end{figure*}


%% file: supplementary/figures/fig_cifar_collage.tex
\begin{figure}[H]
\centering
\centerline{\includegraphics[width=.95\linewidth,keepaspectratio]{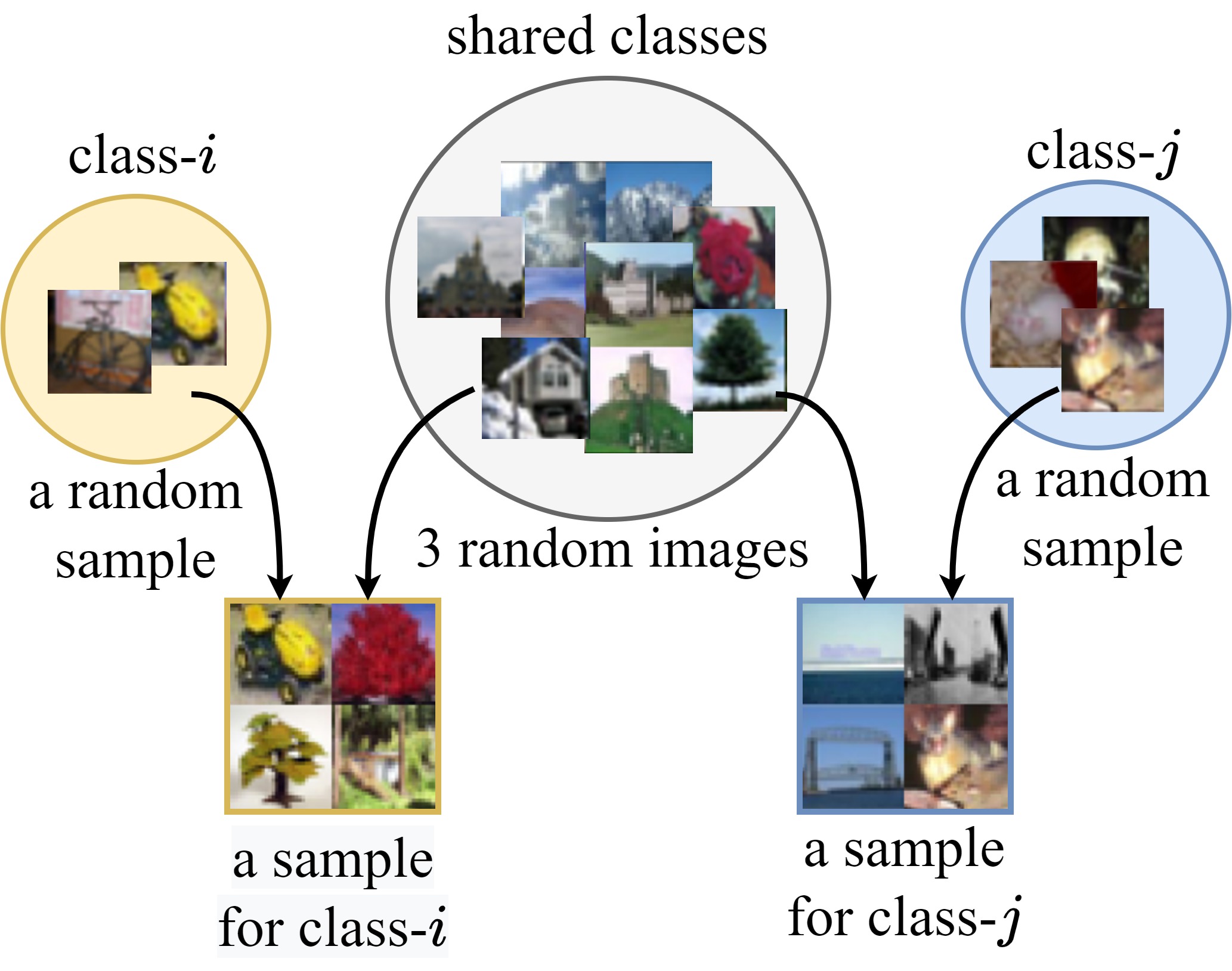}}
\caption{Sample generation for Cifar Collage dataset} 
\label{fig:cifarcollage}
\end{figure}

%% file: paper/figures/fig_ablation_zsl_attr.tex
%
  \begin{figure}[H]
\centerline{\includegraphics[width=.98\linewidth,keepaspectratio]{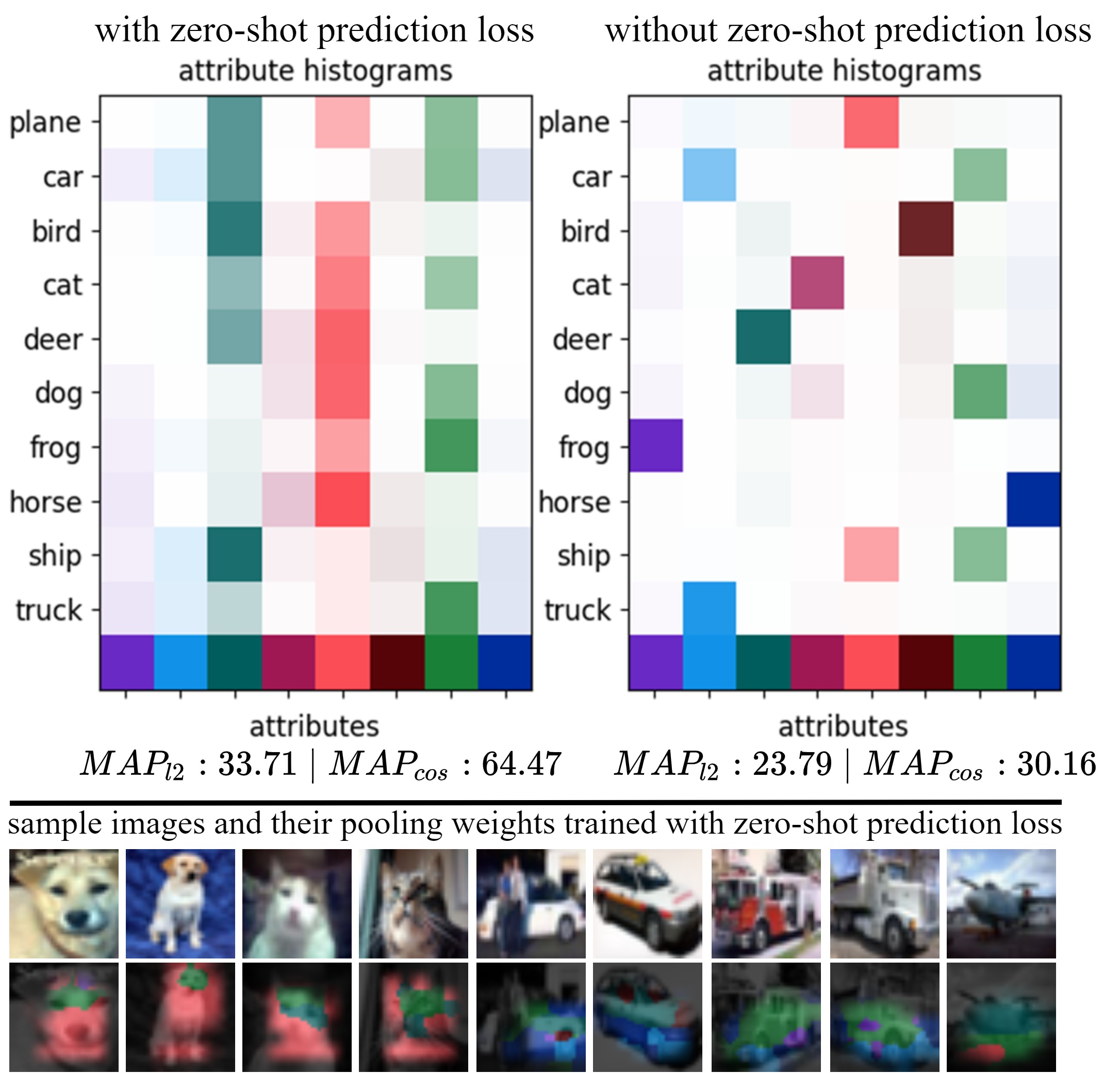}}
\caption{Comparing the distributions of the learned 8 prototypes across classes of Cifar10 dataset with and without $\mathcal{L}_\text{ZS}$. Pooling weights are coloured according to the dominant prototype at that location.}
\label{fig:attributes_eg}
\end{figure}

%% file: paper/figures/fig_summary.tex
\begin{figure*}[ht]
\centering
  \centerline{\includegraphics[width=1.0\linewidth,keepaspectratio]{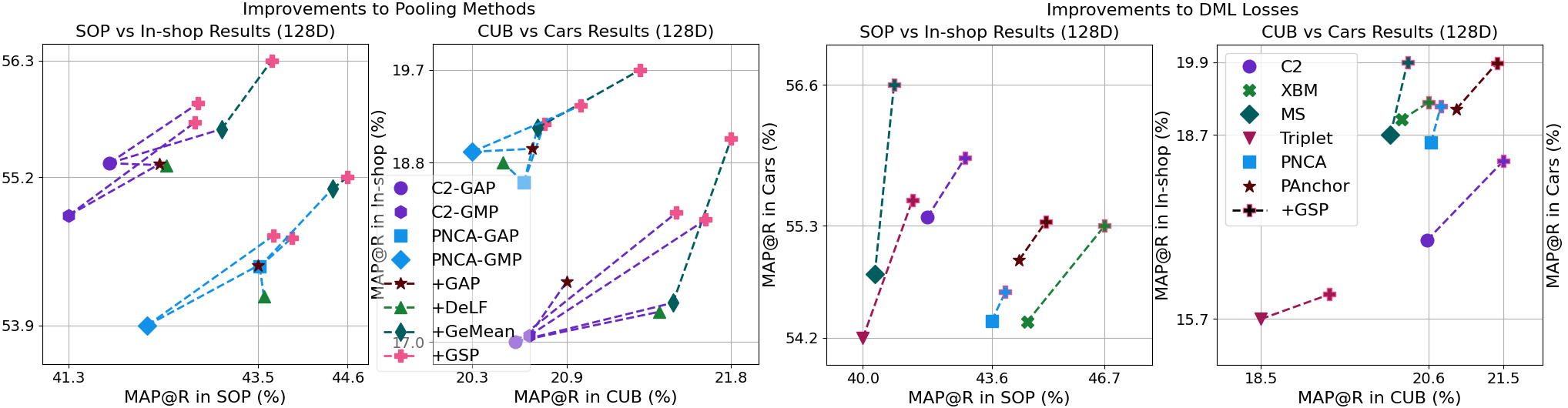}}
\caption{Summary of relative improvements: Each mark without a dashed line represents a baseline DML loss using GAP, unless stated otherwise. Similar losses are color-coded. Dashed lines indicate performance changes when replacing GAP with the associated pooling method (mark with the dashed line). Pooling methods applied on top of GMP or GMean are combined with them instead of replacing.}
\label{fig:test_summary_gsp}
\end{figure*}

%% file: paper/figures/table_components.tex
\begin{table}[b]
\centering
\caption{Effects of $\mathcal{L}_\text{ZS}$  and 
GSP with C2 loss}
\label{tab:components}
\resizebox{\linewidth}{!}{%
\begin{tabular}{@{}cccccccccc@{}}
\toprule
\multicolumn{2}{c}{} & \multicolumn{8}{c}{\textbf{MAP@R}}\\ \cmidrule(lr){3-10}
\multicolumn{2}{l}{\textbf{}}    & \multicolumn{2}{c}{\textbf{SOP}}                    & \multicolumn{2}{c}{\textbf{In-shop}}                & \multicolumn{2}{c}{\textbf{CUB}}                    & \multicolumn{2}{c}{\textbf{Cars196}} \\  \cmidrule(lr){3-4} \cmidrule(lr){5-6} \cmidrule(lr){7-8} \cmidrule(lr){9-10}
$\mathcal{L}_\text{ZS}$                  & GSP                 & 512D           & \multicolumn{1}{c}{128D}           & 512D           & \multicolumn{1}{c}{128D}           & 512D           & \multicolumn{1}{c}{128D}           & 512D              & 128D             \\ \midrule
                     &                     & 45.85          & \multicolumn{1}{c}{41.79}          & 59.07          & \multicolumn{1}{c}{55.38}          & 25.95          & \multicolumn{1}{c}{20.58}          & 24.38             & 17.02            \\
                     & \checkmark          & 46.78          & \multicolumn{1}{c}{42.66}          & 59.46          & \multicolumn{1}{c}{55.50}          & 26.25          & \multicolumn{1}{c}{20.85}          & 25.54             & 17.88            \\
\checkmark           &                     & 46.60          & \multicolumn{1}{c}{42.55}          & 59.38          & \multicolumn{1}{c}{55.43}          & 26.49          & \multicolumn{1}{c}{21.08}          & 25.54             & 17.67            \\
\checkmark           & \checkmark          & \textbf{46.81} & \multicolumn{1}{c}{\textbf{42.84}} & \textbf{60.01} & \multicolumn{1}{c}{\textbf{55.94}} & \textbf{27.12} & \multicolumn{1}{c}{\textbf{21.52}} & \textbf{26.25}    & \textbf{18.31}  \\ \bottomrule
\end{tabular}%
}
\end{table}

%% file: paper/figures/fig_ablation_duo.tex
\begin{figure}[H]
\centering
\begin{minipage}{.50\linewidth}

  \centerline{\includegraphics[width=1.0\linewidth,keepaspectratio]{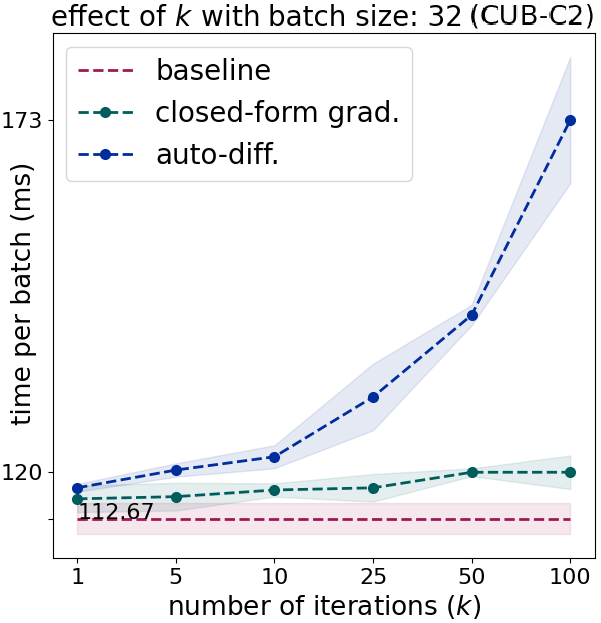}}

\end{minipage}
\begin{minipage}{.48\linewidth}

  \centerline{\includegraphics[width=1.0\linewidth,keepaspectratio]{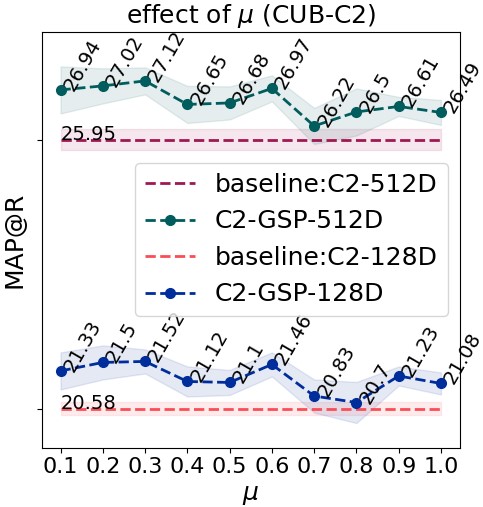}}

\end{minipage}

\caption{Efficiency of closed-form gradient (left) and effect of $\mu$ (right). Shaded regions represent $\mp$\textit{std}.}
\label{fig:ablation_duo}

\end{figure}

%% file: supplementary/figures/table_all_results.tex

\begin{table*}[!hb]
\renewcommand{\arraystretch}{1.5}
\centering
\caption{Comparison with the existing methods for the retrieval task on SOP, In-shop, CUB, Cars. Experimental setting follows MLRC \cite{musgrave2020metric}. $\mp$ denotes 1 \textit{std}. Red: the best, Blue: the second best, Bold: the loss term specific best.}
\label{tab:all_results}
\resizebox{\linewidth}{!}{%
\begin{tabular}{@{}l@{\hspace{2mm}}c@{\hspace{1mm}}c@{\hspace{1mm}}c@{\hspace{1mm}}c@{\hspace{4mm}}c@{\hspace{1mm}}c@{\hspace{1mm}}c@{\hspace{1mm}}c@{\hspace{4mm}}c@{\hspace{1mm}}c@{\hspace{1mm}}c@{\hspace{1mm}}c@{\hspace{4mm}}c@{\hspace{1mm}}c@{\hspace{1mm}}c@{\hspace{1mm}}c@{}}
\toprule
Dataset $\rightarrow$        & \multicolumn{4}{c}{\textbf{SOP}}& \multicolumn{4}{c}{\textbf{In-shop}}& \multicolumn{4}{c}{\textbf{CUB}}& \multicolumn{4}{c}{\textbf{Cars196}}\\ \cmidrule(lr){2-5} \cmidrule(lr){6-9} \cmidrule(lr){10-13} \cmidrule(lr){14-17}
Dim. $\rightarrow$           & \multicolumn{2}{c}{512D}& \multicolumn{2}{c}{128D}& \multicolumn{2}{c}{512D}& \multicolumn{2}{c}{128D}& \multicolumn{2}{c}{512D}& \multicolumn{2}{c}{128D}& \multicolumn{2}{c}{512D} & \multicolumn{2}{c}{128D}\\ 
\cmidrule(lr){2-3} \cmidrule(lr){4-5} \cmidrule(lr){6-7} \cmidrule(lr){8-9} \cmidrule(lr){10-11} \cmidrule(lr){12-13} \cmidrule(lr){14-15} \cmidrule(lr){16-17}
{Method}$\downarrow$  & \textbf{P@1}                                                                                     & \multicolumn{1}{c}{\textbf{MAP@R}}                                                                                   & \textbf{P@1}                                                                                     & \textbf{MAP@R}                                                                                   & \textbf{P@1}                                                                                     & \multicolumn{1}{c}{\textbf{MAP@R}}                                                                                   & \textbf{P@1}                                                                                     & \textbf{MAP@R}                                                                                   & \textbf{P@1}                                                                                     & \multicolumn{1}{c}{\textbf{MAP@R}}                                                                                   & \textbf{P@1}                                                                                     & \textbf{MAP@R}                                                                                   & \textbf{P@1}                                                                            & \multicolumn{1}{c}{\textbf{MAP@R}}                                                                                   & \textbf{P@1}                                                                                     & \textbf{MAP@R}                                                                                   \\ \midrule
C1+ & & & & & & & & & & & & & & & & \\
GAP                             & $\underset{\scriptsize{\mp0.11}}{69.29}$ & $\underset{\scriptsize{\mp0.15}}{40.40}$  & $\underset{\scriptsize{\mp0.10}}{65.15}$  & $\underset{\scriptsize{\mp0.11}}{36.50}$  & $\underset{\scriptsize{\mp0.19}}{80.11}$ & $\underset{\scriptsize{\mp0.14}}{50.32}$ & $\underset{\scriptsize{\mp0.14}}{75.83}$ & $\underset{\scriptsize{\mp0.13}}{46.42}$ & $\underset{\scriptsize{\mp0.57}}{63.32}$ & $\underset{\scriptsize{\mp0.31}}{23.49}$ & $\underset{\scriptsize{\mp0.35}}{56.34}$ & $\underset{\scriptsize{\mp0.29}}{19.37}$ & $\underset{\scriptsize{\mp0.38}}{78.01}$ & $\underset{\scriptsize{\mp0.33}}{22.87}$ & $\underset{\scriptsize{\mp0.59}}{65.61}$ & $\underset{\scriptsize{\mp0.14}}{16.06}$ \\
XBM+GAP                              & $\underset{\scriptsize{\mp0.32}}{76.54}$ & $\underset{\scriptsize{\mp0.47}}{48.58}$ & $\underset{\scriptsize{\mp0.48}}{73.22}$ & $\underset{\scriptsize{\mp0.57}}{44.55}$ & $\underset{\scriptsize{\mp0.26}}{87.76}$ & $\underset{\scriptsize{\mp0.41}}{57.53}$ & $\underset{\scriptsize{\mp0.37}}{85.26}$ & $\underset{\scriptsize{\mp0.45}}{54.40}$  & $\underset{\scriptsize{\mp0.48}}{65.56}$ & $\underset{\scriptsize{\mp0.24}}{25.65}$ & $\underset{\scriptsize{\mp0.41}}{57.48}$ & $\underset{\scriptsize{\mp0.19}}{20.27}$ & $\underset{\scriptsize{\mp0.35}}{83.55}$ & $\underset{\scriptsize{\mp0.22}}{27.53}$ & $\underset{\scriptsize{\mp0.30}}{72.17}$  & $\underset{\scriptsize{\mp0.17}}{18.98}$ \\
XBM+GSP                          & $\underset{\scriptsize{\mp0.18}}{{\color[HTML]{FE0000}\mathbf{77.88}}}$ & $\underset{\scriptsize{\mp0.28}}{{\color[HTML]{FE0000}\mathbf{50.65}}}$ & $\underset{\scriptsize{\mp0.19}}{{\color[HTML]{FE0000}\mathbf{74.84}}}$ & $\underset{\scriptsize{\mp0.28}}{{\color[HTML]{FE0000}\mathbf{46.69}}}$ & $\underset{\scriptsize{\mp0.19}}{{\color[HTML]{FE0000}\mathbf{88.33}}}$ & $\underset{\scriptsize{\mp0.29}}{\mathbf{58.55}}$ & $\underset{\scriptsize{\mp0.21}}{{\color[HTML]{00D2CB}\mathbf{85.95}}}$ & $\underset{\scriptsize{\mp0.21}}{\mathbf{55.30}}$  & $\underset{\scriptsize{\mp0.49}}{\mathbf{67.00}}$    & $\underset{\scriptsize{\mp0.15}}{\mathbf{26.05}}$ & $\underset{\scriptsize{\mp0.49}}{\mathbf{58.89}}$ & $\underset{\scriptsize{\mp0.16}}{\mathbf{20.60}}$  & $\underset{\scriptsize{\mp0.22}}{\mathbf{83.31}}$ & $\underset{\scriptsize{\mp0.23}}{\mathbf{27.88}}$ & $\underset{\scriptsize{\mp0.39}}{\mathbf{73.04}}$ & $\underset{\scriptsize{\mp0.19}}{\mathbf{19.26}}$ \\ \midrule
C2+ & & & & & & & & & & & & & & & & \\
GAP                             & $\underset{\scriptsize{\mp0.23}}{74.20}$  & $\underset{\scriptsize{\mp0.31}}{45.85}$ & $\underset{\scriptsize{\mp0.19}}{70.54}$ & $\underset{\scriptsize{\mp0.26}}{41.79}$ & $\underset{\scriptsize{\mp0.15}}{86.47}$ & $\underset{\scriptsize{\mp0.21}}{59.07}$ & $\underset{\scriptsize{\mp0.12}}{83.42}$ & $\underset{\scriptsize{\mp0.13}}{55.38}$ & $\underset{\scriptsize{\mp0.50}}{67.35}$  & $\underset{\scriptsize{\mp0.21}}{25.95}$ & $\underset{\scriptsize{\mp0.36}}{58.87}$ & $\underset{\scriptsize{\mp0.13}}{20.58}$ & $\underset{\scriptsize{\mp0.48}}{80.96}$ & $\underset{\scriptsize{\mp0.58}}{24.38}$ & $\underset{\scriptsize{\mp0.42}}{69.55}$ & $\underset{\scriptsize{\mp0.31}}{17.02}$ \\
GSP                              & $\underset{\scriptsize{\mp0.12}}{74.91}$ & $\underset{\scriptsize{\mp0.17}}{46.81}$ & $\underset{\scriptsize{\mp0.11}}{71.43}$ & $\underset{\scriptsize{\mp0.14}}{42.84}$ & $\underset{\scriptsize{\mp0.17}}{86.90}$  & $\underset{\scriptsize{\mp0.29}}{\mathbf{60.01}}$ & $\underset{\scriptsize{\mp0.18}}{83.57}$ & $\underset{\scriptsize{\mp0.17}}{55.94}$ & $\underset{\scriptsize{\mp0.41}}{{\color[HTML]{00D2CB}\mathbf{68.85}}}$  & $\underset{\scriptsize{\mp0.27}}{27.12}$  & $\underset{\scriptsize{\mp0.36}}{60.42}$ & $\underset{\scriptsize{\mp0.16}}{21.52}$ & $\underset{\scriptsize{\mp0.27}}{82.83}$ & $\underset{\scriptsize{\mp0.34}}{26.25}$ & $\underset{\scriptsize{\mp0.27}}{71.40}$  & $\underset{\scriptsize{\mp0.22}}{18.31}$ \\
DeLF                             & $\underset{\scriptsize{\mp0.15}}{74.59}$ & $\underset{\scriptsize{\mp0.19}}{46.54}$ & $\underset{\scriptsize{\mp0.18}}{45.53}$ & $\underset{\scriptsize{\mp0.17}}{42.47}$ & $\underset{\scriptsize{\mp0.16}}{86.65}$ & $\underset{\scriptsize{\mp0.22}}{59.20}$  & $\underset{\scriptsize{\mp0.09}}{83.51}$ & $\underset{\scriptsize{\mp0.12}}{55.36}$ & $\underset{\scriptsize{\mp0.32}}{68.66}$ & $\underset{\scriptsize{\mp0.18}}{27.06}$ & $\underset{\scriptsize{\mp0.18}}{59.85}$ & $\underset{\scriptsize{\mp0.16}}{21.42}$ & $\underset{\scriptsize{\mp0.41}}{81.85}$ & $\underset{\scriptsize{\mp0.38}}{24.77}$ & $\underset{\scriptsize{\mp0.38}}{69.95}$ & $\underset{\scriptsize{\mp0.25}}{17.32}$ \\
GeMean                          & $\underset{\scriptsize{\mp0.13}}{74.92}$ & $\underset{\scriptsize{\mp0.15}}{46.99}$ & $\underset{\scriptsize{\mp0.11}}{71.53}$ & $\underset{\scriptsize{\mp0.12}}{43.12}$ & $\underset{\scriptsize{\mp0.15}}{86.62}$ & $\underset{\scriptsize{\mp0.19}}{59.12}$ & $\underset{\scriptsize{\mp0.09}}{83.83}$ & $\underset{\scriptsize{\mp0.12}}{55.70}$  & $\underset{\scriptsize{\mp0.36}}{68.79}$ & $\underset{\scriptsize{\mp0.19}}{27.12}$ & $\underset{\scriptsize{\mp0.30}}{60.37}$  & $\underset{\scriptsize{\mp0.15}}{21.50}$  & $\underset{\scriptsize{\mp0.60}}{82.43}$  & $\underset{\scriptsize{\mp0.63}}{25.27}$ & $\underset{\scriptsize{\mp0.55}}{70.23}$ & $\underset{\scriptsize{\mp0.45}}{17.41}$ \\
GeMean+GSP                       & $\underset{\scriptsize{\mp0.08}}{\mathbf{75.32}}$ & $\underset{\scriptsize{\mp0.13}}{\mathbf{47.69}}$ & $\underset{\scriptsize{\mp0.10}}{\mathbf{71.93}}$  & $\underset{\scriptsize{\mp0.13}}{\mathbf{43.71}}$ & $\underset{\scriptsize{\mp0.15}}{\mathbf{86.94}}$ & $\underset{\scriptsize{\mp0.21}}{59.98}$ & $\underset{\scriptsize{\mp0.19}}{\mathbf{84.35}}$ & $\underset{\scriptsize{\mp0.14}}{{\color[HTML]{00D2CB}\mathbf{56.34}}}$ & $\underset{\scriptsize{\mp0.49}}{{\color[HTML]{FE0000}\mathbf{69.11}}}$ & $\underset{\scriptsize{\mp0.18}}{{\color[HTML]{FE0000}\mathbf{27.56}}}$ & $\underset{\scriptsize{\mp0.34}}{{\color[HTML]{FE0000}\mathbf{60.81}}}$ & $\underset{\scriptsize{\mp0.19}}{{\color[HTML]{FE0000}\mathbf{21.84}}}$ & $\underset{\scriptsize{\mp0.36}}{\mathbf{83.62}}$ & $\underset{\scriptsize{\mp0.31}}{\mathbf{26.98}}$ & $\underset{\scriptsize{\mp0.28}}{\mathbf{72.38}}$ & $\underset{\scriptsize{\mp0.22}}{\mathbf{19.05}}$ \\
GMP                              & $\underset{\scriptsize{\mp0.15}}{74.09}$ & $\underset{\scriptsize{\mp0.19}}{46.13}$ & $\underset{\scriptsize{\mp0.20}}{69.68}$  & $\underset{\scriptsize{\mp0.22}}{41.31}$ & $\underset{\scriptsize{\mp0.12}}{86.38}$ & $\underset{\scriptsize{\mp0.10}}{59.04}$  & $\underset{\scriptsize{\mp0.13}}{83.04}$ & $\underset{\scriptsize{\mp0.07}}{54.89}$ & $\underset{\scriptsize{\mp0.40}}{68.13}$  & $\underset{\scriptsize{\mp0.21}}{26.43}$ & $\underset{\scriptsize{\mp0.34}}{58.99}$ & $\underset{\scriptsize{\mp0.18}}{20.66}$ & $\underset{\scriptsize{\mp0.62}}{81.83}$ & $\underset{\scriptsize{\mp0.72}}{25.11}$ & $\underset{\scriptsize{\mp0.61}}{69.05}$ & $\underset{\scriptsize{\mp0.47}}{17.08}$ \\
GMP+GAP                          & $\underset{\scriptsize{\mp0.11}}{74.71}$ & $\underset{\scriptsize{\mp0.15}}{46.70}$  & $\underset{\scriptsize{\mp0.10}}{70.83}$  & $\underset{\scriptsize{\mp0.15}}{42.38}$ & $\underset{\scriptsize{\mp0.16}}{86.58}$ & $\underset{\scriptsize{\mp0.18}}{59.22}$ & $\underset{\scriptsize{\mp0.12}}{83.41}$ & $\underset{\scriptsize{\mp0.15}}{55.37}$ & $\underset{\scriptsize{\mp0.48}}{67.88}$ & $\underset{\scriptsize{\mp0.23}}{26.63}$ & $\underset{\scriptsize{\mp0.32}}{59.24}$ & $\underset{\scriptsize{\mp0.17}}{20.88}$ & $\underset{\scriptsize{\mp0.40}}{82.14}$  & $\underset{\scriptsize{\mp0.44}}{25.66}$ & $\underset{\scriptsize{\mp0.38}}{69.81}$ & $\underset{\scriptsize{\mp0.32}}{17.62}$ \\
GMP+GSP                          & $\underset{\scriptsize{\mp0.1}}{75.08}$  & $\underset{\scriptsize{\mp0.17}}{47.12}$ & $\underset{\scriptsize{\mp0.15}}{71.18}$ & $\underset{\scriptsize{\mp0.18}}{42.80}$  & $\underset{\scriptsize{\mp0.16}}{86.79}$ & $\underset{\scriptsize{\mp0.28}}{59.43}$ & $\underset{\scriptsize{\mp0.15}}{83.86}$ & $\underset{\scriptsize{\mp0.19}}{55.76}$ & $\underset{\scriptsize{\mp0.58}}{68.47}$ & $\underset{\scriptsize{\mp0.36}}{{\color[HTML]{00D2CB}\mathbf{27.49}}}$ & $\underset{\scriptsize{\mp0.41}}{60.19}$ & $\underset{\scriptsize{\mp0.35}}{{\color[HTML]{00D2CB}\mathbf{21.69}}}$ & $\underset{\scriptsize{\mp0.46}}{82.54}$ & $\underset{\scriptsize{\mp0.43}}{26.30}$  & $\underset{\scriptsize{\mp0.48}}{71.03}$ & $\underset{\scriptsize{\mp0.29}}{18.24}$ \\ \midrule
MS+ & & & & & & & & & & & & & & & & \\
GAP        & $\underset{\scriptsize{\mp0.14}}{72.81}$ & $\underset{\scriptsize{\mp0.21}}{44.19}$ & $\underset{\scriptsize{\mp0.10}}{69.09}$  & $\underset{\scriptsize{\mp0.16}}{40.34}$ & $\underset{\scriptsize{\mp0.20}}{87.01}$  & $\underset{\scriptsize{\mp0.37}}{58.79}$ & $\underset{\scriptsize{\mp0.21}}{83.87}$ & $\underset{\scriptsize{\mp0.34}}{54.85}$ & $\underset{\scriptsize{\mp0.46}}{65.43}$ & $\underset{\scriptsize{\mp0.15}}{24.95}$ & $\underset{\scriptsize{\mp0.27}}{\mathbf{57.57}}$ & $\underset{\scriptsize{\mp0.12}}{20.13}$ & $\underset{\scriptsize{\mp0.34}}{83.73}$ & $\underset{\scriptsize{\mp0.43}}{27.16}$ & $\underset{\scriptsize{\mp0.43}}{72.54}$ & $\underset{\scriptsize{\mp0.31}}{18.73}$ \\
GSP                              & $\underset{\scriptsize{\mp0.11}}{\mathbf{73.05}}$ & $\underset{\scriptsize{\mp0.17}}{\mathbf{44.72}}$ & $\underset{\scriptsize{\mp0.15}}{\mathbf{69.44}}$ & $\underset{\scriptsize{\mp0.19}}{\mathbf{40.87}}$ & $\underset{\scriptsize{\mp0.21}}{{\color[HTML]{00D2CB}\mathbf{88.28}}}$ & $\underset{\scriptsize{\mp0.24}}{{\color[HTML]{00D2CB}\mathbf{60.49}}}$ & $\underset{\scriptsize{\mp0.19}}{\mathbf{85.28}}$ & $\underset{\scriptsize{\mp0.26}}{{\color[HTML]{FE0000}\mathbf{56.62}}}$ & $\underset{\scriptsize{\mp0.33}}{\mathbf{65.50}}$  & $\underset{\scriptsize{\mp0.21}}{\mathbf{25.09}}$ & $\underset{\scriptsize{\mp0.15}}{57.39}$ & $\underset{\scriptsize{\mp0.22}}{\mathbf{20.34}}$ & $\underset{\scriptsize{\mp0.35}}{{\color[HTML]{00D2CB}\mathbf{84.27}}}$ & $\underset{\scriptsize{\mp0.40}}{{\color[HTML]{FE0000}\mathbf{28.58}}}$  & $\underset{\scriptsize{\mp0.32}}{{\color[HTML]{00D2CB}\mathbf{73.74}}}$ & $\underset{\scriptsize{\mp0.31}}{{\color[HTML]{FE0000}\mathbf{19.91}}}$ \\ \midrule
Triplet+ & & & & & & & & & & & & & & & & \\
GAP                           & $\underset{\scriptsize{\mp0.24}}{74.54}$ & $\underset{\scriptsize{\mp0.30}}{45.88}$  & $\underset{\scriptsize{\mp0.38}}{69.41}$ & $\underset{\scriptsize{\mp0.39}}{40.01}$ & $\underset{\scriptsize{\mp0.36}}{85.99}$ & $\underset{\scriptsize{\mp0.46}}{59.67}$ & $\underset{\scriptsize{\mp0.38}}{81.75}$ & $\underset{\scriptsize{\mp0.45}}{54.25}$ & $\underset{\scriptsize{\mp0.66}}{64.11}$ & $\underset{\scriptsize{\mp0.40}}{23.65}$  & $\underset{\scriptsize{\mp0.46}}{55.62}$ & $\underset{\scriptsize{\mp0.31}}{18.54}$ & $\underset{\scriptsize{\mp0.60}}{77.58}$  & $\underset{\scriptsize{\mp0.58}}{22.67}$ & $\underset{\scriptsize{\mp0.59}}{64.61}$ & $\underset{\scriptsize{\mp0.34}}{15.74}$ \\
GSP                         & $\underset{\scriptsize{\mp0.23}}{\mathbf{75.59}}$ & $\underset{\scriptsize{\mp0.32}}{\mathbf{47.35}}$ & $\underset{\scriptsize{\mp0.20}}{\mathbf{70.65}}$  & $\underset{\scriptsize{\mp0.22}}{\mathbf{41.38}}$ & $\underset{\scriptsize{\mp0.27}}{\mathbf{86.75}}$ & $\underset{\scriptsize{\mp0.47}}{{\color[HTML]{FE0000}\mathbf{60.85}}}$ & $\underset{\scriptsize{\mp0.33}}{\mathbf{82.74}}$ & $\underset{\scriptsize{\mp0.46}}{\mathbf{55.54}}$ & $\underset{\scriptsize{\mp0.52}}{\mathbf{66.09}}$ & $\underset{\scriptsize{\mp0.33}}{\mathbf{24.80}}$  & $\underset{\scriptsize{\mp0.42}}{\mathbf{57.12}}$ & $\underset{\scriptsize{\mp0.25}}{\mathbf{19.38}}$ & $\underset{\scriptsize{\mp0.30}}{\mathbf{78.93}}$  & $\underset{\scriptsize{\mp0.29}}{\mathbf{23.44}}$ & $\underset{\scriptsize{\mp0.35}}{\mathbf{65.81}}$ & $\underset{\scriptsize{\mp0.21}}{\mathbf{16.14}}$ \\ \midrule
PNCA+ & & & & & & & & & & & & & & & & \\
GAP        & $\underset{\scriptsize{\mp0.15}}{75.18}$ & $\underset{\scriptsize{\mp0.16}}{47.11}$ & $\underset{\scriptsize{\mp0.06}}{72.15}$ & $\underset{\scriptsize{\mp0.08}}{43.57}$ & $\underset{\scriptsize{\mp0.14}}{87.26}$ & $\underset{\scriptsize{\mp0.14}}{57.43}$ & $\underset{\scriptsize{\mp0.08}}{84.86}$ & $\underset{\scriptsize{\mp0.10}}{54.41}$ & $\underset{\scriptsize{\mp0.51}}{65.74}$ & $\underset{\scriptsize{\mp0.23}}{25.27}$ & $\underset{\scriptsize{\mp0.36}}{58.19}$ & $\underset{\scriptsize{\mp0.20}}{20.63}$  & $\underset{\scriptsize{\mp0.25}}{82.33}$ & $\underset{\scriptsize{\mp0.22}}{26.21}$ & $\underset{\scriptsize{\mp0.18}}{70.75}$ & $\underset{\scriptsize{\mp0.08}}{18.61}$ \\
GSP                            & $\underset{\scriptsize{\mp0.11}}{75.68}$ & $\underset{\scriptsize{\mp0.14}}{47.74}$ & $\underset{\scriptsize{\mp0.06}}{72.37}$ & $\underset{\scriptsize{\mp0.06}}{43.95}$ & $\underset{\scriptsize{\mp0.10}}{87.35}$  & $\underset{\scriptsize{\mp0.12}}{57.65}$ & $\underset{\scriptsize{\mp0.10}}{85.13}$  & $\underset{\scriptsize{\mp0.08}}{54.68}$ & $\underset{\scriptsize{\mp0.38}}{65.80}$  & $\underset{\scriptsize{\mp0.25}}{25.48}$ & $\underset{\scriptsize{\mp0.22}}{58.20}$  & $\underset{\scriptsize{\mp0.19}}{20.75}$ & $\underset{\scriptsize{\mp0.27}}{82.70}$  & $\underset{\scriptsize{\mp0.18}}{26.93}$ & $\underset{\scriptsize{\mp0.32}}{71.55}$ & $\underset{\scriptsize{\mp0.17}}{19.20}$  \\
DeLF                         & $\underset{\scriptsize{\mp0.09}}{75.29}$ & $\underset{\scriptsize{\mp0.11}}{47.44}$ & $\underset{\scriptsize{\mp0.06}}{72.05}$ & $\underset{\scriptsize{\mp0.07}}{43.62}$ & $\underset{\scriptsize{\mp0.11}}{87.19}$ & $\underset{\scriptsize{\mp0.16}}{57.44}$ & $\underset{\scriptsize{\mp0.04}}{84.55}$ & $\underset{\scriptsize{\mp0.10}}{54.13}$  & $\underset{\scriptsize{\mp0.34}}{65.42}$ & $\underset{\scriptsize{\mp0.16}}{25.31}$ & $\underset{\scriptsize{\mp0.24}}{57.98}$ & $\underset{\scriptsize{\mp0.14}}{20.51}$ & $\underset{\scriptsize{\mp0.35}}{82.37}$ & $\underset{\scriptsize{\mp0.22}}{26.63}$ & $\underset{\scriptsize{\mp0.27}}{71.06}$ & $\underset{\scriptsize{\mp0.14}}{18.81}$ \\
GeMean                       & $\underset{\scriptsize{\mp0.09}}{75.64}$ & $\underset{\scriptsize{\mp0.09}}{47.82}$ & $\underset{\scriptsize{\mp0.07}}{72.75}$ & $\underset{\scriptsize{\mp0.06}}{44.43}$ & $\underset{\scriptsize{\mp0.10}}{87.63}$  & $\underset{\scriptsize{\mp0.13}}{57.88}$ & $\underset{\scriptsize{\mp0.14}}{85.48}$ & $\underset{\scriptsize{\mp0.12}}{55.14}$ & $\underset{\scriptsize{\mp0.33}}{66.33}$ & $\underset{\scriptsize{\mp0.20}}{25.74}$  & $\underset{\scriptsize{\mp0.39}}{58.52}$ & $\underset{\scriptsize{\mp0.20}}{20.71}$  & $\underset{\scriptsize{\mp0.29}}{\mathbf{83.83}}$ & $\underset{\scriptsize{\mp0.15}}{27.44}$ & $\underset{\scriptsize{\mp0.28}}{\mathbf{72.14}}$ & $\underset{\scriptsize{\mp0.12}}{19.16}$ \\
GeMean+GSP                     & $\underset{\scriptsize{\mp0.11}}{\mathbf{75.89}}$ & $\underset{\scriptsize{\mp0.12}}{\mathbf{48.17}}$ & $\underset{\scriptsize{\mp0.04}}{\mathbf{72.91}}$ & $\underset{\scriptsize{\mp0.06}}{\mathbf{44.61}}$ & $\underset{\scriptsize{\mp0.10}}{\mathbf{87.64}}$  & $\underset{\scriptsize{\mp0.16}}{\mathbf{58.12}}$ & $\underset{\scriptsize{\mp0.07}}{\mathbf{85.58}}$ & $\underset{\scriptsize{\mp0.08}}{\mathbf{55.25}}$ & $\underset{\scriptsize{\mp0.53}}{\mathbf{67.39}}$ & $\underset{\scriptsize{\mp0.26}}{\mathbf{26.19}}$ & $\underset{\scriptsize{\mp0.40}}{\mathbf{59.39}}$  & $\underset{\scriptsize{\mp0.21}}{\mathbf{21.31}}$ & $\underset{\scriptsize{\mp0.25}}{83.09}$ & $\underset{\scriptsize{\mp0.30}}{\mathbf{27.96}}$  & $\underset{\scriptsize{\mp0.27}}{71.95}$ & $\underset{\scriptsize{\mp0.19}}{\mathbf{19.74}}$ \\
GMP                            & $\underset{\scriptsize{\mp0.08}}{74.43}$ & $\underset{\scriptsize{\mp0.08}}{46.33}$ & $\underset{\scriptsize{\mp0.07}}{70.80}$  & $\underset{\scriptsize{\mp0.08}}{42.24}$ & $\underset{\scriptsize{\mp0.13}}{86.94}$ & $\underset{\scriptsize{\mp0.13}}{56.79}$ & $\underset{\scriptsize{\mp0.08}}{84.53}$ & $\underset{\scriptsize{\mp0.09}}{53.86}$ & $\underset{\scriptsize{\mp0.51}}{65.74}$ & $\underset{\scriptsize{\mp0.29}}{25.36}$ & $\underset{\scriptsize{\mp0.38}}{57.61}$ & $\underset{\scriptsize{\mp0.29}}{20.33}$ & $\underset{\scriptsize{\mp0.33}}{83.06}$ & $\underset{\scriptsize{\mp0.27}}{26.96}$ & $\underset{\scriptsize{\mp0.25}}{71.19}$ & $\underset{\scriptsize{\mp0.15}}{18.92}$ \\
GMP+GAP                        & $\underset{\scriptsize{\mp0.09}}{75.19}$ & $\underset{\scriptsize{\mp0.11}}{47.26}$ & $\underset{\scriptsize{\mp0.04}}{71.97}$ & $\underset{\scriptsize{\mp0.06}}{43.55}$ & $\underset{\scriptsize{\mp0.14}}{87.21}$ & $\underset{\scriptsize{\mp0.15}}{57.34}$ & $\underset{\scriptsize{\mp0.09}}{84.95}$ & $\underset{\scriptsize{\mp0.10}}{54.42}$  & $\underset{\scriptsize{\mp0.35}}{65.91}$ & $\underset{\scriptsize{\mp0.26}}{25.56}$ & $\underset{\scriptsize{\mp0.37}}{57.92}$ & $\underset{\scriptsize{\mp0.20}}{20.68}$  & $\underset{\scriptsize{\mp0.41}}{82.92}$ & $\underset{\scriptsize{\mp0.36}}{26.92}$ & $\underset{\scriptsize{\mp0.22}}{71.33}$ & $\underset{\scriptsize{\mp0.19}}{18.95}$ \\
GMP+GSP                        & $\underset{\scriptsize{\mp0.12}}{75.41}$ & $\underset{\scriptsize{\mp0.12}}{47.50}$  & $\underset{\scriptsize{\mp0.07}}{72.10}$  & $\underset{\scriptsize{\mp0.09}}{43.73}$ & $\underset{\scriptsize{\mp0.10}}{87.43}$  & $\underset{\scriptsize{\mp0.14}}{57.68}$ & $\underset{\scriptsize{\mp0.10}}{85.10}$   & $\underset{\scriptsize{\mp0.08}}{54.70}$  & $\underset{\scriptsize{\mp0.48}}{66.14}$ & $\underset{\scriptsize{\mp0.23}}{25.85}$ & $\underset{\scriptsize{\mp0.32}}{58.12}$ & $\underset{\scriptsize{\mp0.18}}{20.96}$ & $\underset{\scriptsize{\mp0.31}}{83.46}$ & $\underset{\scriptsize{\mp0.21}}{27.12}$ & $\underset{\scriptsize{\mp0.39}}{72.04}$ & $\underset{\scriptsize{\mp0.20}}{19.38}$  \\ \midrule
PAnchor+ & & & & & & & & & & & & & & & & \\
GAP                              & $\underset{\scriptsize{\mp0.19}}{76.48}$ & $\underset{\scriptsize{\mp0.26}}{48.08}$ & $\underset{\scriptsize{\mp0.14}}{73.50}$  & $\underset{\scriptsize{\mp0.20}}{44.33}$  & $\underset{\scriptsize{\mp0.21}}{88.02}$ & $\underset{\scriptsize{\mp0.25}}{58.02}$ & $\underset{\scriptsize{\mp0.18}}{85.83}$ & $\underset{\scriptsize{\mp0.22}}{54.98}$ & $\underset{\scriptsize{\mp0.41}}{68.04}$ & $\underset{\scriptsize{\mp0.21}}{26.20}$  & $\underset{\scriptsize{\mp0.34}}{59.91}$ & $\underset{\scriptsize{\mp0.15}}{20.94}$ & $\underset{\scriptsize{\mp0.31}}{85.26}$ & $\underset{\scriptsize{\mp0.20}}{27.14}$  & $\underset{\scriptsize{\mp0.23}}{75.08}$ & $\underset{\scriptsize{\mp0.13}}{19.15}$ \\
GSP                         & $\underset{\scriptsize{\mp0.16}}{{\color[HTML]{00D2CB}\mathbf{77.13}}}$ & $\underset{\scriptsize{\mp0.22}}{{\color[HTML]{00D2CB}\mathbf{49.05}}}$ & $\underset{\scriptsize{\mp0.13}}{{\color[HTML]{00D2CB}\mathbf{74.07}}}$ & $\underset{\scriptsize{\mp0.17}}{{\color[HTML]{00D2CB}\mathbf{45.07}}}$ & $\underset{\scriptsize{\mp0.11}}{\mathbf{88.10}}$  & $\underset{\scriptsize{\mp0.14}}{\mathbf{58.44}}$ & $\underset{\scriptsize{\mp0.06}}{{\color[HTML]{FE0000}\mathbf{85.97}}}$ & $\underset{\scriptsize{\mp0.13}}{\mathbf{55.34}}$ & $\underset{\scriptsize{\mp0.45}}{\mathbf{68.40}}$  & $\underset{\scriptsize{\mp0.25}}{\mathbf{26.59}}$ & $\underset{\scriptsize{\mp0.31}}{{\color[HTML]{00D2CB}\mathbf{60.80}}}$  & $\underset{\scriptsize{\mp0.17}}{\mathbf{21.44}}$ & $\underset{\scriptsize{\mp0.39}}{{\color[HTML]{FE0000}\mathbf{86.46}}}$ & $\underset{\scriptsize{\mp0.33}}{{\color[HTML]{00D2CB}\mathbf{28.43}}}$ & $\underset{\scriptsize{\mp0.25}}{{\color[HTML]{FE0000}\mathbf{75.88}}}$ & $\underset{\scriptsize{\mp0.20}}{{\color[HTML]{00D2CB}\mathbf{19.90}}}$   \\ \bottomrule
\end{tabular}%
}
\end{table*}

%% file: supplementary/supplementary_append.tex




\setcounter{section}{0}
\twocolumn[{\section*{\Large \centering Supplemental Material for {\large\emph{"Generalized Sum Pooling for Metric Learning"}}}}]

\input{supplementary/experiments}

\appendix
\input{supplementary/appendix}

%% file: supplementary/experiments.tex
\section{Extended Empirical Study for DML}
In the following sections, we explain our empirical study in detail and provide additional experiments on effect of hyperparameters as well as evaluation with the \textit{conventional} experimental settings.

\subsection*{Reproducibility}
\hypertarget{code}{We} provide full detail of our experimental setup and recapitulate the implementation details for the sake of complete transparency and reproducibility. Code is available at: \href{\codeurl}{GSP-DML Framework}.
%
%

\input{supplementary/figures/table_conventional}


\subsection{Conventional Evaluation}
We additionally follow the relatively old-fashioned conventional procedure \cite{oh2016deep} for the evaluation of our method. We use BN-Inception \cite{normalization2015accelerating} and ResNet50 \cite{he2016identity} architectures as the backbones. We obtain 512D (BN-Inception and ResNet50) embeddings through linear transformation after global pooling layer. Aligned with the recent approaches \cite{venkataramanan2022it,teh2020proxynca++,kim2020proxy, wang2020cross}, we use global max pooling as well as global average pooling. The rest of the settings are disclosed in \cref{sec:experiment_details}.

We evaluate our method with XBM. We provide R@1 results in \cref{tab:conventional} for the comparison with SOTA. In our evaluations, we also provide MAP@R scores in parenthesis under R@1 scores. We also provide baseline XBM evaluation in our framework. The results are mostly consistent with the ones reported in the original paper \cite{wang2020cross} except for CUB and Cars datasets. In XBM \cite{wang2020cross}, the authors use proxy-based trainable memory for CUB and Cars datasets. On the other hand, we use the official implementation provided by the authors, which does not include such proxy-based extensions. 

We observe that our method improves XBM and XBM+GSP reaches SOTA performance in large scale datasets. With that being said, the improvement margins are less substantial than the ones in \textit{fair evaluation}. Such a result is expected since training is terminated by \textit{early-stopping} which is a common practice to regularize the generalization of training \cite{dong2020generalization, lei2021generalization}. In \textit{conventional evaluation}, early-stopping is achieved by monitoring the test data performance, enabling good generalization to test data. Therefore, observing less improvement in generalization with GSP is something we expect owing to generalization boost that test data based early-stopping already provides.

We also observe that in a few cases, the R@1 performance of GSP is slightly worse than the baseline. Nevertheless, once we compare the MAP@R performances, GSP consistently brings improvement with no exception. We should recapitulate that R@1 is a myopic metric to assess the quality of the embedding space geometry \cite{musgrave2020metric} and hence, pushing R@1 does not necessarily reflect the true order of the improvements that the methods bring.

As we observe from MAP@R comparisons in Table~2 (main paper), the methods sharing similar R@1 (\ie, P@1) performances can differ in MAP@R performance relatively more significantly. In that manner, we firmly believe that comparing MAP@R performances instead of R@1 technically sounds more in showing the improvements of our method.

Finally, we also apply our method with LIBC \cite{intrabatch} to further show wide applicability of our method. We use the official implementation of LIBC and follow their default experimental settings. The evaluations on 4 benchmarks show that GSP improve LIBC by $\approx0.5$pp R@1. To offer a complete outlook on the conventional evaluation, we have included the recall at K (R@K) scores in \Cref{tab:r_at_k} as well.

\input{supplementary/figures/table_r_at_k}

\subsection{Application of GSP to Other Problems}
GSP is applicable to any problem and architecture with a pooling layer. We believe GSP is particularly relevant to the problem of metric learning due to the geometry it enforces. Our pooling method enhances local geometry by reducing overlap of class convex hulls and improves unseen class generalization. 

In order to evaluate the applicability of GSP beyond metric learning, we applied GSP to ImageNet classification tasks using ResNetV2 \cite{he2016identity} and EfficientNetV2 \cite{tan2021efficientnetv2} models. We took the official \href{https://www.tensorflow.org/api_docs/python/tf/keras/applications}{Tensorflow Keras models} and only replaced pooling layers with GSP.

\input{supplementary/figures/table_imagenet}

The results suggests that our method is applicable beyond metric lerning as it improves ImageNet classification accuracy for both ResNetV2 and EfficientNetV2 models. We additionally assessed the metric learning performance of the embedding vectors pre-classification. By reducing the embedding dimension to 512 through LDA, we then evaluated the resulting embedding geometry using P@1 and MAP@R metrics, and observed that GSP yields better feature geometry.

\input{supplementary/figures/table_pooling}
\subsection{Evaluation of Other Pooling Alternatives}

We evaluate 14 additional pooling alternatives on \textit{Ciffar Collage} and CUB datasets with \textit{contrastive} (C2) and \textit{Proxy-NCA++} (PNCA) losses. We pick \textit{contrastive}  since  it is one of the best performing sample-based loss. We pick \textit{Proxy-NCA++} since most of the pooling methods are tailored for landmark-based image retrieval and use classification loss akin to \textit{Proxy-NCA++}. We particularly consider \textit{Cifar Collage} dataset since the images of different classes share a considerable amount of semantic entities, enabling us to assess the methods w.r.t. their ability to discard the nuisance information.

\textbf{Compared methods.} In addition to our method (GSP) and global average pooling (GAP), we consider: $i)$ global max pooling (GMP), $ii)$ GAP+GMP \cite{kim2020proxy}, $iii)$ CBAM \cite{cbam}, $iv)$ CroW \cite{crow}, $v)$ DeLF \cite{delf}, $vi)$ generalized max pooling (GeMax) \cite{murray2016interferences}, $vii)$ generalized mean pooling (GeMean) \cite{gmeanp}, $viii)$ GSoP \cite{gsop}, $ix)$ optimal transport based aggregation (OTP) \cite{mialon2021trainable,kolouri2020wasserstein}, $x)$ SOLAR \cite{solar}, $xi)$ trainable SMK (T-SMK) \cite{tamsk}, $xii)$ NetVLAD \cite{vlad}, $xiii)$ WELDON \cite{weldon}, and $xiv)$ visual transformer encoder with class token (TFM) \cite{dosovitskiy2021image}. Among those, OTP and VLAD are ensemble based methods and typically necessitate large embedding dimensions. Thus, we both experimented their 128D version -($8\sxtimes 16)$ (8 prototypes of 16D vectors) and 8192D version -($64\sxtimes 128)$ (64 prototypes of 128D vectors). We note that some of our baselines utilize attention based pooling. Notably, attention mechanism is the key part of DeLF, SOLAR, and GSoP. In fact, DeLF can be seen as equivalent to a single-layer residual-free transformer layer with a class token. To perform a more direct comparison with transformers, we conducted experiments by replacing GAP with transformer layers using a class token. We evaluated 1, 2, 4, and 8-layer transformers (TFM-\#).


\textbf{Setting.} For CUB dataset, the experimental setting follows \cref{sec:experiment_details}-\textit{Fair evaluation} and we report MAP@R performance of the 4-model average at 128 dimensional embeddings each. For \textit{Cifar Collage dataset}, the experimental setting follows \cref{sec:cifar_collage_details} and we report MAP@R performance. We provide the results in \Cref{tab:pooling}.

\textbf{Results.} Evaluations show that our method is superior to other pooling alternatives including prototype based VLAD and OTP. Predominantly, for 128 dimensional embeddings, our method outperforms prototype based methods by large margin. In CUB dataset, the pooling methods either are inferior to or perform on par with GAP. The performance improvements of the superior methods are less than 1\%, implying that our improvements in the order of 1-2\% reported in Table~2 (main paper) is substantial. Besides, the methods that mask the feature map outperform GAP by large margin in \textit{Cifar Collage} dataset. That said, our method outperforms all the methods except for Contrastive+VLAD by large margin in \textit{Cifar Collage} dataset, yet another evidence for better feature selection mechanism of our method. For instance in CUB dataset, DeLF and GeMean are on par with our method which has slightly better performance. Yet, our method outperforms both by large margin in \textit{Cifar Collage} dataset.

\textbf{Superior selection mechanism.} Comparing to CroW, T-SMK and CBAM, our method outperforms them by large margin. Those methods are the built on feature magnitude based saliency, assuming that the backbone functions must be able to zero-out nuisance information. Yet, such a requirement is restrictive for the parameter space and annihilation of the non-discriminative information might not be feasible in some problems. We experimentally observe such a weakness of CroW, T-SMK and CBAM in \textit{Cifar Collage} dataset where the nuisance information cannot be zeroed-out by the backbone. Our formulation do not have such a restrictive assumption and thus substantially superior to them.

\textbf{Superior attention mechanism.} Similarly, attention-based weighting methods, DeLF and GSoP, do not have explicit control on feature selection behavior and might result in poor models when jointly trained with the feature extractor \cite{delf}, which we also observe in \textit{Cifar Collage} experiments. On the contrary, we have explicit control on the pooling behavior with $\mu$ parameter and the behavior of our method is stable and consistent across datasets and with different loss functions. We also found that our method outperforms direct application of transformer based pooling. 

\textbf{Simpler and interpretable.} Attention-based methods DeLF, GSoP, and SOLAR typically introduce several convolution layers to compute the feature weights. We only introduce an $m$-kernel $1\sxtimes1$ convolution layer (\ie, $m$-many trainable prototypes) and obtain better results. We should note that our pooling operation is as simple as performing a convolution (\ie, distance computation) and alternating normalization of a vector and a scalar. Additionally, we are able to incorporate a zero-shot regularization into our problem naturally by using the prototype assignment weights. We can as well incorporate such a loss in DeLF which has $1\sxtimes1$ convolution to compute prototype similarities. However, we first need a mechanism to aggregate the per prototype similarities (\eg, sum and normalization). Indeed, normalizing the similarities channel-wise and spatially summing them correspond to solving our problem with $\mu=1$.

Other pooling methods, \ie, GAP, GMP, GAP+GMP, GeMax, GeMean, WELDON, VLAD, OTP, do not build on discriminative feature selection. Thus, our method substantially outperforms those.

\input{supplementary/figures/fig_inference_computation}

\subsection{Computational Analysis}

Forward and backward computation of proposed GSP method can be implemented using only matrix-vector products. Moreover, having closed-form matrix-inversion-free expression for the loss gradient enables memory efficient back propagation since the output of each iteration must be stored otherwise.

We perform $k$ iterations to obtain the pooling weights and at each iteration, we only perform matrix-vector products. In this sense, the back propagation can be achieved using \textit{automatic-differentiation}. One problem with automatic differentiation is that the computation load increases with increasing $k$. On the other hand, with closed-form gradient expression, we do not have such issue and in fact we have constant back propagation complexity. Granted that the closed-form expression demands exact solution of the problem (\ie, $k\to\infty$), forward computation puts a little computation overhead and is memory efficient since we discard the intermediate outputs. Moreover, our initial empirical study show that our problems typically converges for $k>50$ and we observe similar performances with $k\geqslant 25$. 

\input{supplementary/figures/fig_computation}

The choice of $k$ is indeed problem dependent (\ie, size of the feature map and the number of prototypes). Thus, its effect on computation load should be analyzed. We study the effect of $k$ with automatic differentiation and with our closed-form gradient expression. We provide the related plots in \cref{fig:computation}. We observe that with closed-form gradients, our method puts a little computation overhead and increasing $k$ has marginal effect. On the contrary, with automatic differentiation, the computational complexity substantially increases.

We have further provided the inference times for various optimization steps ($k$) in \cref{fig:inference_computation}. The additional computational complexity introduced by our method is minor, resulting in a less than 6\% increase in the time per image (from 43.1 ms to 45.6 ms) within the typical operation interval of $k$ (25-50). Therefore, our method remains computationally feasible for real-time processing.

\input{supplementary/figures/fig_parameter_search}

\subsection{Hyperparameter Selection}
\label{sec:tuning}

We first perform a screening experiment to see the effect of the parameters. We design a 2-level fractional factorial (\ie, a subset of the all possible combinations) experiment.

We provide the results in \cref{tab:fractional_factorial}. In our analysis, we find that \textit{lower the better} for $\lambda$ and $\mu$. Thus, we set $\mu=0.3$ and $\lambda=0.1$. $\varepsilon$ is observed to have the most effect and number of prototypes, $m$, seems to have no significant effect. Nevertheless, we jointly search for  $m$ and $\varepsilon$. To this end, we perform grid search in CUB dataset with Contrastive (C2) and Proxy NCA++  (PNCA) losses. We provide the results in \cref{fig:parameter_search}-(a). We see that both losses have their best performance when $m=64$. On the other hand, $\varepsilon=5.0$ works better for C2 while $\varepsilon=0.5$ works better for PNCA. We additionally perform a small experiment to see whether $\varepsilon=0.5$ is the case for Proxy Anchor loss as well and observe that $\varepsilon=0.5$ is a better choice over $\varepsilon=5.0$. As the result of $m$-$\varepsilon$ search, we set $\varepsilon=5.0$ for non-proxy based losses and $\varepsilon=0.5$ for proxy-based losses.

%
\input{supplementary/figures/table_fractional_factorial}

Fixing $\mu=0.3,\lambda=0.1,\varepsilon=0.5\text{( or }5.0\text{)}$, we further experiment the effect of number of prototypes $m$ in large datasets (\ie, SOP and In-shop). We provide the corresponding performance plots in \cref{fig:parameter_search}-(b). Supporting our initial analysis, $m$ seemingly does not have a significant effect once it is not small (\eg, $m\geqslant 64$). We observe that any choice of $m\geqslant 64$ provides performance improvement. With that being said, increasing $m$ does not bring substantial improvement over relatively smaller values. Considering the results of the experiment, we set $m=128$ for SOP and In-shop datasets since both C2 and PNCA losses perform slightly better with $m=128$ than the other choices of $m$.

\subsection{Experimental Setup}
\label{sec:experiment_details}

\subsubsection{Datasets}

We perform our experiments on 4 widely-used benchmark datasets: Stanford Online Products (SOP) \cite{oh2016deep}, In-shop \cite{liu2016deepfashion}, Cars196 \cite{krause2014submodular} and, CUB-200-2011 (CUB) \cite{wah2011caltech}.

\textbf{SOP} has 22,634 classes with 120,053 product images. The first 11,318 classes (59,551 images) are split for training and the other 11,316 (60,502 images) classes are used for testing.

\textbf{In-shop} has 7,986 classes with 72,712 images. We use 3,997 classes with 25,882 images as the training set. For the evaluation, we use 14,218 images of 3,985 classes as the query and 12,612 images of 3,985 classes as the gallery set.

\textbf{Cars196} contains 196 classes with 16,185 images. The first 98 classes (8,054 images) are used for training and remaining 98 classes (8,131 images) are reserved for testing.

\textbf{CUB-200-2011} dataset consists of 200 classes with 11,788 images. The first 100 classes (5,864 images) are split for training, the rest of 100 classes (5,924 images) are used for testing.

\textbf{Data augmentation} follows \cite{musgrave2020metric}. During training, we  resize each image so that its shorter side has length 256, then make a random crop between 40 and 256, and aspect ratio between $\nicefrac{3}{4}$ and $\nicefrac{4}{3}$. We resize the resultant image to $227\sxtimes 227$ and apply random horizontal flip with $50\%$ probability. During evaluation, images are resized to 256 and then center cropped to $227\sxtimes 227$.

\subsubsection{Training Splits}

\textbf{Fair evaluation.} We split datasets into disjoint training, validation and test sets according to \cite{musgrave2020metric}. In particular, we partition $\nicefrac{50\%}{50\%}$ for training and test, and further split training data to 4 partitions where 4 models are to be trained by exploiting $\nicefrac{1}{4}$ as validation while training on $\nicefrac{3}{4}$.

\textbf{Conventional evaluation.} Following relatively \textit{old-fashioned} conventional evaluation \cite{oh2016deep}, we use the whole train split of the dataset for training and we use the test split for evaluation as well as monitoring the training for early stopping.

\textbf{Hyperparameter tuning.} For the additional experiments related to the effect of hyperparameters, we split training set into 5 splits and train a single model on the $\nicefrac{4}{5}$ of the set while using $\nicefrac{1}{5}$ for the validation.

\subsubsection{Evaluation Metrics} 
\label{sec:metrics}
We consider precision at 1 (P@1) and mean average precision (MAP@R) at R where R is defined for each query\footnote{A query is an image for which similar images are to be retrieved, and the references are the images in the searchable database.} and is the total number of true references as the query. Among those, MAP@R performance metric is shown to better reflect the geometry of the embedding space and to be less noisy as the evaluation metric \cite{musgrave2020metric}. Thus, we use MAP@R to monitor training in our experiments except for conventional evaluation setting where we monitor P@1.

\textit{P@1:} Find the nearest reference to the query. The score for that query is 1 if the reference is of the same class, 0 otherwise. Average over all queries gives P@1 metric.

\textit{P@R:} For a query, $i$, find $R_i$ nearest references to the query and let $r_i$ be the number of true references in those $R_i$-neighbourhood. The score for that query is $\text{P@R}_i=\nicefrac{r_i}{R_i}$. Average over all queries gives P@R metric, \ie, $\text{P@R} = \tfrac{1}{n}\sum\limits_{i\in[n]}\text{P@R}_i$, where $n$ is the number of queries.

\textit{MAP@R:} For a query, $i$, we define $\text{MAP@R}_i \coloneqq \tfrac{1}{R_i}\sum\limits_{i\in[R_i]} P(i)$, where $P(i)=\text{P@R}_i$ if $i^{th}$ retrieval is correct or 0 otherwise. Average over all queries gives MAP@R metric, \ie, $\text{MAP@R} = \tfrac{1}{n}\sum\limits_{i\in[n]}\text{MAP@R}_i$, where $n$ is the number of queries.

\subsubsection{Training Procedure}

\textbf{Fair evaluation.} We use Adam \cite{kingma2014adam} optimizer with constant $10^{\shortminus 5}$ learning rate, $10^{\shortminus 4}$ weight decay, and default moment parameters, $\beta_1{=}.9$ and $\beta_2{=}.99$. We use batch size of 32 (4 samples per class). We evaluate validation MAP@R for every 100 steps of training in CUB and Cars196, for 1000 steps in SOP and In-shop. We stop training if no improvement is observed for 15 steps in CUB and Cars196, and 10 steps in SOP and In-shop. We recover the parameters with the best validation performance. Following \cite{musgrave2020metric}, we train 4 models for each $\nicefrac{3}{4}$ partition of the train set. Each model is trained 3 times. Hence, we have $3^4=81$ many realizations of 4-model collections. We present the average performance as well as the standard deviation (\textit{std}) of such 81 models' evaluations.

\textbf{Conventional evaluation.} We use Adam \cite{kingma2014adam} optimizer with default moment parameters, $\beta_1{=}.9$ and $\beta_2{=}.99$. Following recent works \cite{kim2020proxy}, we use \textit{reduce on plateau} learning rate scheduler with patience 4. The initial learning rate is $10^{\shortminus 5}$ for CUB, and $10^{\shortminus 4}$ for Cars, SOP and In-shop. We use $10^{\shortminus 4}$ weight decay for BNInception backbone and $4\,10^{\shortminus 4}$ wight decay for ResNet50 backbone. We use batch size of 128 (4 samples per class) for BNInception backbone and 112 (4 samples per class) for ResNet backbone (following \cite{roth2020revisiting}). We evaluate validation P@1 for every 25 steps of training in CUB and Cars196, for 250 steps in SOP and In-shop. We stop training if no improvement is observed for 15 steps in CUB and Cars196, and 10 steps in SOP and In-shop. We recover the parameters with the best validation performance. We repeat each experiment 3 times and report the best result. For the evaluations on LIBC framework \cite{intrabatch}, we follow their experimental setting.

\textbf{Hyperparameter tuning.} We use Adam \cite{kingma2014adam} optimizer with constant $10^{\shortminus 5}$ learning rate, $10^{\shortminus 4}$ weight decay, and default moment parameters, $\beta_1{=}.9$ and $\beta_2{=}.99$. We use batch size of 32 (4 samples per class). We evaluate validation MAP@R for every 100 steps of training in CUB and Cars196, for 1000 steps in SOP and In-shop. We stop training if no improvement is observed for 10 steps in CUB and Cars196, and 7 steps in SOP and In-shop. We recover the parameters with the best validation performance. We train a single model on the $\nicefrac{4}{5}$ of the training set while using $\nicefrac{1}{5}$ for the validation. We repeat each experiment 3 times and report the averaged result.

\subsubsection{Embedding vectors}

\textbf{Fair evaluation.} Embedding dimension is fixed to 128. During training and evaluation, the embedding vectors are $\ell2$ normalized. We follow the evaluation method proposed in \cite{musgrave2020metric} and produce two results: $i)$ Average performance (128 dimensional) of 4-fold models and $ii)$ Ensemble performance (concatenated 512 dimensional) of 4-fold models where the embedding vector is obtained by concatenated 128D vectors of the individual models before retrieval.

\textbf{Conventional evaluation.} Embedding dimension is 512 in both BNInception and ResNet50 experiments.

\textbf{Hyperparameter tuning.} Embedding dimension is fixed to 128.

\subsubsection{Baselines with GSP}

We evaluate our method with \textit{C1+XBM+GSP}: Cross-batch memory (XBM) \cite{wang2020cross} with original Contrastive loss (C1) \cite{hadsell2006dimensionality}, \textit{C2+GSP}: Contrastive loss with positive margin \cite{wu2017sampling}, \textit{MS+GSP}: Multi-similarity (MS) loss \cite{Wang_2019_CVPR_MS}, \textit{Triplet+GSP}: Triplet loss \cite{schroff2015facenet}, \textit{PNCA+GSP}: ProxyNCA++ loss \cite{teh2020proxynca++}, \textit{PAnchor+GSP}: ProxyAnchor loss \cite{kim2020proxy}.

\subsubsection{Hyperparameters}

For the hyperparameter selection, we exploit the work \cite{musgrave2020metric} that performed parameter search via Bayesian optimization on variety of losses. We further experiment the suggested parameters from the original papers and official implementations. We pick the best performing parameters. We perform no further parameter tuning for the baseline methods' parameters when applied with our method to purely examine the effectiveness of our method. 

\textbf{C1}: We adopted XBM's official implementation for fair comparison. We use 0.5 margin for all datasets.

\textbf{C2}: C2 has two parameters, $(m^{+}, m^{-})$: positive margin, $m^{+}$, and negative margin. We set $(m^{+}, m^{-})$ to $(0, 0.3841)$, $(0.2652, 0.5409)$, $(0.2858, 0.5130)$, $(0.2858, 0.5130)$ for CUB, Cars196, In-shop and SOP, respectively.

\textbf{Triplet}: We set its margin to 0.0961, 0.1190, 0.0451, 0.0451 for CUB, Cars196, In-shop and SOP, respectively.

\textbf{MS}: We set its parameters $(\alpha, \beta, \lambda)$ to $(2, 40, 0.5)$, $(14.35, 75.83, 0.66)$, $(8.49, 57.38, 0.41)$, $(2, 40, 0.5)$ for CUB, Cars196, In-shop and SOP, respectively.

\textbf{ProxyAnchor}: We set its two paremeters $(\delta, \alpha)$ to $(0.1, 32)$ for all datasets. We use 1 sample per class in batch setting (\ie, 32 classes with 1 samples per batch), we perform 1 epoch warm-up training of the embedding layer, and we apply learning rate multiplier of 100 for the proxies during training. For SOP dataset, we use $5\,10^{\shortminus 6}$ learning rate.

\textbf{ProxyNCA++}: We set its temperature parameter to 0.11 for all datasets. We use 1 sample per class in batch setting (\ie, 32 classes with 1 samples per batch), we perform 1 epoch warm-up training of the embedding layer, and we apply learning rate multiplier of 100 for the proxies.

\textbf{XBM}: We evaluate XBM with C1 since in the original paper, contrastive loss is reported to be the best performing baseline with XBM. We set the memory size of XBM according to the dataset. For CUB and Cars196, we use memory size of 25 batches. For In-shop, we use 400 batches and for SOP we use 1400 batches. We perform 1K steps of training with the baseline loss prior to integrate XBM loss in order to ensure XBM's \textit{slow drift} assumption.

\textbf{GSP}: For the hyperparameters of our method, we perform parameters search, details of which are provided in \cref{sec:tuning}. We use 64-many prototypes in CUB and Cars, and 128-many prototypes in SOP and In-shop. We use $\varepsilon{=}0.5$ for proxy-based losses and $\varepsilon{=}5.0$ for non-proxy losses. For the rest, we set  $\mu{=}0.3$, $\epsilon{=}0.05$, and we iterate until $k{=}100$ in Proposition 4.1. For zero-shot prediction loss coefficient (\ie, $(1{\shortminus}\lambda)\mathcal{L}_{DML} + \lambda\mathcal{L}_{ZS}$), we set $\lambda{=}0.1$.

\section{Details of the Other Empirical Work}

\subsection{Synthetic Study}
We design a synthetic empirical study to evaluate GSP in a fully controlled manner. We consider 16-class problem such that classes are defined over trainable tokens. In this setting, tokens correspond to semantic entities but we choose to give a specific working to emphasize that they are trained as part of the learning. Each class is defined with 4 distinct tokens and there are also 4 background tokens shared by all classes. For example, a \emph{"car"} class would have tokens like \emph{"tire"} and \emph{"window"} as well as background tokens of \emph{"tree"} and \emph{"road"}.

We sample class representations from both class specific and background tokens according to a mixing ratio $\Tilde{\mu}\sim\mathcal{N}(0.5, 0.1)$. We sample a total of 50 tokens and such a 50-many feature collection will correspond to a training sample (\ie, we are mimicking CNN's output with trainable tokens). For instance, given class tokens for class-$c$, $\nu^{(c)}=\{\nu^{(c)}_1,\nu^{(c)}_2,\nu^{(c)}_3,\nu^{(c)}_4\}$ and shared tokens, $\nu^{(b)}=\{\nu^{(b)}_1,\nu^{(b)}_2,\nu^{(b)}_3,\nu^{(b)}_4\}$; we first sample $\mu=0.4$ and then sample 20 tokens from  $\nu^{(c)}$ with replacement, and 30 tokens from $\nu^{(b)}$, forming a feature collection for a class-$c$, \ie, $f^{(c)}=\{\nu^{(c)}_3, \nu^{(c)}_1, \nu^{(c)}_1, \nu^{(c)}_3,\ldots, \nu^{(b)}_4, \nu^{(b)}_3,\nu^{(b)}_4, \nu^{(b)}_1, \ldots\}$ We then obtain global representations using GAP and GSP.

We do not apply $\ell2$ normalization on the global representations. We also constrain the range of the token vectors to be in between $[\shortminus 0.3, 0.3]$ to bound the magnitude of the learned vectors. We use default Adam optimizer with  $10^{\shortminus 4}$ learning rate and perform early stopping with 30 epoch patience by monitoring MAP@R. In each batch, we use 4 samples from 16 classes.

\subsection{Cifar Collage}
\label{sec:cifar_collage_details}

We consider the 20 \textit{super-classes} of Cifar100 dataset \cite{cifar} where each has 5 sub-classes. For each super-class, we split the sub-classes for train (2), validation (1), and test (2). We consider 4 super-classes as the shared classes and compose $4{\sxtimes}4$-stitched collage images for the rest 16 classes. In particular, we sample an image from a class and then sample 3 images from shared classes. We illustrate a sample formation process in \cref{fig:cifarcollage_1}.

\input{supplementary/figures/fig_cifar_collage_1}

We should note that the classes exploited in training, validation and test are disjoint. For instance, if a \textit{tree} class is used as a shared class in training, then that \textit{tree} class does not exist in validation or test set as a shared feature. Namely, in our problem setting, both the background and the foreground classes are disjoint across training, validation and test sets. Such a setting is useful to analyze zero-shot transfer capability of our method. 

We use ResNet20 (\ie, 3 stages, 3 blocks) backbone pretrained on Cifar100 classification task. We use $\ell2$ normalization on global representations. We use default Adam optimizer with initial $0.001$ learning rate. We use \textit{reduce on plateau} with 0.5 decay factor and 5 epochs patience. For GSP, we set $m=128,\mu=0.2,\varepsilon=10, \lambda=0.5$. We use 4 samples from 16 classes in a batch.

\subsection{Evaluation of Zero-shot Prediction Loss}

We train on Cifar10 \cite{cifar} dataset with 8 prototypes using ProxyNCA++ \cite{teh2020proxynca++} (PNCA) loss with and without $\mathcal{L}_{ZS}$. We then use test set to compute attribute histograms for each class. Namely, we aggregate the marginal transport plans of each sample in a class to obtain the histogram. Similarly, for each class, we compute the mean embedding vector (\ie, we average embedding vectors of the samples of a class). Our aim is to fit a linear predictor to map attribute vectors to the mean embeddings.

To quantify the zero-shot prediction performance, we randomly split the classes into half and apply cross-batch zero-shot prediction. Specifically, we fit a linear predictor using 5 classes and then use that transformation to map the other 5 classes to their mean embeddings. We then compute pairwise distance between the predicted means and the true means. We then evaluate the nearest neighbour classification performance. We use both $\ell2$ distance and cosine distance while computing the pairwise distances. We repeat the experiment 1000 times with different class splits. 

%% file: supplementary/figures/table_conventional.tex
\begin{table*}[!hb]
\centering
\caption{Comparison with the existing methods for the retrieval task in conventional experimental settings with BN-Inception and ResNet50 backbones where superscripts denote embedding size. Red: the best. Blue: the second best. Bold: previous SOTA. \, \textsuperscript{$\ddagger$}\textit{Results obtained from } \cite{intrabatch}.}
\label{tab:conventional}
\begin{subtable}[t]{0.5\linewidth}
\centering
\caption{}
\label{tab:conventional_bni}
\resizebox{\linewidth}{!}{%
\begin{tabular}[t]{@{}lcccc@{}}
\toprule
{{Backbone $\rightarrow$}}  & \multicolumn{4}{c}{\textbf{BN-Inception-512D}}                                                                                                                         \\ \cmidrule(lr){2-5}
{{Dataset} $\rightarrow$}            & \textbf{CUB}                          & \textbf{Cars196}                      & \textbf{SOP}                          & \textbf{In-shop}                      \\ \cmidrule(lr){2-2} \cmidrule(lr){3-3} \cmidrule(lr){4-4} \cmidrule(lr){5-5}
{{Method} $\downarrow$}              & \textbf{R@1}                          & \textbf{R@1}                          & \textbf{R@1}                          & \textbf{R@1}                          \\ \midrule
{C1+XBM \cite{wang2020cross}}     & 65.80                                 & 82.00                                 & {\color[HTML]{00D2CB} \textbf{79.50}} & 89.90                                 \\
{ProxyAnchor \cite{kim2020proxy}} & 68.40                                 & 86.10                                 & 79.10                                 & {\color[HTML]{FE0000} \textbf{91.50}} \\
{DiVA \cite{milbich2020diva}}     & 66.80                                 & 84.10                                 & 78.10                                 & -                                     \\
{ProxyFewer \cite{zhu2020fewer}}  & 66.60                                 & 85.50                                 & 78.00                                 & -                                     \\
{Margin+S2SD \cite{roth2021s2sd}  } & {\color[HTML]{FE0000} \textbf{68.50}} & {\color[HTML]{FE0000} \textbf{87.30}} & 79.30                                 & -                                     \\ \midrule
{C1+XBM}                          & $\underset{\scriptsize{(23.59)}}{64.32}$                              & $\underset{\scriptsize{(21.67)}}{77.63}$                                 & $\underset{\scriptsize{(52.59)}}{79.29}$                                  & $\underset{\scriptsize{(61.39)}}{90.16}$                                \\
{C1+XBM+GSP}                      & $\underset{\scriptsize{(25.35)}}{64.99}$                              & $\underset{\scriptsize{(22.51)}}{79.07}$                                & {\color[HTML]{FE0000} $\underset{\scriptsize{(52.70)}}{\mathbf{79.59}}$ } & {\color[HTML]{00D2CB} $\underset{\scriptsize{(63.25)}}{\mathbf{90.92}}$} \\ \bottomrule
\end{tabular}%
}
\end{subtable}
%
%
\begin{subtable}[t]{.49\linewidth}
\centering
\caption{}
\label{tab:conventional_resnet}
\resizebox{\linewidth}{!}{%
\begin{tabular}[t]{@{}lcccc@{}}
\toprule
{{Backbone $\rightarrow$}}                & \multicolumn{4}{c}{\textbf{ResNet50}}                                                                                                                                 \\ \cmidrule(lr){2-5}
{{Dataset $\rightarrow$}}                 & \textbf{CUB}                          & \textbf{Cars196}                      & \textbf{SOP}                                  & \textbf{In-shop}                      \\ \cmidrule(lr){2-2} \cmidrule(lr){3-3} \cmidrule(lr){4-4} \cmidrule(lr){5-5}
{{Method $\downarrow$}}                   & \textbf{R@1}                          & \textbf{R@1}                          & \textbf{R@1}                                  & \textbf{R@1}                          \\ \midrule
{C1+XBM$^{128}$  \cite{wang2020cross}}           & -                                     & -                                     & 80.60                                         & 91.30                                 \\
{ProxyAnchor$^{512}$ \cite{kim2020proxy}}       & 69.70 & 87.70                                 & 80.00\textsuperscript{$\ddagger$}              & 92.10\textsuperscript{$\ddagger$}      \\
{DiVA$^{512}$ \cite{milbich2020diva}}           & 69.20                                 & 87.60                                 & 79.60                                         & -                                     \\
{ProxyNCA++$^{512}$ \cite{teh2020proxynca++}}  & 66.30                                 & 85.40                                 & 80.20                                         & 88.60                                 \\
{Margin+S2SD$^{512}$ \cite{roth2021s2sd}}       & 69.00                                 & {\color[HTML]{00D2CB} \textbf{89.50}} & 81.20                                         & -                                     \\
{LIBC$^{512}$ \cite{intrabatch}}               & 70.30                                 & 88.10                                 & \textbf{81.40}                                & {\color[HTML]{00D2CB} \textbf{92.80}} \\
{MS+Metrix$^{512}$ \cite{venkataramanan2022it}} & {\color[HTML]{FE0000} \textbf{71.40}} & {\color[HTML]{FE0000} \textbf{89.60}} & 81.00                                         &  92.20 \\
{PAnchor+DIML$^{128}$ \cite{zhao2021towards}} & $\underset{\scriptsize{(25.58)}}{66.46}$                        & $\underset{\scriptsize{(28.11)}}{86.13}$                           & $\underset{\scriptsize{(43.04)}}{79.22}$                                & -                          \\ \midrule
{LIBC+GSP$^{512}$}                                & {\color[HTML]{00D2CB} \textbf{70.70 }}                        & 88.43                           & {\color[HTML]{FE0000} \textbf{81.65}}                              & {\color[HTML]{FE0000} \textbf{93.30 }}                        \\
{C1+XBM$^{512}$}                                & $\underset{\scriptsize{(25.38)}}{66.68}$                           & $\underset{\scriptsize{(25.34)}}{82.83}$                          &  $\underset{\scriptsize{(55.66)}}{81.44}$ & $\underset{\scriptsize{(64.00)}}{91.56}$                         \\
{C1+XBM+GSP$^{512}$}                            & $\underset{\scriptsize{(25.51)}}{66.63}$                         & $\underset{\scriptsize{(25.76)}}{82.60}$                          & {\color[HTML]{00D2CB} $\underset{\scriptsize{(55.91)}}{\mathbf{81.54}}$} & $\underset{\scriptsize{(64.43)}}{91.75}$                         \\ \bottomrule
\end{tabular}%
}
\end{subtable}
\end{table*}

%% file: supplementary/figures/table_r_at_k.tex
\begin{table}[H]
\centering
\caption{R@K performances using 512D embeddings from LIBC \cite{intrabatch} and XBM \cite{wang2020cross} with ResNet50 backbone}
\label{tab:r_at_k}
\resizebox{\columnwidth}{!}{%
\begin{tabular}{@{}lcccccccc@{}}
\toprule
Dataset$\rightarrow$ & \multicolumn{4}{c}{\textbf{CUB}}                              & \multicolumn{4}{c}{\textbf{Cars196}}                         \\ \cmidrule(lr){2-5} \cmidrule(lr){6-9}
Method$\downarrow$   & \textbf{R@1} & \textbf{R@2}  & \textbf{R@4}  & \textbf{R@8}   & \textbf{R@1} & \textbf{R@2}  & \textbf{R@4}  & \textbf{R@8}  \\ \midrule
LIBC+GSP             & 70.70        & 80.72         & 88.18         & 92.64          & 88.43        & 93.03         & 95.78         & 97.69         \\
XBM+GSP              & 66.63        & 77.43         & 85.26         & 91.02          & 82.60        & 89.04         & 92.67         & 95.62         \\ \midrule
Dataset$\rightarrow$ & \multicolumn{4}{c}{\textbf{SOP}}                              & \multicolumn{4}{c}{\textbf{In-shop}}                         \\ \cmidrule(lr){2-5} \cmidrule(lr){6-9}
Method$\downarrow$   & \textbf{R@1} & \textbf{R@10} & \textbf{R@50} & \textbf{R@100} & \textbf{R@1} & \textbf{R@10} & \textbf{R@20} & \textbf{R@40} \\ \midrule
LIBC+GSP             & 81.65        & 91.37         & 94.85         & 96.00          & 93.30        & 98.54         & 98.96         & 99.25         \\
XBM+GSP              & 81.54        & 91.84         & 95.18         & 96.29          & 91.75        & 97.83         & 98.52         & 99.01         \\ \bottomrule
\end{tabular}%
}
\end{table}

%% file: supplementary/figures/table_imagenet.tex
\setlength{\columnsep}{10pt}
\begin{table}[h]
\centering
\caption{Evaluation in classification task}
\label{tab:imagenet}
\begin{tabular}{@{}lccc@{}}
\toprule
{ImageNet} & \textbf{Acc.}  & \textbf{P@1}   & \textbf{MAP@R} \\ \midrule
{RN50V2}  & 75.26          & 69.30          & 41.18          \\
{+GSP}    & \textbf{76.53} & \textbf{71.34} & \textbf{42.58} \\ \midrule
{ENV2B3}  & 80.03    & 79.23       & 59.98          \\
{+GSP}    & \textbf{82.00} & \textbf{80.80} & \textbf{62.75} \\ \bottomrule
\end{tabular}%
\end{table}

%% file: supplementary/figures/table_pooling.tex
\begin{table*}[hb]
\centering
\caption{Evaluation of feature pooling methods on \textit{Cifar Collage} and CUB datasets with Contrastive and ProxyNCA++ losses for DML task. Red: the best, Blue: the second best, Bold: the third best.}
\label{tab:pooling}
\begin{tabular}{@{}lcccc@{}}
\toprule
                 & \multicolumn{4}{c}{128D - MAP@R}                                                                                                                              \\ \cmidrule(l){2-5} 
Dataset$\rightarrow$                        & \multicolumn{2}{c}{\textbf{Cifar Collage}}                                    & \multicolumn{2}{c}{\textbf{CUB}}                                              \\ \cmidrule(lr){2-3} \cmidrule(lr){4-5} 
{Method$\downarrow$}  {Loss$\rightarrow$} & \textbf{Contrastive}                  & \textbf{ProxyNCA++}                   & \textbf{Contrastive}                  & \textbf{ProxyNCA++}                   \\ \midrule
CBAM \cite{cbam}                                                             & 7.87                                  & 10.99                                 & 18.45                                 & 18.21                                 \\
CroW    \cite{crow}                                                         & 10.09                                 & 11.48                                 & 20.88                                 & 20.42                                 \\
DeLF \cite{delf}                                                            & 11.44                                 & {\color[HTML]{00D2CB} \textbf{24.83}} & \textbf{21.42}                        & 20.51                                 \\
GeMax  \cite{murray2016interferences}                                                          & 7.04                                  & 7.83                                  & 18.85                                 & 17.83                                 \\
GeMean    \cite{gmeanp}                                                        & 10.97                                 & 10.60                                  & {\color[HTML]{00D2CB} \textbf{21.50}} & {\color[HTML]{00D2CB} \textbf{20.71}} \\
GSoP        \cite{gsop}                                                      & 11.15                                 & 17.73                                 & 20.52                                 & 15.72                                 \\
OTP-$(8\sxtimes 16)$ \cite{mialon2021trainable}                                                               & 7.02                                  & 11.55                                 & 15.19                                  & 13.57                                  \\
 OTP-$(64\sxtimes 128)$   \cite{mialon2021trainable}      & 7.57                                  & 11.79                                 & 20.88                                 & 20.48                               \\
SOLAR \cite{solar}                                                              & 17.30                         & 20.36                                 & 19.89                                 & 20.14                                 \\
TFM-1 \cite{dosovitskiy2021image}                                                           & 8.84               & 10.83              & 17.82           & 19.13          \\
 TFM-2 \cite{dosovitskiy2021image}          & 16.48              & 21.00              & 17.16           & 18.13          \\
 TFM-4 \cite{dosovitskiy2021image}                                     & 18.51              & \textbf{21.56}              & 16.91           & 18.22          \\
TFM-8 \cite{dosovitskiy2021image}                  & 18.18              & 19.68              & 16.31           & 17.47          \\
T-SMK \cite{tamsk}                                                           & 9.21                                  & 13.15                                 & 21.01                                 & 20.23                                 \\
VLAD-$(8\sxtimes 16)$ \cite{vlad}                                                            & {\textbf{21.73}} & 19.68                                 & 15.19                                 & 13.08                                 \\
VLAD-$(64\sxtimes 128)$ \cite{vlad}   & {\color[HTML]{00D2CB} \textbf{22.52}} & 21.15 & 16.67 & 16.53 \\
WELDON \cite{weldon}                                                       & 13.81                                 & 20.38                        & 20.67                                 & 20.31                                 \\
GAP                                                               & 8.09                                  & 10.68                                 & 20.58                                 & 20.63         \\
GMP                                                               & 9.53                                  & 11.25                                 & 20.66                                 & 20.33                                 \\
GMP+GAP                                                           & 10.01                                 & 11.85                                 & 20.88                                 & \textbf{20.68}                        \\
\textbf{GSP}                                                              & {\color[HTML]{FE0000} \textbf{22.68}} & {\color[HTML]{FE0000} \textbf{27.61}} & {\color[HTML]{FE0000} \textbf{21.52}} & {\color[HTML]{FE0000} \textbf{20.75}} \\ \bottomrule
\end{tabular}%
\end{table*}

%% file: supplementary/figures/fig_inference_computation.tex
\begin{figure}[b]
\centering
\centerline{\includegraphics[width=1.0\linewidth,keepaspectratio]{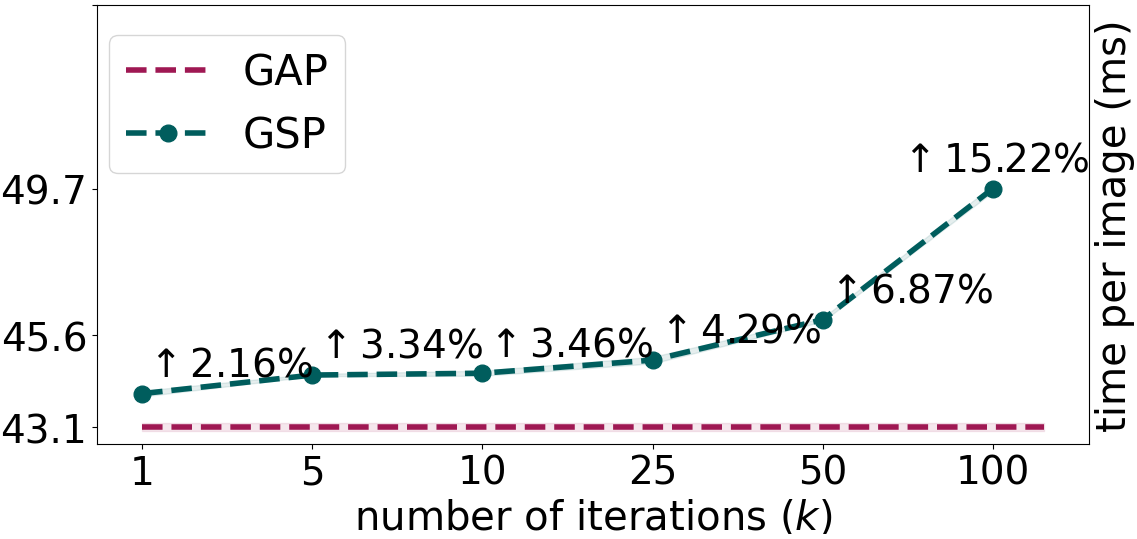}}
\caption{Computation increase ($\uparrow$) in inference with GSP using $k$ iterations} 
\label{fig:inference_computation}
\end{figure}

%% file: supplementary/figures/fig_computation.tex
\begin{figure}[ht]
\centering
\centerline{\includegraphics[width=0.8\linewidth,keepaspectratio]{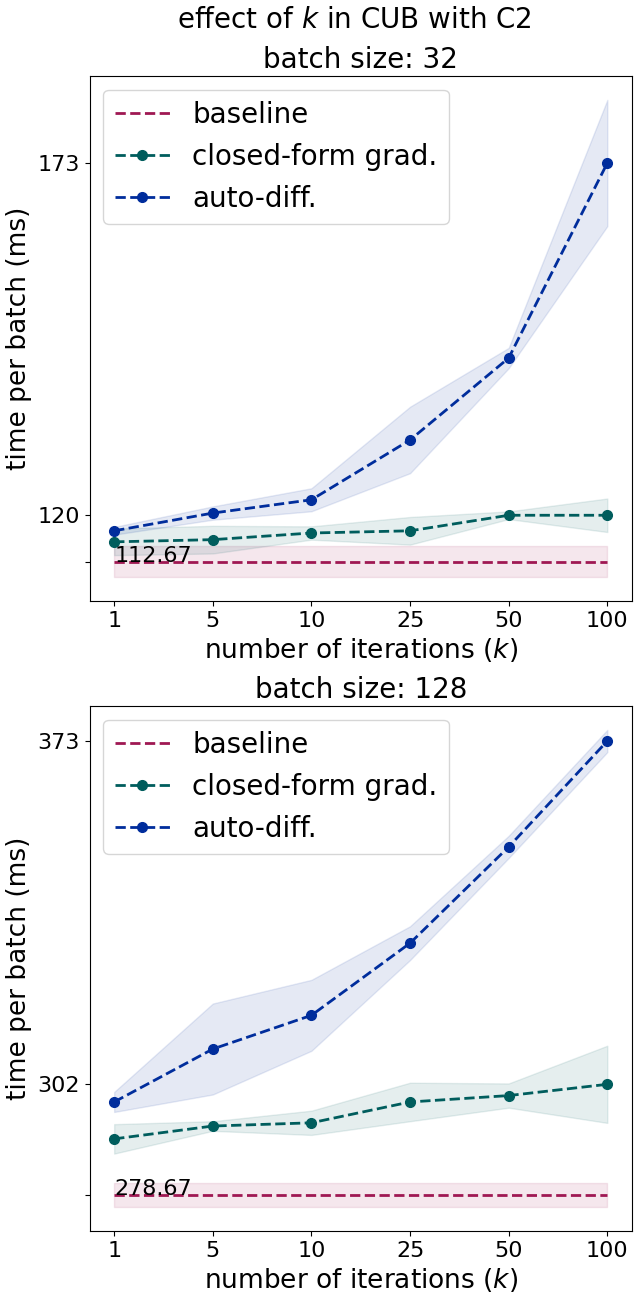}}
\caption{Comparing closed-form gradient with automatic differentiation through analyzing the effect of $k$ on computation in CUB  with C2 loss. Shaded regions represent $\mp$\textit{std}.} 
\label{fig:computation}
\end{figure}

%% file: supplementary/figures/fig_parameter_search.tex
\begin{figure*}[hb]

\begin{minipage}[b]{0.49\linewidth}
  \centering
  \centerline{\includegraphics[width=1.0\linewidth,keepaspectratio]{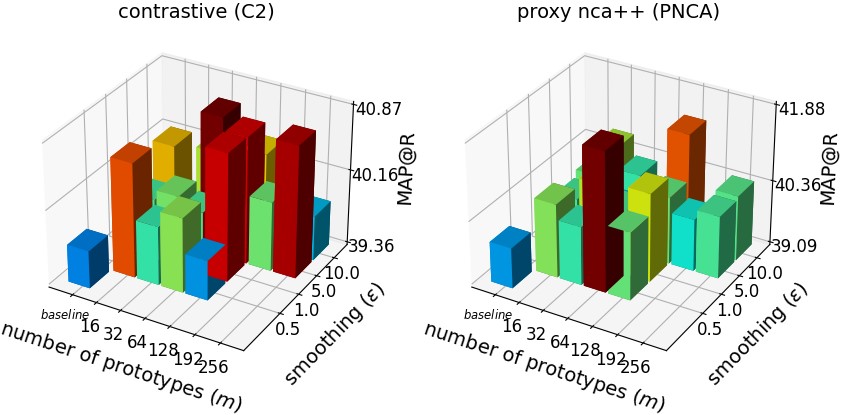}}
\centerline{(a)}
\label{fig:param2d}
\end{minipage}
\begin{minipage}[b]{0.49\linewidth}
  \centering
  \centerline{\includegraphics[width=1.0\linewidth,keepaspectratio]{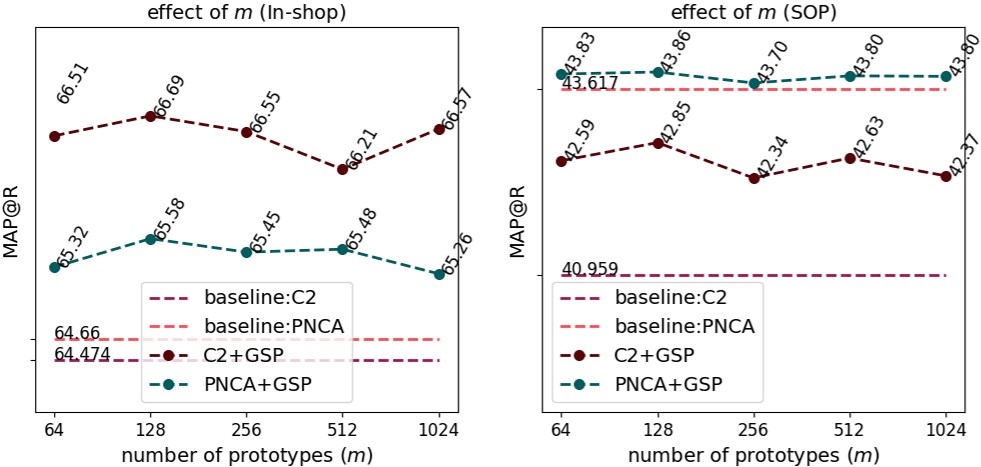}}
\centerline{(b)}
\label{fig:sweeping_m}
\end{minipage}

\caption{Parameter search for $m:$ number of prototoypes and $\varepsilon$: entropy smoothing coefficient. We fix $\mu=0.3$ and $\lambda=0.1$. (a) Searching $m-\varepsilon$ space in CUB dataset. (b) Effect of $m$ once we fix $\varepsilon=5$ for Contrastive (C2) and $\varepsilon=0.5$ for Proxy NCA++ (PNCA).}
\label{fig:parameter_search}
\end{figure*}

%% file: supplementary/figures/table_fractional_factorial.tex
\begin{table}[t]
\centering
\caption{Initial 2-level fractional factorial screening experiments for parameter tuning (conducted in CUB)}
\label{tab:fractional_factorial}
\begin{tabular}{@{}cccc|cc@{}}
\toprule
\multicolumn{4}{c|}{\textbf{Setting}}   & \multicolumn{2}{c}{\textbf{MAP@R}} \\ \midrule
$m$ & $\mu$ & $\varepsilon$ & $\lambda$ & C2               & PNCA            \\ \midrule
16  & 0.3   & 0.5           & 0.1       & 40.63            & 40.59           \\
16  & 0.7   & 0.5           & 0.5       & 40.41            & 40.34           \\
128 & 0.3   & 0.5           & 0.5       & 40.22            & 40.35           \\
128 & 0.7   & 0.5           & 0.1       & 40.07            & 40.85           \\
16  & 0.3   & 20            & 0.5       & 36.11            & 40.51           \\
16  & 0.7   & 20            & 0.1       & 39.11            & 39.88           \\
128 & 0.3   & 20            & 0.1       & 39.61            & 39.12           \\
128 & 0.7   & 20            & 0.5       & 35.36            & 39.92           \\ \midrule
\multicolumn{4}{c|}{Baseline}           & 39.77            & 39.90           \\ \bottomrule
\end{tabular}%
\end{table}

%% file: supplementary/figures/fig_cifar_collage_1.tex
\begin{figure}[H]
\centering
\centerline{\includegraphics[width=.95\linewidth,keepaspectratio]{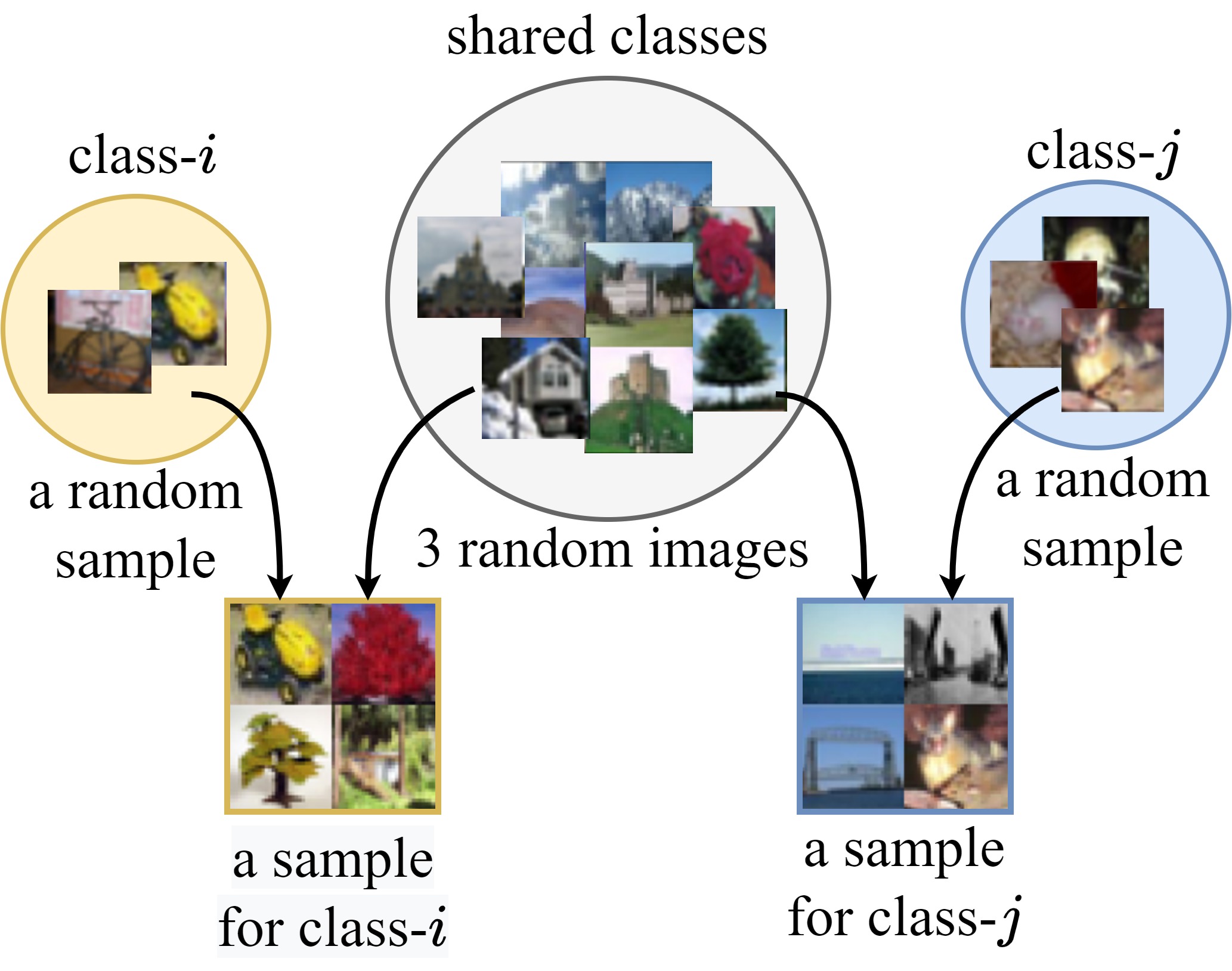}}
\caption{Sample generation for Cifar Collage dataset} 
\label{fig:cifarcollage_1}
\end{figure}

%% file: supplementary/appendix.tex
\onecolumn
\input{supplementary/proofs}
\vspace{160pt}
\input{supplementary/figures/table_method_diff}
\twocolumn
\input{supplementary/ot_operators}

\newpage
\onecolumn
\section{Implementations with Pseudo Codes}
\input{supplementary/figures/algo_pooling}

%% file: supplementary/proofs.tex
\section{Appendix}
\subsection{Proof for Theorem 4.1}

\begin{proof}
$\rho^{\ast}$ is obtained as the solution of the following optimal transport problem:
\begin{equation*}\label{eq:ot_problem}
\rho^{\ast}, \pi^{\ast} = \!\!\!\!\!\!\! \argmin_{\substack{\rho,\pi\geqslant0 \\
\rho_j + \Sigma_i\pi_{ij}=\nicefrac{1}{n}\\
\Sigma_{ij}\pi_{ij}=\mu}}
\!\!\!\!\!\!\! \textstyle\sum_{ij}c_{ij}\pi_{ij}.
\end{equation*}
Given solutions $(\rho^\ast, \pi^\ast)$, for $\mu{=}1$, from the 3$^{rd}$ constraint, we have $\Sigma_{ij}\pi_{ij}^\ast{=}1$. Then, using the $2^{nd}$ constraint, we write:
$$ \textstyle\sum_j\rho^\ast_j + \textstyle\sum_j\textstyle\sum_i \pi^\ast_{ij} = \textstyle\sum_j\tfrac{1}{n}$$
where $j{\in}[n]$ for $n$-many convolutional features. Hence, we have $\sum_j\rho^\ast=0$ which implies $\rho^\ast{=}0$ owing to non-negativity constraint. Finally, pooling weights becomes $p_i=\tfrac{\nicefrac{1}{n}-\cancel{\rho^\ast_i}}{\underset{=1}{\mu}} = \nicefrac{1}{n}$.

\end{proof}

\subsection{Proof for Proposition 4.2}
Before starting our proof, we first derive an iterative approach for the solution of (P2). We then prove that the iterations in Proposition 4.2 can be used to obtain the solution.

We can write (P2) equivalently as:
\begin{equation*}\label{eq:smoothed_problem1}
\begin{split}
\rho^{(\varepsilon)}, \pi^{(\varepsilon)} = \!\!\!\!\!\! \argmin_{\substack{\rho,\pi\geqslant0 \\
\rho_j + \Sigma_i\pi_{ij}=\nicefrac{1}{n}\\
\Sigma_{ij}\pi_{ij}=\mu}}
\!\!\!\!\! \textstyle\sum_{ij}c_{ij}\pi_{ij}&+ 
\tfrac{1}{\varepsilon}(\textstyle\sum_{ij}\pi_{ij}\log\pi_{ij} + \textstyle\sum_j \rho_j\log\rho_j)\\
&+ \textstyle\sum_j 0\rho_j  - \textstyle\sum_{ij}\pi_{ij} - \textstyle\sum_j \rho_j + \textstyle\sum_{ij}\mathrm{e}^{\shortminus\varepsilon c_{ij}}  + \textstyle\sum_{j}\mathrm{e}^{\shortminus\varepsilon 0}
\end{split}
\end{equation*}
Rearranging the terms we get:
\begin{equation*}
    \rho^{(\varepsilon)}, \pi^{(\varepsilon)} = \!\!\!\!\!\! \argmin_{\substack{\rho,\pi\geqslant0 \\
\rho_j + \Sigma_i\pi_{ij}=\nicefrac{1}{n}\\
\Sigma_{ij}\pi_{ij}=\mu}}
\!\!\!\!\!\! \textstyle\sum_{ij}\pi_{ij}\log\tfrac{\pi_{ij}}{\mathrm{e}^{\shortminus\varepsilon c_{ij}}}
+ \textstyle\sum_{j}\rho_{j}\log\tfrac{\rho_{j}}{\mathrm{e}^{\shortminus\varepsilon 0}}
- \textstyle\sum_{ij}\pi_{ij} - \textstyle\sum_j \rho_j + 
\textstyle\sum_{ij}\mathrm{e}^{\shortminus\varepsilon c_{ij}}  + \textstyle\sum_{j}\mathrm{e}^{\shortminus\varepsilon 0}
\end{equation*}
which is generalized \textit{Kullback–Leibler divergence} (KLD) between $(\rho, \pi)$ and $(\exp{({\shortminus}\varepsilon 0)}, \exp{({\shortminus}\varepsilon c)})$. Hence, we reformulate the problem as a KLD prjoection onto a convex set, which can be solved by \textit{Bregman Projections} (\ie, alternating projections onto constraint sets) \cite{bregman1967relaxation, bauschke2000dykstras}. Defining $\mathcal{C}_1 \coloneqq \{(\rho, \pi)\mid \rho_j + \textstyle\sum_{ij}\pi_{ij}=\nicefrac{1}{n} \,\,\forall j\}$  and $\mathcal{C}_2 \coloneqq \{(\rho, \pi)\mid  \textstyle\sum_{ij}\pi_{ij}=\mu \}$, and omitting constants, we can write the problem as:
\begin{equation*}\tag{P2$^\prime$}
    \rho^{(\varepsilon)}, \pi^{(\varepsilon)} = \!\!\!\! \argmin_{\substack{\rho,\pi\geqslant0 \\
(\rho,\pi)\in\mathcal{C}_1\cap\mathcal{C}_2}} \textstyle\sum_{ij}\pi_{ij}(\log\tfrac{\pi_{ij}}{\mathrm{e}^{\shortminus\varepsilon c_{ij}}}\shortminus 1)
+ \textstyle\sum_{j}\rho_{j}(\log\tfrac{\rho_{j}}{\mathrm{e}^{\shortminus\varepsilon 0}}\shortminus 1)
\end{equation*}
Given, $(\rho^{(k)},\pi^{(k)})$), at iteration $k$, KLD projection onto $\mathcal{C}_1$, \ie,  $(\rho^{(k{+}1)},\pi^{(k{+}1)})\coloneqq \mathcal{P}_{\mathcal{C}_1}^{KL}(\rho^{(k)},\pi^{(k)})$, reads:
$$\rho^{(k{+}1)}_j = \nicefrac{1}{n}(\rho^{(k)}_j+\textstyle\sum_{i}\pi^{(k)}_{ij})^{\shortminus 1}\rho^{(k)}_j\,,$$
$$\pi^{(k{+}1)} = \nicefrac{1}{n}(\rho^{(k)}_j+\textstyle\sum_{i}\pi^{(k)}_{ij})^{\shortminus 1}\pi^{(k)}_{ij} $$
where the results follow from \textit{method of Lagrange multipliers}.
Similarly, for $\mathcal{P}_{\mathcal{C}_2}^{KL}(\rho^{(k)},\pi^{(k)})$, we have:
$$\rho^{(k{+}1)} = \rho^{(k)} \,, \quad \pi^{(k{+}1)} = \tfrac{\mu}{\textstyle\sum_{ij}\pi^{(k)}_{ij}}\pi^{(k)}\, . $$

Given initialization, $(\rho^{(0)},\pi^{(0)})=(\bm{1}_n, \exp({\shortminus}\varepsilon c))$, we obtain the pairs $(\rho^{(k)}, \pi^{(k)})$ for $k=0,1,2,\ldots$ as:
\begin{equation}
\begin{split}
    \rho^{(k{+}1)} &= \nicefrac{1}{n}(\rho^{(k)}+\pi^{(k)\intercal}\bm{1}_m)^{\shortminus 1}\odot\rho^{(k)}\, , \quad \pi^{(k{+}1)} = \mu(\bm{1}_m\T\hat{\pi}\bm{1}_n)^{\shortminus 1}\hat{\pi} \\
    &\text{where}\quad \hat{\pi} = \pi^{(k)}Diag\big(\nicefrac{1}{n}(\rho^{(k)}+\pi^{(k)\intercal}\bm{1}_m)^{\shortminus 1}\big)
\end{split}
\end{equation}

\begin{proof}
We will prove by induction. From Proposition 4.2, we have
\[
\rho^{(k{+1})} = \nicefrac{1}{n}\,(1 + t^{(k)}\exp({\shortminus}\varepsilon c)^\intercal\bm{1}_m)^{\shortminus 1}\!,\,\,\, t^{(k{+}1)}=\mu\,(\bm{1}_m^\intercal\exp({\shortminus}\varepsilon c) \rho^{(k{+}1)})^{\shortminus 1}
\]
and $\pi^{(k)} = t^{(k)}\exp({\shortminus}\varepsilon c)Diag(\rho^{(k)})$. It is easy to show that the expressions hold for the pair $(\rho^{(1)}, \pi^{(1)})$. Now, assuming that the expressions also holds for arbitrary $(\rho^{(k^\prime)}, \pi^{(k^\prime)})$. We have
\[
\rho^{(k^\prime{+}1)} = \nicefrac{1}{n}(\rho^{(k^\prime)}+\pi^{(k^\prime)\intercal}\bm{1}_m)^{\shortminus 1}\odot\rho^{(k^\prime)}
\]
Replacing $\pi^{(k^\prime)} = t^{(k^\prime)}\exp({\shortminus}\varepsilon c)Diag(\rho^{(k^\prime)})$ we get:
\[
\rho^{(k^\prime{+}1)} = \nicefrac{1}{n}(\rho^{(k^\prime)}+Diag(\rho^{(k^\prime)})t^{(k^\prime)}\exp({\shortminus}\varepsilon c)\T\bm{1}_m)^{\shortminus 1}\odot\rho^{(k^\prime)}
\]
where $\rho^{(k^\prime)}$ terms cancel out, resulting in:
\[
\rho^{(k^\prime{+}1)} = \nicefrac{1}{n}(1+t^{(k^\prime)}\exp({\shortminus}\varepsilon c)\T\bm{1}_m)^{\shortminus 1}.
\]
Similarly, we express $\hat{\pi}$ as:
\[
\hat{\pi} = t^{(k^\prime)} \exp({\shortminus}\varepsilon c) Diag(\rho^{k^\prime})
Diag\Big(\nicefrac{1}{n}\big(\rho^{(k^\prime)}+Diag(\rho^{(k^\prime)})t^{(k^\prime)}\exp({\shortminus}\varepsilon c)\T\bm{1}_m\big)^{\shortminus 1}\Big)
\]
again  $\rho^{(k^\prime)}$ terms cancel out, resulting in:
\[
\hat{\pi} = t^{(k^\prime)} \exp({\shortminus}\varepsilon c)
Diag(\nicefrac{1}{n}(1+t^{(k^\prime)}\exp({\shortminus}\varepsilon c)\T\bm{1}_m)^{\shortminus 1}) = t^{(k^\prime)} \exp({\shortminus}\varepsilon c)
Diag(\rho^{(k^\prime+1)}).
\]
Hence, $\pi^{(k^\prime+1)}$ becomes:
\begin{equation*}
    \begin{split}
        \pi^{(k^\prime+1)} &=\mu(\bm{1}_m\T t^{(k^\prime)} \exp({\shortminus}\varepsilon c)Diag(\rho^{(k^\prime+1)})\bm{1}_n)^{\shortminus 1}t^{(k^\prime)} \exp({\shortminus}\varepsilon c)Diag(\rho^{(k^\prime+1)})\\
        & = \tfrac{1}{t^{(k^\prime)}}\underbrace{\mu(\bm{1}_m\T \exp({\shortminus}\varepsilon c)\rho^{(k^\prime+1)})^{\shortminus 1}}_{=t^{(k^\prime+1)}}t^{(k^\prime)} \exp({\shortminus}\varepsilon c)Diag(\rho^{(k^\prime+1)})\\
        &= t^{(k^\prime+1)}\exp({\shortminus}\varepsilon c)Diag(\rho^{(k^\prime+1)}),
    \end{split}
\end{equation*}
meaning that the expressions also hold for the pair $(\rho^{(k^\prime+1)}, \pi^{(k^\prime+1)})$.
\end{proof}

\subsection{Proof for Proposition 4.3}
\label{sec:proof4.2}

\begin{proof}
We start our proof by expressing (P2$^\prime$) in a compact form as:
\begin{equation*}
    x^{(\varepsilon)} =  \argmin_{\substack{x\geqslant 0 \\
Ax=b}} \bar{c}\T x + \tfrac{1}{\varepsilon}x\T(\log x -\bm{1}_{(m+1)n})
\end{equation*}
where $x$ denotes the vector formed by stacking $\rho$ and the row vectors of $\pi$, $\bar{c}$ denotes the vector formed by stacking $n$-dimensional zero vector and the row vectors of $c$, and $A$ and $b$ are such that $Ax=b$ imposes transport constraints as:
\[
A = \begin{bmatrix}
I_{n\sxtimes n}&{\smash{\overbrace{\begin{matrix}
        I_{n\sxtimes n}&\dotsb&I_{n\sxtimes n}
    \end{matrix}}^m}}\\
\bm{0}\T_n & \bm{1}_{m\,n}\T
\end{bmatrix}\, , 
\quad
b = [\nicefrac{1}{n}\bm{1}_n\T \quad \mu]\T
\]
From \textit{Lagrangian dual}, we have:
\[
x^{(\varepsilon)} = \exp({\shortminus}\varepsilon(\bar{c}{+}A\T\lambda^\ast))
\]
where $\lambda^\ast$ is the optimal dual Lagrangian satisfying:
\[
A\exp({\shortminus}\varepsilon(\bar{c}{+}A\T\lambda^\ast))=b
\]
Defining $[\tfrac{\partial x}{\partial c}]_{ij}\coloneqq \tfrac{\partial x_j}{\partial c_i}$, we have;
\[
\tfrac{\partial x^{(\varepsilon)}}{\partial c} = -\varepsilon\bar{I}(I +\tfrac{\partial\lambda^\ast}{\partial \bar{c}}A)Diag(x^{(\varepsilon)})
\]
where $\bar{I}\coloneqq[\bm{0}_{(mn){\sxtimes} n}\quad I_{(mn) {\sxtimes} ((m{+}1)n)}]$. Similarly, for the dual variable, we have:
\[
-\varepsilon(I+\tfrac{\partial\lambda^\ast}{\partial \bar{c}}A)Diag(x^{(\varepsilon)})A\T = 0 \Rightarrow \tfrac{\partial\lambda^\ast}{\partial \bar{c}} = \shortminus Diag(x^{(\varepsilon)})A\T(A Diag(x^{(\varepsilon)})A\T)^{\shortminus 1}.
\]
Putting back the expression for $\tfrac{\partial\lambda^\ast}{\partial \bar{c}}$ in  $\tfrac{\partial x^{(\varepsilon)}}{\partial c}$, we obtain:
\[
\tfrac{\partial x^{(\varepsilon)}}{\partial c} = -\varepsilon\bar{I}\big(Diag(x^{(\varepsilon)}) - Diag(x^{(\varepsilon)})A\T(A Diag(x^{(\varepsilon)})A\T)^{\shortminus 1}A Diag(x^{(\varepsilon)})\big),
\]
which includes $(m{+}1)$ by $n$ matrix inversion, $H\coloneqq A Diag(x^{(\varepsilon)})A\T$. We now show that $H^{\shortminus 1}$ can be obtained without explicit matrix inversion.

$H$ can be expressed as:
\[
H = \begin{bmatrix}
        \nicefrac{1}{n}I & \nicefrac{1}{n} - \rho\\
        \nicefrac{1}{n} - \rho\T & \mu
    \end{bmatrix}
\]
$H$ is Hermitian and positive definite. Using block matrix inversion formula for such matrices (Corrolary 4.1 of \cite{lu2002inverses}), we obtain the inverse as;
\[
H^{\shortminus 1} = \begin{bmatrix}
        nI + k^{\shortminus 1}\hat{\rho}\hat{\rho}\T  & -k^{\shortminus 1}\hat{\rho}\\
        -k^{\shortminus 1}\hat{\rho}\T & k^{\shortminus 1}
    \end{bmatrix}
\]
where $k=1-\mu-n\rho^{(\varepsilon)\intercal}\rho^{(\varepsilon)}$ and $\hat{\rho}=1-n\rho^{(\varepsilon)}$.

Given $\tfrac{\partial\mathcal{L}}{\partial x^{(\varepsilon)}}$, \ie, $(\tfrac{\partial\mathcal{L}}{\partial \rho^{(\varepsilon)}}, \tfrac{\partial\mathcal{L}}{\partial \pi^{(\varepsilon)}})$, the rest of the proof to obtain $\tfrac{\partial\mathcal{L}}{\partial c}$ follows from right multiplying the Jacobian, \ie, $\tfrac{\partial\mathcal{L}}{\partial c}=\tfrac{\partial x^{(\varepsilon)}}{\partial c}\,\tfrac{\partial\mathcal{L}}{\partial x^{(\varepsilon)}}$ and rearranging the terms.
\end{proof}

%% file: supplementary/figures/table_method_diff.tex
\begin{table}[hb]
\renewcommand{\arraystretch}{1.5}
\centering
\caption{Comparing our pooling method with OT-based pooling}
\label{tab:diff_pooling}
\resizebox{\textwidth}{!}{%
\begin{tabular}{@{}llll@{}}
\toprule
Item$\downarrow$ Method$\rightarrow$ & $\underset{\text{\cite{naderializadeh2021pooling}}}{\text{Ensemble of SWD Monge Maps}}$ & $\underset{\text{\cite{kolouri2020wasserstein, mialon2021trainable}}}{\text{OT Monge Map}}$                                                        & Ours (GSP)                                                                                                                                                              \\ \midrule
optimization problem                 & $s$\textsuperscript{$\dagger$}-many 1D OT                                                           & $\underset{\substack{\pi\geq 0 \\ \Sigma_i \pi_{ij} = \nicefrac{1}{n} \\ \Sigma_j \pi_{ij} = \nicefrac{1}{m}}}{\mathrm{argmin}} \Sigma_{ij}c_{ij}\pi_{ij}$ & $\underset{\substack{\rho, \pi\geq 0 \\ \rho_j+\Sigma_i \pi_{ij} = \nicefrac{1}{n} \\ \Sigma_{ij} \pi_{ij} = \mu}}{\mathrm{argmin}} \Sigma_{ij}c_{ij}\pi_{ij}$ \\
image representation                 & $[g_1 \mid g_2 \mid \cdots\mid g_s]$\textsuperscript{$\dagger$}            & $\sqrt{m}[f_1 \mid f_2 \mid \cdots\mid f_n]\pi^{\intercal}\quad$                                                                                                 & $\Sigma_i \tfrac{1 - n\rho_i}{n\mu}f_i$                                                                                                                                 \\
dimension                            & $m\times s$                                                                             & $m\times d$                                                                                                                                                      & $d$                                                                                                                                                                     \\
feature selection                    & \xmark                                                                                & \xmark                                                                                                                                                         & \cmark                                                                                                                                                                \\
gradient computation                 & auto-diff                                                                               & auto-diff                                                                                                                                                        & closed form expression                                                                                                                                                  \\
matrix-inverse-free gradient         & \cmark                                                                                & \xmark                                                                                                                                                         & \cmark                                                                                                                                                                \\ \bottomrule
\end{tabular}%
}
{\footnotesize\textsuperscript{$\dagger$} $s$: number of slices, $g_i$: projection of $\{f_j\}_j$ to slice-$i$ with sorted $j$ according to Monge Map.}
\end{table}

%% file: supplementary/ot_operators.tex
\section{Optimal Transport Based Operators}
\label{sec:otp_comparison}
In this section, we briefly discuss optimal transport (OT) based aggregation and selection operators. We provide their formulations to show how our formulation differs from them.

\subsection{Feature Aggregation}
Given a cost map $c_{ij}=\Vert \omega_i \shortminus f_j\Vert_2$ which is an $m$ (number of prototypes) by $n$ (number of features) matrix representing the closeness of prototypes $\omega_i$ and features $f_j$, \cite{mialon2021trainable} consider the following OT problem:
\begin{equation}\tag{P4}\label{eq:mialon}
\pi^{\ast} = \!\!\!\argmin_{\substack{\pi\geqslant0 \\
\Sigma_i\pi_{ij}=\nicefrac{1}{n}\\
\Sigma_j\pi_{ij}=\nicefrac{1}{m}}}
\!\!\! \textstyle\sum_{ij}c_{ij}\pi_{ij}
\end{equation}
and defines their aggregated feature as:
\begin{equation}\label{eq:ensemble_g}
    g = \sqrt{m}[f_1 \mid f_2 \mid \cdots \mid f_n] \pi\T \,\, .
\end{equation}
Namely, $g$ is an ensemble of $m$ vectors each of which is the weighted aggregation of $\lbrace f_i\rbrace_{i\in[n]}$ with the weights proportional to the assignment weights to the corresponding prototype. The same aggregation scheme is also discovered within the context of linear \emph{Wasserstein} embeddings via \emph{Monge maps} and is shown to be a barycentric projection of the feature set with the transport plan to approximate \emph{Monge map} \cite{kolouri2020wasserstein}. Similar to them, ensemble of Monge maps corresponding to sliced-Wasserstein distances (SWD) are further employed in set aggregation \cite{naderializadeh2021pooling}. In such ensemble representations, there is no feature selection mechanism and thus, all the features are somehow included in the image representation. 

For instance, if we further sum those $m$ vectors of $g$ in \cref{eq:ensemble_g} to obtain a single global representation, we end up with global average pooling: \( g\T\bm{1}_m = \sqrt{m}[f_1 \mid f_2 \mid \cdots \mid f_n] \pi\T\bm{1}_m = \nicefrac{\sqrt{m}}{n}[f_1 \mid f_2 \mid \cdots \mid f_n] \bm{1}_n = \nicefrac{\sqrt{m}}{n}\Sigma_i f_i\). 

Briefly, those optimal transport based set aggregators, \cite{kolouri2020wasserstein, mialon2021trainable, naderializadeh2021pooling} map a set of features to another set of features without discarding any and do not provide a natural way to aggregate the class-discriminative subset of the features. Such representation schemes are useful for structural matching. Albeit enabling $\ell 2$ metric as a similarity measure for the sets, their ensemble based representation results in very high dimensional embeddings. On the contrary, our formulation effectively enables learning to select discriminative features and maps a set of features to a single feature that is of the same dimension and is distilled from nuisance information. We summarize the comparison of our pooling method with the optimal transport based counterparts in \Cref{tab:diff_pooling}.

\subsection{Top-$k$ Selection}
Given $n$-many scalars as $x=[ x_i ]_{i\in[n]}$ and $m$-many scalars as $y=[ y_i ]_{i\in[m]}$ with $y$ is in an increasing family, \ie, $y_1{<}y_2<\ldots$, \cite{xie2020differentiable} considers the following OT:
\begin{equation}\tag{P5}\label{eq:xie}
\pi^{\ast} = \!\!\argmin_{\substack{\pi\geqslant0 \\
\Sigma_i\pi_{ij}=q_j\\
\Sigma_j\pi_{ij}=p_i}}
\!\! \textstyle\sum_{ij}c_{ij}\pi_{ij}
\end{equation}
where $c_{ij} = \vert y_i - x_j \vert$ and $p$ is $m$-dimensional probability simplex, \ie, $p\in\lbrace p{\in} \Real_{\leq 0}^m \mid \Sigma_i p_i {=}1\rbrace$. Then, top-$k$ selection is formulated with the setting $q=\nicefrac{1}{n}\bm{1}_n$, $y=[0, 1]$ and $p=[\tfrac{k}{n}\,\,\tfrac{n{\shortminus}k}{n}]\T$. Similarly, sorted top-$k$ selection is formulated with the setting $y=[k]$ and $p=[\tfrac{1}{n}\cdots\tfrac{1}{n}\,\,\tfrac{n{\shortminus}k}{n}]\T$. Solving the same problem \eqref{eq:xie}, \cite{cuturi2019differentiable} formulate top-$k$ ranking by letting $q$ and $p$ be arbitrary probability simplex and $y$ be in an increasing family.

Such top-$k$ formulations are suitable for selecting/ranking scalars. In our problem, the aim is to select a subset of features that are closest to the prototypes which are representatives for the discriminative information. Namely, we have a problem of subset selection from set-to-set distances. If we had our problem in the form of set-to-vector, then we would be able to formulate the problem using \eqref{eq:xie}. However, there is no natural extension of the methods in \cite{xie2020differentiable, cuturi2019differentiable} to our problem. Therefore, we rigorously develop an OT based formulation to express a discriminative subset selection operation analytically in a differentiable form.

Our formulation in (P1) differs from the typical optimal transport problem exploited in \eqref{eq:xie}. In optimal transport, one matches two distributions and transports all the mass from one to the other. Differently, we transport $\mu$ portion of the uniformly distributed masses to the prototypes that have no restriction on their mass distribution. In our formulation, we have a portion constraint instead of a target distribution constraint, and we use an additional decision variable, $\rho$, accounting for residual masses. If we do not have $\rho$ and set $\mu = 1$, then the problem becomes a specific case of an optimal transport barycenter problem with 1 distribution. 

Our problem can be expressed in a compact form by absorbing $\rho$ into $\pi$ with zero costs associated in the formulation, as in the proof (\cref{sec:proof4.2}). We choose to explicitly define $\rho$ in the problem (P1) to show its role. We believe its residual mass role is more understandable this way. The benefits of our formulation include that we can perform feature selection with matrix inversion free Jacobian and we can change the role of the prototypes as background representatives simply by using $\rho$ to weight the features instead of $\nicefrac{1}{n} {\shortminus} \rho$ in Eq. (4.1). Our specific formulation further allows us to tailor a zero-shot regularization loss built on the learned prototypes within our pooling layer.

%% file: supplementary/figures/algo_pooling.tex
\renewcommand{\algorithmicwhile}{\textbf{repeat}}
\renewcommand{\algorithmicdo}{\ }

\renewcommand{\algorithmiccomment}[1]{\hfill\eqparbox[h]{COMMENT}{// #1}}
\renewcommand{\algorithmicrequire}{\textbf{input:}}
\renewcommand{\algorithmicensure}{\textbf{trainable parameters:}}

\makeatletter
\newcommand{\algrule}[1][.2pt]{\par\vskip.5\baselineskip\hrule height #1\par\vskip.5\baselineskip}
\makeatother

\begin{algorithm}[H]
\caption{Deep Metric Learning Loss with GSP and ZSR}\label{algo:dml_with_gsp}
\begin{algorithmic}
\Require $(X,Y) = (\lbrace x_i\rbrace, \lbrace y_i\rbrace)_{i\in b}$ \algorithmiccomment{a batch of image-label pairs}
\State $F \gets \mathrm{Backbone}(X)$ \algorithmiccomment{a CNN backbone such as \textit{BN-Inception, ResNet}}
\State $(X_p, Z) \gets \lbrace \mathrm{GSP}(f)\rbrace_{f\in F}$ \algorithmiccomment{get pooled features and attribute predictions, see \cref{algo:gsp_pooling}}
\State $\mathcal{L}_{ZSR} \gets \mathrm{ZSR}(Z, Y)$  \algorithmiccomment{compute ZSR loss, see \cref{algo:zsr_loss}}
\State $\mathcal{L}_{DML} \gets \mathrm{LossDML}(X_p, Y)$  \algorithmiccomment{a DML loss such as \textit{contrastive, triplet, XBM, LIBC, ...}}
\State $\mathcal{L} \gets (1{\shortminus}\lambda)\mathcal{L}_{DML} + \lambda\mathcal{L}_{ZSR}$ \algorithmiccomment{we set $\lambda{=}0.1$}
\end{algorithmic}
\textbf{return}  $\mathcal{L}$
\end{algorithm}
\begin{algorithm}[H]
\caption{GSP($f$)}\label{algo:gsp_pooling}
\textbf{trainable parameters:} $\omega=\lbrace \omega_i\rbrace_{i\in[m]}$ \quad \algorithmiccomment{$m$-many prototypes}
\algrule[.5pt]
\begin{algorithmic}
\Require $f=\lbrace f_i\rbrace_{i\in[n]}$ \algorithmiccomment{feature map, $n=w\,h$ (\ie $\mathrm{width}\sxtimes\mathrm{height}$)}
\State $\bar{\omega}_i\gets \nicefrac{\omega_i}{\max\lbrace 1, \Vert \omega_i \Vert_2\rbrace}$ $\forall i{\in}[m]$
\State $\bar{f}_j\gets \nicefrac{f_j}{\max\lbrace 1, \Vert f_j \Vert_2\rbrace}$ $\forall j{\in}[n]$ 
\State $c_{ij} \gets \Vert \bar{\omega}_i \shortminus \bar{f}_j \Vert_2$ \algorithmiccomment{cost map, $c=\{c_{ij}\}_{(i,j)\in[m]\sxtimes[n]}$}
\State $\rho, \pi \gets \mathrm{WeightTransport}(c)$ \algorithmiccomment{see \cref{algo:feature_selection}}
\State $f \gets \tfrac{1{\shortminus}n\rho}{\mu}\odot f $ \algorithmiccomment{re-weight features, $\odot$: element-wise multiplication}
\State $x_p \gets (\tfrac{1}{n}\textstyle\sum\limits_{i\in[n]} f_i^p)^{\nicefrac{1}{p}}$ \algorithmiccomment{pooled feature, GSP for $p{=}1$, GeMean+GSP for $p{>}1$}
\State $z_i \gets \tfrac{1}{\mu}\textstyle\sum\limits_{j\in[n]}\pi_{ij}$ $\forall i{\in}[m]$ \algorithmiccomment{\textit{attribute} predictions, $z=\lbrace z_i\rbrace_{i\in[m]}$}
\end{algorithmic}
\textbf{return} $x_p$, $z$
\end{algorithm}

\begin{algorithm}[H]
\caption{WeightTransport($c$)}\label{algo:feature_selection}
\textbf{hyperparameters:} $\mu:$ transport ratio, $\varepsilon:$ entropy regularization weight, $k:$ number of iterations
\algrule[.5pt]
\textbf{forward:} gets cost map, $c$, returns residual weights, $\rho$, and transport plan $\pi$
\algrule
\begin{algorithmic}
\Require $c=\{c_{ij}\}_{(i,j)\in[m]\sxtimes[n]}$ \algorithmiccomment{cost map of $m$-many prototypes and $n$-many features}
\State $\kappa \gets \mathrm{exp}({\shortminus}\varepsilon c)$, $t\gets 1$ \algorithmiccomment{$\mathrm{exp}$ is element-wise}
\While{$k$ times}
	\State $\rho \gets \nicefrac{1}{n}(1+t\,\kappa\T\bm{1}_m)^{\shortminus 1}$ \algorithmiccomment{$A\T\bm{1}_m$: sum $A$ along rows, $(\cdot)^{\shortminus 1}$ is element-wise}
	\State $t \gets \mu(\bm{1}_m\T\kappa\,\rho)^{\shortminus 1}$
\EndWhile
\end{algorithmic}
\textbf{return} $\rho,\, t\,\kappa\,Diag(\rho)$ \algorithmiccomment{$\pi \gets t\,\kappa\,Diag(\rho)$\\}
\algrule[.5pt]
\textbf{backward:} gets the solution ($\rho$, $\pi$) and the gradients ($\frac{\partial \mathcal{L}}{\partial \rho}, \frac{\partial \mathcal{L}}{\partial \pi}$), returns $\frac{\partial \mathcal{L}}{\partial c}$
\algrule
\begin{algorithmic}
\Require $\rho,\,\, \pi,\,\, \frac{\partial \mathcal{L}}{\partial \rho},\,\, \frac{\partial \mathcal{L}}{\partial \pi} $ \algorithmiccomment{results of forward pass and the loss gradient w.r.t. them}
\State $q \gets \rho \odot\frac{\partial \mathcal{L}}{\partial \rho} + (\pi\odot\frac{\partial \mathcal{L}}{\partial \pi})^\intercal\bm{1}_m$ \algorithmiccomment{$A\T\bm{1}_m$: sum $A$ along rows, $\odot$: element-wise multiplication}
\State $\eta \gets (\rho \odot \frac{\partial \mathcal{L}}{\partial \rho})^\intercal\bm{1}_n\shortminus n\,q^\intercal\rho$
\State $\frac{\partial \mathcal{L}}{\partial c} \gets \shortminus\varepsilon\Big(\pi \odot\frac{\partial \mathcal{L}}{\partial \pi} - n\pi Diag\big(q- \tfrac{\eta}{1\shortminus\mu\shortminus n\rho\T\rho}
\big)\rho\Big)$
\end{algorithmic}
\textbf{return} $\frac{\partial \mathcal{L}}{\partial c}$
\end{algorithm}

\begin{algorithm}[H]
\caption{ZSR($Z$, $Y$)}\label{algo:zsr_loss}
\textbf{trainable parameters:} $\Upsilon=\lbrace\upsilon_i\rbrace_{i\in[\#\text{classes}]}$ \quad \algorithmiccomment{a semantic embedding vector for each class label}
\algrule[.5pt]
\begin{algorithmic}
\Require $Z{=}\lbrace z_i\rbrace_{i\in b}$, $Y{=}\lbrace y_i\rbrace_{i\in b}$ \algorithmiccomment{a batch, $b$, of attribute prediction vectors, $z_i$, and their labels, $y_i$}
\State $(b_1, b_2) \gets$ split $b$ into two class-disjoint halves s.t. $\lbrace y_i \rbrace_{i\in b_1 } {\cap} \lbrace y_i \rbrace_{i\in b_2 } = \emptyset$
\State $\Upsilon_k \gets [\upsilon_{y_i}]_{i\in b_k}$ for $k{=}1,2$ \algorithmiccomment{label embedding matrix for batch-$k$, \ie prediction targets}
\State $Z_k \gets [z_i]_{i\in b_k}$ for $k{=}1,2$ \algorithmiccomment{attribute prediction matrix for batch-$k$, \ie prediction inputs}
\State $A_k \gets \Upsilon_k(Z_k\T Z_k + \epsilon I)^{\shortminus 1}Z_k\T$ for $k{=}1,2$ \quad \algorithmiccomment{fit label embedding predictor for batch-$k$, $\epsilon{=}0.05$}
\State $\hat{\Upsilon}_1 \gets A_2 Z_1$, $\hat{\Upsilon}_2 \gets A_1 Z_2$ \algorithmiccomment{use predictor for $b_k$ to predict the label embeddings of $b_{k^{\prime}}$}
\State $\hat{\Upsilon} \gets [\hat{\Upsilon}_1 \mid \hat{\Upsilon}_2]$ \algorithmiccomment{concatenate predictions}
\State $S \gets \mathrm{SoftMax}(\hat{\Upsilon}\T \Upsilon)$ \algorithmiccomment{similarity scores between predictions and label embeddings}
\State $\mathcal{L}_{ZSR} \gets \mathrm{CrossEntropy}(S, Y)$
\end{algorithmic}
\textbf{return} $\mathcal{L}_{ZSR}$
\end{algorithm}

%% file: iccv_2023_paper.bbl
\begin{thebibliography}{10}

\bibitem{kim2018attention}
W.Kim, B.Goyal, K.Chawla, J.Lee, and K.Kwon,
\newblock ``Attention-based ensemble for deep metric learning,''
\newblock in {\em Proceedings of the European Conference on Computer Vision
  (ECCV)}, 2018, pp. 736--751.

\bibitem{lin2018deep}
X.Lin, Y.Duan, Q.Dong, J.Lu, and J.Zhou,
\newblock ``Deep variational metric learning,''
\newblock in {\em Proceedings of the European Conference on Computer Vision
  (ECCV)}, 2018.

\bibitem{ermolov2022hyperbolic}
A.Ermolov, L.Mirvakhabova, V.Khrulkov, N.Sebe, and I.Oseledets,
\newblock ``Hyperbolic vision transformers: Combining improvements in metric
  learning,''
\newblock in {\em Proceedings of the IEEE/CVF Conference on Computer Vision and
  Pattern Recognition}, 2022, pp. 7409--7419.

\bibitem{musgrave2020metric}
K.Musgrave, S.Belongie, and S.-N.Lim,
\newblock ``A metric learning reality check,''
\newblock in {\em European Conference on Computer Vision}. Springer, 2020, pp.
  681--699.

\bibitem{roth2020revisiting}
K.Roth, T.Milbich, S.Sinha, P.Gupta, B.Ommer, and J.~P.Cohen,
\newblock ``Revisiting training strategies and generalization performance in
  deep metric learning,''
\newblock in {\em International Conference on Machine Learning}. PMLR, 2020,
  pp. 8242--8252.

\bibitem{venkataramanan2022it}
S.Venkataramanan, B.Psomas, E.Kijak, laurent amsaleg, K.Karantzalos, and
  Y.Avrithis,
\newblock ``It takes two to tango: Mixup for deep metric learning,''
\newblock in {\em International Conference on Learning Representations}, 2022.

\bibitem{gurbuz2023generalizable}
Y.~Z.Gurbuz and A.~A.Alatan,
\newblock ``Generalizable embeddings with cross-batch metric learning,''
\newblock {\em arXiv preprint arXiv:2307.07620}, 2023.

\bibitem{zeiler2014visualizing}
M.~D.Zeiler and R.Fergus,
\newblock ``Visualizing and understanding convolutional networks,''
\newblock in {\em European conference on computer vision}. Springer, 2014, pp.
  818--833.

\bibitem{zhou2016learning}
B.Zhou, A.Khosla, A.Lapedriza, A.Oliva, and A.Torralba,
\newblock ``Learning deep features for discriminative localization,''
\newblock in {\em Proceedings of the IEEE conference on computer vision and
  pattern recognition}, 2016.

\bibitem{zhou2018interpreting}
B.Zhou, D.Bau, A.Oliva, and A.Torralba,
\newblock ``Interpreting deep visual representations via network dissection,''
\newblock {\em IEEE transactions on pattern analysis and machine intelligence},
  2018.

\bibitem{gurbuz2019novel}
Y.~Z.G{\"u}rb{\"u}z and A.~A.Alatan,
\newblock ``A novel bovw mimicking end-to-end trainable cnn classification
  framework using optimal transport theory,''
\newblock in {\em 2019 IEEE International Conference on Image Processing
  (ICIP)}. IEEE, 2019, pp. 3053--3057.

\bibitem{xu2020attribute}
W.Xu, Y.Xian, J.Wang, B.Schiele, and Z.Akata,
\newblock ``Attribute prototype network for zero-shot learning,''
\newblock {\em Advances in Neural Information Processing Systems}, vol. 33, pp.
  21969--21980, 2020.

\bibitem{cuturi2019differentiable}
M.Cuturi, O.Teboul, and J.-P.Vert,
\newblock ``Differentiable ranking and sorting using optimal transport,''
\newblock {\em Advances in neural information processing systems}, vol. 32,
  2019.

\bibitem{demirel2017attributes2classname}
B.Demirel, R.Gokberk~Cinbis, and N.Ikizler-Cinbis,
\newblock ``Attributes2classname: A discriminative model for attribute-based
  unsupervised zero-shot learning,''
\newblock in {\em Proceedings of the IEEE international conference on computer
  vision}, 2017, pp. 1232--1241.

\bibitem{ko2020embedding}
B.Ko and G.Gu,
\newblock ``Embedding expansion: Augmentation in embedding space for deep
  metric learning,''
\newblock in {\em Proceedings of the IEEE/CVF Conference on Computer Vision and
  Pattern Recognition}, 2020, pp. 7255--7264.

\bibitem{liu2021noise}
C.Liu, H.Yu, B.Li, Z.Shen, Z.Gao, P.Ren, X.Xie, L.Cui, and C.Miao,
\newblock ``Noise-resistant deep metric learning with ranking-based instance
  selection,''
\newblock in {\em Proceedings of the IEEE/CVF Conference on Computer Vision and
  Pattern Recognition}, 2021, pp. 6811--6820.

\bibitem{proxysynthesis}
G.Gu, B.Ko, and H.-G.Kim,
\newblock ``Proxy synthesis: Learning with synthetic classes for deep metric
  learning,''
\newblock in {\em Proceedings of the AAAI Conference on Artificial
  Intelligence}, 2021, vol.~35, pp. 1460--1468.

\bibitem{wang2020cross}
X.Wang, H.Zhang, W.Huang, and M.~R.Scott,
\newblock ``Cross-batch memory for embedding learning,''
\newblock in {\em Proceedings of the IEEE/CVF Conference on Computer Vision and
  Pattern Recognition}, 2020, pp. 6388--6397.

\bibitem{teh2020proxynca++}
E.~W.Teh, T.DeVries, and G.~W.Taylor,
\newblock ``Proxynca++: Revisiting and revitalizing proxy neighborhood
  component analysis,''
\newblock in {\em European Conference on Computer Vision (ECCV)}. Springer,
  2020.

\bibitem{dong2020generalization}
M.Dong, X.Yang, R.Zhu, Y.Wang, and J.Xue,
\newblock ``Generalization bound of gradient descent for non-convex metric
  learning,''
\newblock Neural Information Processing Systems Foundation, 2020.

\bibitem{lei2021generalization}
Y.Lei, M.Liu, and Y.Ying,
\newblock ``Generalization guarantee of sgd for pairwise learning,''
\newblock {\em Advances in Neural Information Processing Systems}, vol. 34,
  2021.

\bibitem{gurbuz2021asap}
Y.~Z.Gurbuz, O.Can, and A.~A.Alatan,
\newblock ``Deep metric learning with chance constraints,''
\newblock {\em arXiv preprint arXiv:2209.09060}, 2022.

\bibitem{Roth_2019_ICCV}
K.Roth, B.Brattoli, and B.Ommer,
\newblock ``Mic: Mining interclass characteristics for improved metric
  learning,''
\newblock in {\em The IEEE International Conference on Computer Vision (ICCV)},
  October 2019.

\bibitem{intrabatch}
J.Seidenschwarz, I.Elezi, and L.Leal{-}Taix{\'{e}},
\newblock ``Learning intra-batch connections for deep metric learning,''
\newblock in {\em Proceedings of the 38th International Conference on Machine
  Learning, {ICML} 2021, 18-24 July 2021, Virtual Event}. 2021, vol. 139 of
  {\em Proceedings of Machine Learning Research}, pp. 9410--9421, {PMLR}.

\bibitem{lim2022hypergraph}
J.Lim, S.Yun, S.Park, and J.~Y.Choi,
\newblock ``Hypergraph-induced semantic tuplet loss for deep metric learning,''
\newblock in {\em Proceedings of the IEEE/CVF Conference on Computer Vision and
  Pattern Recognition}, 2022.

\bibitem{patel2022recall}
Y.Patel, G.Tolias, and J.Matas,
\newblock ``Recall@ k surrogate loss with large batches and similarity mixup,''
\newblock in {\em Proceedings of the IEEE/CVF Conference on Computer Vision and
  Pattern Recognition}, 2022, pp. 7502--7511.

\bibitem{Jacob_2019_ICCV}
P.Jacob, D.Picard, A.Histace, and E.Klein,
\newblock ``Metric learning with horde: High-order regularizer for deep
  embeddings,''
\newblock in {\em The IEEE International Conference on Computer Vision (ICCV)},
  October 2019.

\bibitem{zhang2020deepSEC}
D.Zhang, Y.Li, and Z.Zhang,
\newblock ``Deep metric learning with spherical embedding,''
\newblock {\em Advances in Neural Information Processing Systems}, vol. 33,
  2020.

\bibitem{profs}
O.Can, Y.~Z.G{\"u}rb{\"u}z, and A.~A.Alatan,
\newblock ``Deep metric learning with alternating projections onto feasible
  sets,''
\newblock in {\em 2021 IEEE International Conference on Image Processing
  (ICIP)}. IEEE, 2021, pp. 1264--1268.

\bibitem{kim2021multi}
Y.Kim and W.Park,
\newblock ``Multi-level distance regularization for deep metric learning,''
\newblock in {\em Proceedings of the AAAI Conference on Artificial
  Intelligence}, 2021, vol.~35.

\bibitem{roth2022non}
K.Roth, O.Vinyals, and Z.Akata,
\newblock ``Non-isotropy regularization for proxy-based deep metric learning,''
\newblock in {\em Proceedings of the IEEE/CVF Conference on Computer Vision and
  Pattern Recognition}, 2022, pp. 7420--7430.

\bibitem{xuan2018deep}
H.Xuan, R.Souvenir, and R.Pless,
\newblock ``Deep randomized ensembles for metric learning,''
\newblock in {\em Proceedings of the European Conference on Computer Vision
  (ECCV)}, 2018, pp. 723--734.

\bibitem{Sanakoyeu_2019_CVPR}
A.Sanakoyeu, V.Tschernezki, U.Buchler, and B.Ommer,
\newblock ``Divide and conquer the embedding space for metric learning,''
\newblock in {\em The IEEE Conference on Computer Vision and Pattern
  Recognition (CVPR)}, June 2019.

\bibitem{zheng2021deep2}
W.Zheng, C.Wang, J.Lu, and J.Zhou,
\newblock ``Deep compositional metric learning,''
\newblock in {\em Proceedings of the IEEE/CVF Conference on Computer Vision and
  Pattern Recognition}, 2021, pp. 9320--9329.

\bibitem{zheng2021deep}
W.Zheng, B.Zhang, J.Lu, and J.Zhou,
\newblock ``Deep relational metric learning,''
\newblock in {\em Proceedings of the IEEE/CVF International Conference on
  Computer Vision}, 2021, pp. 12065--12074.

\bibitem{milbich2020diva}
T.Milbich, K.Roth, H.Bharadhwaj, S.Sinha, Y.Bengio, B.Ommer, and J.~P.Cohen,
\newblock ``Diva: Diverse visual feature aggregation for deep metric
  learning,''
\newblock in {\em European Conference on Computer Vision}. Springer, 2020, pp.
  590--607.

\bibitem{roth2021s2sd}
K.{Roth}, T.{Milbich}, B.{Ommer}, J.~P.{Cohen}, and M.{Ghassemi},
\newblock ``S2sd: Simultaneous similarity-based self-distillation for deep
  metric learning,''
\newblock in {\em ICML 2021: 38th International Conference on Machine
  Learning}, 2021, pp. 9095--9106.

\bibitem{zhang2017deep}
H.Zhang, J.Xue, and K.Dana,
\newblock ``Deep ten: Texture encoding network,''
\newblock in {\em Proceedings of the IEEE conference on computer vision and
  pattern recognition}, 2017, pp. 708--717.

\bibitem{vlad}
R.Arandjelovic, P.Gronat, A.Torii, T.Pajdla, and J.Sivic,
\newblock ``Netvlad: Cnn architecture for weakly supervised place
  recognition,''
\newblock in {\em Proceedings of the IEEE conference on computer vision and
  pattern recognition}, 2016, pp. 5297--5307.

\bibitem{kolouri2020wasserstein}
S.Kolouri, N.Naderializadeh, G.~K.Rohde, and H.Hoffmann,
\newblock ``Wasserstein embedding for graph learning,''
\newblock in {\em International Conference on Learning Representations}, 2020.

\bibitem{mialon2021trainable}
G.Mialon, D.Chen, A.d'Aspremont, and J.Mairal,
\newblock ``A trainable optimal transport embedding for feature aggregation and
  its relationship to attention,''
\newblock in {\em ICLR 2021-The Ninth International Conference on Learning
  Representations}, 2021.

\bibitem{kancoded}
S.KAN, Y.Liang, M.Li, Y.Cen, J.Wang, and Z.He,
\newblock ``Coded residual transform for generalizable deep metric learning,''
\newblock in {\em Advances in Neural Information Processing Systems}.

\bibitem{crow}
Y.Kalantidis, C.Mellina, and S.Osindero,
\newblock ``Cross-dimensional weighting for aggregated deep convolutional
  features,''
\newblock in {\em European conference on computer vision}. Springer, 2016, pp.
  685--701.

\bibitem{tamsk}
G.Tolias, T.Jenicek, and O.Chum,
\newblock ``Learning and aggregating deep local descriptors for instance-level
  recognition,''
\newblock in {\em European Conference on Computer Vision}. Springer, 2020, pp.
  460--477.

\bibitem{cbam}
S.Woo, J.Park, J.-Y.Lee, and I.~S.Kweon,
\newblock ``Cbam: Convolutional block attention module,''
\newblock in {\em Proceedings of the European conference on computer vision
  (ECCV)}, 2018, pp. 3--19.

\bibitem{delf}
H.Noh, A.Araujo, J.Sim, T.Weyand, and B.Han,
\newblock ``Large-scale image retrieval with attentive deep local features,''
\newblock in {\em Proceedings of the IEEE International Conference on Computer
  Vision (ICCV)}, Oct 2017.

\bibitem{gsop}
Z.Gao, J.Xie, Q.Wang, and P.Li,
\newblock ``Global second-order pooling convolutional networks,''
\newblock in {\em Proceedings of the IEEE/CVF Conference on Computer Vision and
  Pattern Recognition}, 2019, pp. 3024--3033.

\bibitem{cuturi2013sinkhorn}
M.Cuturi,
\newblock ``Sinkhorn distances: Lightspeed computation of optimal transport,''
\newblock {\em Advances in neural information processing systems}, vol. 26,
  2013.

\bibitem{zhao2021towards}
W.Zhao, Y.Rao, Z.Wang, J.Lu, and J.Zhou,
\newblock ``Towards interpretable deep metric learning with structural
  matching,''
\newblock in {\em Proceedings of the IEEE/CVF International Conference on
  Computer Vision}, 2021, pp. 9887--9896.

\bibitem{xie2020differentiable}
Y.Xie, H.Dai, M.Chen, B.Dai, T.Zhao, H.Zha, W.Wei, and T.Pfister,
\newblock ``Differentiable top-k with optimal transport,''
\newblock {\em Advances in Neural Information Processing Systems}, vol. 33, pp.
  20520--20531, 2020.

\bibitem{naderializadeh2021pooling}
N.Naderializadeh, J.~F.Comer, R.Andrews, H.Hoffmann, and S.Kolouri,
\newblock ``Pooling by sliced-wasserstein embedding,''
\newblock {\em Advances in Neural Information Processing Systems}, vol. 34, pp.
  3389--3400, 2021.

\bibitem{luise2018differential}
G.Luise, A.Rudi, M.Pontil, and C.Ciliberto,
\newblock ``Differential properties of sinkhorn approximation for learning with
  wasserstein distance,''
\newblock {\em Advances in Neural Information Processing Systems}, vol. 31,
  2018.

\bibitem{wu2017sampling}
C.-Y.Wu, R.Manmatha, A.~J.Smola, and P.Krahenbuhl,
\newblock ``Sampling matters in deep embedding learning,''
\newblock in {\em Proceedings of the IEEE International Conference on Computer
  Vision}, 2017, pp. 2840--2848.

\bibitem{schroff2015facenet}
F.Schroff, D.Kalenichenko, and J.Philbin,
\newblock ``Facenet: A unified embedding for face recognition and clustering,''
\newblock in {\em Proceedings of the IEEE conference on computer vision and
  pattern recognition}, 2015, pp. 815--823.

\bibitem{Wang_2019_CVPR_MS}
X.Wang, X.Han, W.Huang, D.Dong, and M.~R.Scott,
\newblock ``Multi-similarity loss with general pair weighting for deep metric
  learning,''
\newblock in {\em The IEEE Conference on Computer Vision and Pattern
  Recognition (CVPR)}, June 2019.

\bibitem{huynh2020fine}
D.Huynh and E.Elhamifar,
\newblock ``Fine-grained generalized zero-shot learning via dense
  attribute-based attention,''
\newblock in {\em Proceedings of the IEEE/CVF conference on computer vision and
  pattern recognition}, 2020, pp. 4483--4493.

\bibitem{bertinetto2018meta}
L.Bertinetto, J.~F.Henriques, P.Torr, and A.Vedaldi,
\newblock ``Meta-learning with differentiable closed-form solvers,''
\newblock in {\em International Conference on Learning Representations}, 2018.

\bibitem{he2016identity}
K.He, X.Zhang, S.Ren, and J.Sun,
\newblock ``Identity mappings in deep residual networks,''
\newblock in {\em European conference on computer vision}. Springer, 2016, pp.
  630--645.

\bibitem{cifar}
A.Krizhevsky and G.Hinton,
\newblock ``Learning multiple layers of features from tiny images,''
\newblock Tech. {R}ep., Citeseer, 2009.

\bibitem{russakovsky2015imagenet}
O.Russakovsky, J.Deng, H.Su, J.Krause, S.Satheesh, S.Ma, Z.Huang, A.Karpathy,
  A.Khosla, M.Bernstein, et~al.,
\newblock ``Imagenet large scale visual recognition challenge,''
\newblock {\em International journal of computer vision}, vol. 115, no. 3, pp.
  211--252, 2015.

\bibitem{normalization2015accelerating}
S.Ioffe and C.Szegedy,
\newblock ``Batch normalization: Accelerating deep network training by reducing
  internal covariate shift,''
\newblock in {\em International conference on machine learning}. PMLR, 2015,
  pp. 448--456.

\bibitem{abadi2016tensorflow}
M.Abadi, P.Barham, J.Chen, Z.Chen, A.Davis, J.Dean, M.Devin, S.Ghemawat,
  G.Irving, M.Isard, et~al.,
\newblock ``Tensorflow: a system for large-scale machine learning.,''
\newblock in {\em OSDI}, 2016, vol.~16, pp. 265--283.

\bibitem{fehervari2019unbiased}
I.Fehervari, A.Ravichandran, and S.Appalaraju,
\newblock ``Unbiased evaluation of deep metric learning algorithms,''
\newblock {\em arXiv preprint arXiv:1911.12528}, 2019.

\bibitem{oh2016deep}
H.Oh~Song, Y.Xiang, S.Jegelka, and S.Savarese,
\newblock ``Deep metric learning via lifted structured feature embedding,''
\newblock in {\em Proceedings of the IEEE Conference on Computer Vision and
  Pattern Recognition}, 2016, pp. 4004--4012.

\bibitem{wah2011caltech}
C.Wah, S.Branson, P.Welinder, P.Perona, and S.Belongie,
\newblock ``The caltech-ucsd birds-200-2011 dataset,''
\newblock 2011.

\bibitem{krause2014submodular}
A.Krause and D.Golovin,
\newblock ``Submodular function maximization,''
\newblock in {\em Tractability: Practical Approaches to Hard Problems}, pp.
  71--104. Cambridge University Press, 2014.

\bibitem{liu2016deepfashion}
Z.Liu, P.Luo, S.Qiu, X.Wang, and X.Tang,
\newblock ``Deepfashion: Powering robust clothes recognition and retrieval with
  rich annotations,''
\newblock in {\em Proceedings of the IEEE conference on computer vision and
  pattern recognition}, 2016, pp. 1096--1104.

\bibitem{kingma2014adam}
D.~P.Kingma and J.Ba,
\newblock ``Adam: A method for stochastic optimization,''
\newblock {\em arXiv preprint arXiv:1412.6980}, 2014.

\bibitem{hadsell2006dimensionality}
R.Hadsell, S.Chopra, and Y.LeCun,
\newblock ``Dimensionality reduction by learning an invariant mapping,''
\newblock in {\em 2006 IEEE Computer Society Conference on Computer Vision and
  Pattern Recognition (CVPR'06)}. IEEE, 2006, vol.~2, pp. 1735--1742.

\bibitem{kim2020proxy}
S.Kim, D.Kim, M.Cho, and S.Kwak,
\newblock ``Proxy anchor loss for deep metric learning,''
\newblock in {\em Proceedings of the IEEE/CVF Conference on Computer Vision and
  Pattern Recognition}, 2020, pp. 3238--3247.

\bibitem{gmeanp}
F.Radenovi{\'c}, G.Tolias, and O.Chum,
\newblock ``Fine-tuning cnn image retrieval with no human annotation,''
\newblock {\em IEEE transactions on pattern analysis and machine intelligence},
  vol. 41, no. 7, pp. 1655--1668, 2018.

\bibitem{zhu2020fewer}
Y.Zhu, M.Yang, C.Deng, and W.Liu,
\newblock ``Fewer is more: A deep graph metric learning perspective using fewer
  proxies,''
\newblock {\em Advances in Neural Information Processing Systems}, vol. 33, pp.
  17792--17803, 2020.

\bibitem{tan2021efficientnetv2}
M.Tan and Q.Le,
\newblock ``Efficientnetv2: Smaller models and faster training,''
\newblock in {\em International conference on machine learning}. PMLR, 2021,
  pp. 10096--10106.

\bibitem{murray2016interferences}
N.Murray, H.J{\'e}gou, F.Perronnin, and A.Zisserman,
\newblock ``Interferences in match kernels,''
\newblock {\em IEEE transactions on pattern analysis and machine intelligence},
  vol. 39, no. 9, pp. 1797--1810, 2016.

\bibitem{solar}
T.Ng, V.Balntas, Y.Tian, and K.Mikolajczyk,
\newblock ``Solar: second-order loss and attention for image retrieval,''
\newblock in {\em European conference on computer vision}. Springer, 2020, pp.
  253--270.

\bibitem{dosovitskiy2021image}
A.Dosovitskiy, L.Beyer, A.Kolesnikov, D.Weissenborn, X.Zhai, T.Unterthiner,
  M.Dehghani, M.Minderer, G.Heigold, S.Gelly, et~al.,
\newblock ``An image is worth 16x16 words: Transformers for image recognition
  at scale,''
\newblock in {\em International Conference on Learning Representations}, 2021.

\bibitem{weldon}
T.Durand, N.Thome, and M.Cord,
\newblock ``Weldon: Weakly supervised learning of deep convolutional neural
  networks,''
\newblock in {\em Proceedings of the IEEE conference on computer vision and
  pattern recognition}, 2016, pp. 4743--4752.

\bibitem{bregman1967relaxation}
L.~M.Bregman,
\newblock ``The relaxation method of finding the common point of convex sets
  and its application to the solution of problems in convex programming,''
\newblock {\em USSR computational mathematics and mathematical physics}, vol.
  7, no. 3, pp. 200--217, 1967.

\bibitem{bauschke2000dykstras}
H.~H.Bauschke and A.~S.Lewis,
\newblock ``Dykstras algorithm with bregman projections: A convergence proof,''
\newblock {\em Optimization}, vol. 48, no. 4, pp. 409--427, 2000.

\bibitem{lu2002inverses}
T.-T.Lu and S.-H.Shiou,
\newblock ``Inverses of 2$\times$ 2 block matrices,''
\newblock {\em Computers \& Mathematics with Applications}, vol. 43, no. 1-2,
  pp. 119--129, 2002.

\end{thebibliography}
